\documentclass[10pt,twocolumn,letterpaper]{article}

\usepackage[review]{cvpr}      

\usepackage[dvipsnames]{xcolor}
\usepackage{multirow} 

\newcommand{\hec}[1]{\textcolor{red}{[Hadar]: {#1}}}

\newcommand{\ruojin}[1]{\textcolor{blue}{[Ruojin]: {#1}}}
\newcommand{\hani}[1]{\textcolor{magenta}{[Hani]: {#1}}}

\long\def\ignorethis#1{}

\newcommand{\dataset}{\emph{ExtremeLandmarkPairs}\xspace}
\newcommand{\shortdataset}{\emph{ELP}\xspace}
\newcommand{\cambridge}{\emph{s}ELP\xspace}
\newcommand{\megadepth}{\emph{w}ELP\xspace}
\newcommand{\fovexp}{$\Delta \text{FoV}$\xspace}
\newcommand{\instructexp}{$\Delta \text{Im}$\xspace}
\newcommand{\casenc}{CascadedAtt Encoder\xspace}

\newbox\jsavebox

 \newif\ifcomment

\definecolor{cvprblue}{rgb}{0.21,0.49,0.74}
\usepackage[pagebackref,breaklinks,colorlinks,citecolor=cvprblue]{hyperref}

\title{Extreme Rotation Estimation in the Wild}

\date{}
\author{{\centering Hana Bezalel$^{1}$ \: Dotan Ankri$^{1}$ \: Ruojin Cai$^{2}$ \: Hadar Averbuch-Elor$^{1,2}$ 
        }
\\
{\parbox{0.9\textwidth}{\centering
$^1$Tel Aviv University \quad
        $^2$Cornell University  
       }
}
\\
\\
{\parbox{\textwidth}{\centering
\small{\url{https://tau-vailab.github.io/ExtremeRotationsInTheWild/}}       }
}
}

\begin{document}
\maketitle
\begin{abstract}
{
We present a technique and benchmark dataset for estimating the relative 3D orientation between a pair of Internet images captured in an extreme setting, where the images have limited or non-overlapping field of views.  
Prior work targeting extreme rotation estimation assume constrained 3D environments and emulate perspective images by cropping regions from panoramic views. However, real images captured in the wild are highly diverse, exhibiting variation in both appearance and camera intrinsics.
In this work, we propose a Transformer-based method for estimating relative rotations in extreme real-world settings, and contribute the \textbf{\dataset{}} dataset, assembled from scene-level Internet photo collections. 
Our evaluation demonstrates that our approach succeeds in estimating the relative rotations in a wide variety of extreme-view Internet image pairs, outperforming various baselines, including dedicated rotation estimation techniques and contemporary 3D reconstruction methods.

}
\end{abstract}

\section{Introduction}
\label{sec:intro}

The problem of estimating the relative 3D orientation between a pair of images is embodied in fundamental computer vision tasks, such as camera localization~\cite{brachmann2017dsac,sattler2018benchmarking,sarlin2021back} and 3D reconstruction~\cite{schonberger2016structure,ozyecsil2017survey,wang2021multi}.
Establishing pixel correspondences (either explicitly or implicitly) is typically a prerequisite for computing the relative rotation between the images.
Correspondences, however, cannot be extracted in extreme settings where the images have little to no overlap. 
%
As dense imagery may not necessarily be available for many practical applications, a natural question arises: How can we estimate the relative rotation between non-overlapping RGB images, without the use of additional data (such as depth or temporal information)? 
\begin{figure}[t]
    \centering
    \includegraphics[width=0.47\textwidth,trim={150 100 150 100},clip]
    {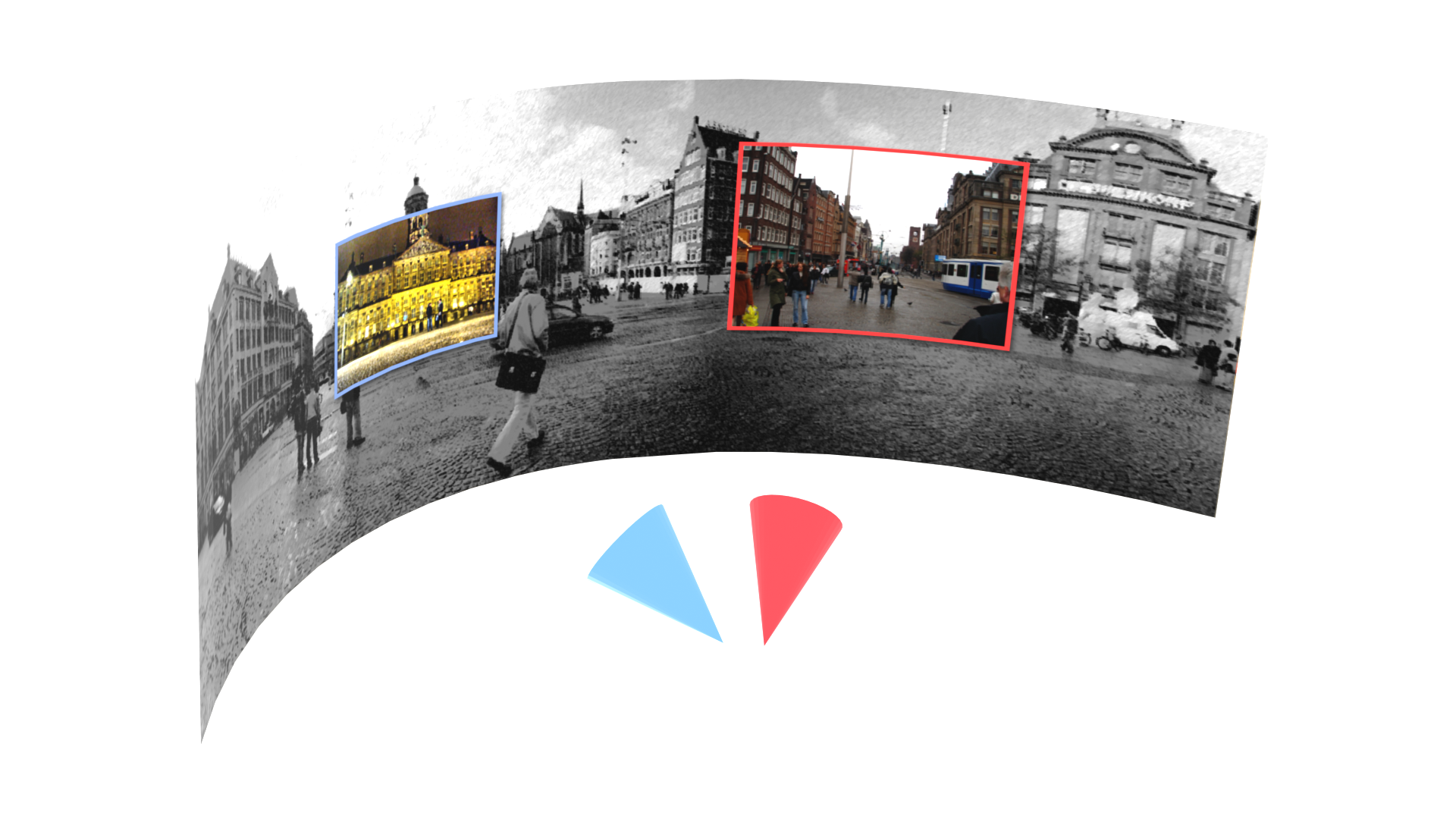}
    \caption{Given a pair of (possibly) non-overlapping images captured \emph{in the wild}---\emph{e.g.}, under arbitrary illumination and intrinsic camera parameters---such as the images of the Dam Square in Amsterdam depicted in red and blue boxes above$^*$, our technique estimates the relative 3D rotation between the images. $^*$The background panorama is illustrated for visualization purposes only.}
    \label{fig:teaser}
\end{figure}

We have recently seen pioneering efforts addressing the task of relative rotation estimation in such extreme non-overlapping settings~\cite{cai2021extreme,dekel2024estimating}. 
Prior work has proposed end-to-end neural architectures, demonstrating that \emph{hidden} cues, such as vanishing points or the directions of cast shadows, can implicitly guide the model for inferring the relative orientation between the images. To facilitate learning and evaluation, datasets constructed from panoramic views were adopted. These datasets emulate perspective views by cropping sub-areas from these panoramas, enabling generation of image pairs with various degrees of overlap. 
However, while such emulated views perhaps capture some of the challenges associated with extreme-view imagery, are they sufficient for representing real images---particularly, images captured \emph{in the wild}?

In this paper, we present a new approach that tackles the problem of extreme rotation estimation in the wild. Consider the boxed images in Figure \ref{fig:teaser}. Internet (\emph{i.e.}, in the wild) images may vary due to a wide range of factors, including  transient objects, weather conditions,  time of day, and the cameras' intrinsic parameters. To explore this problem, we introduce a new dataset, \dataset{} (\shortdataset{}), assembled from publicly-available scene-level Internet image collections. We observe that the set of real extreme-view image pairs is limited, as Internet datasets are typically scene-centric, with nearby cameras commonly capturing overlapping views. Therefore, to facilitate training, we propose a progressive learning scheme that leverages and augments images cropped from panoramic views, allowing for gradually generalizing the model onto real Internet data. In particular, we construct datasets with varying field of views, that better resemble the distribution of real data, and perform image-level appearance augmentations by leveraging recent advancements in text-to-image Diffusion models~\cite{rombach2022high,hertz2022prompt,brooks2023instructpix2pix}.

To estimate extreme rotations in the wild, we propose a Transformer-based model that is provided with auxiliary channels, including the spatial distribution of local keypoints and matches and semantic segmentation maps, allowing for better reasoning over real image pairs with little or no overlap. Our results demonstrate that our model can accurately predict the relative rotations for a wide variety of extreme-view image pairs that vary in illumination, dynamic regions, and intrinsic parameters. We conduct extensive experiments, quantifying performance both over real Internet data and also over emulated perspective images cropped from panoramic views. Our evaluation shows that our model significantly improves over strong baselines when considering real images, while achieving comparable performance over emulated perspective image pairs.

\section{Related Work}

Relative pose estimation is a fundamental task in computer vision, typically studied for overlapping camera views. Traditionally, this task has been divided into two stages: correspondence estimation from local feature matching, followed by geometry-based pose estimation. 
In recent years, feature matching methods have advanced from using heuristic feature descriptors~\cite{lowe2004distinctive, bay2006surf, rublee2011orb} and RANSAC-based matches~\cite{fischler1981random} to learning-based feature extraction~\cite{detone2018superpoint,Dusmanu2019CVPR,tyszkiewicz2020disk,edstedt2023dedode,wang2020learning,balntas2018relocnet,Ding2019camnet} and matching techniques~\cite{sarlin2020superglue, lindenberger2023lightglue,baboud2011automatic}
, 
with several methods performing both feature extraction and matching using unified learning-based frameworks~\cite{sun2021loftr,edstedt2023dkm,balntas2018relocnet}. 

These methods are generally 
invariant to changes in illumination and appearance, demonstrating robust performance across various scene scales and also over in-the-wild datasets. 
However, their reliability diminishes in extreme view scenarios due to their dependence on visual overlap.

Extreme-view scenarios lacking local pixelwise correspondences necessitate the use of end-to-end learning-based pose estimation methods which directly infer the 3D relationships and geometry from sparse and extreme-view images.
Indeed, large-scale 3D object datasets have paved the way for learning-based methods which estimate camera pose directly from sparse views~\cite{zhang2022relpose,yang2022fvor,fan2023pope,sinha2023sparsepose,lin2023relpose++,wang2023posediffusion,zhang2024cameras,wang2023pf}. However, these methods primarily concentrate on object-centric scenes, under experimental settings that typically assume similar lighting and camera intrinsics for the input views. Furthermore, these methods often utilize bounding box inputs defining the object of interest, which is less suitable in the case of images depicting large-scale scenes. 

Several prior works address a sparse view setting at scene-scale. In particular, Chen \emph{et al.}~\cite{chen2021wide} propose to learn a discrete distributions of pose space, Agarwala \emph{et al.}~\cite{agarwala2022planeformers} simplify scene reconstruction using a plane representation, and Rockwell \emph{et al.}~\cite{rockwell20228} introduce an inductive bias of the 8-point algorithm into a vision transformer architecture. Recently, models that directly predict pixel-aligned point maps from input pairs (or sparse image collections), such as DUSt3R~\cite{wang2023dust3r} and Mast3R~\cite{leroy2024groundingimagematching3d}, have demonstrated promising results on pose estimation and scene-scale 3D reconstruction of Internet data with a wide baseline.

Camera pose estimation for non-overlapping views presents a greater challenge.
Earlier efforts~\cite{caspi2002aligning} explored searching for consistent temporal behavior.
Several work utilize pairwise RGB and depth scan data to estimate relative pose among such extreme pairs~\cite{yang2019extreme,yang2020extreme}.
Cai et al.~\cite{cai2021extreme} tackle pose estimation for non-overlapping views without the use of additional data, by introducing a learning-based network leveraging cross-correlation volume to exploit implicit cues. This correlation volume is later enhanced through the integration of transformer attention modules~\cite{dekel2024estimating}. Nonetheless, these approaches assume constrained 3D environments, including the assumption of consistent lighting and camera intrinsics, and are designed for camera distribution of emulated perspective views cropped from panoramas. 
In this work, we aim to address pose estimation for \emph{realistic} in-the-wild non-overlapping image pairs, enhancing the applicability of extreme pose estimation to Internet photos and real-world data. 

\section{The \dataset{} Dataset}
\label{sec:dataset}
Prior works on extreme pose estimation use panoramic views, cropping from it sub-areas to emulate perspective views~\cite{cai2021extreme,dekel2024estimating}. To evaluate and train models on real perspective image pairs, we propose a new benchmark and dataset, \dataset{} (\shortdataset{}), constructed from Internet image pairs from the MegaDepth~\cite{li2018megadepth}, Cambridge Landmarks~\cite{kendall2015cambridgelandmarks}, and MegaScenes~\cite{tung2024megascenes} datasets. In this section, we first describe the dataset construction procedure (Section \ref{sec:construction}), and then present details regarding dataset size and train and test splits (Table \ref{tab:dataset-statistics-new}).

\begin{figure}[t]
    \centering
    \includegraphics[width=\columnwidth]{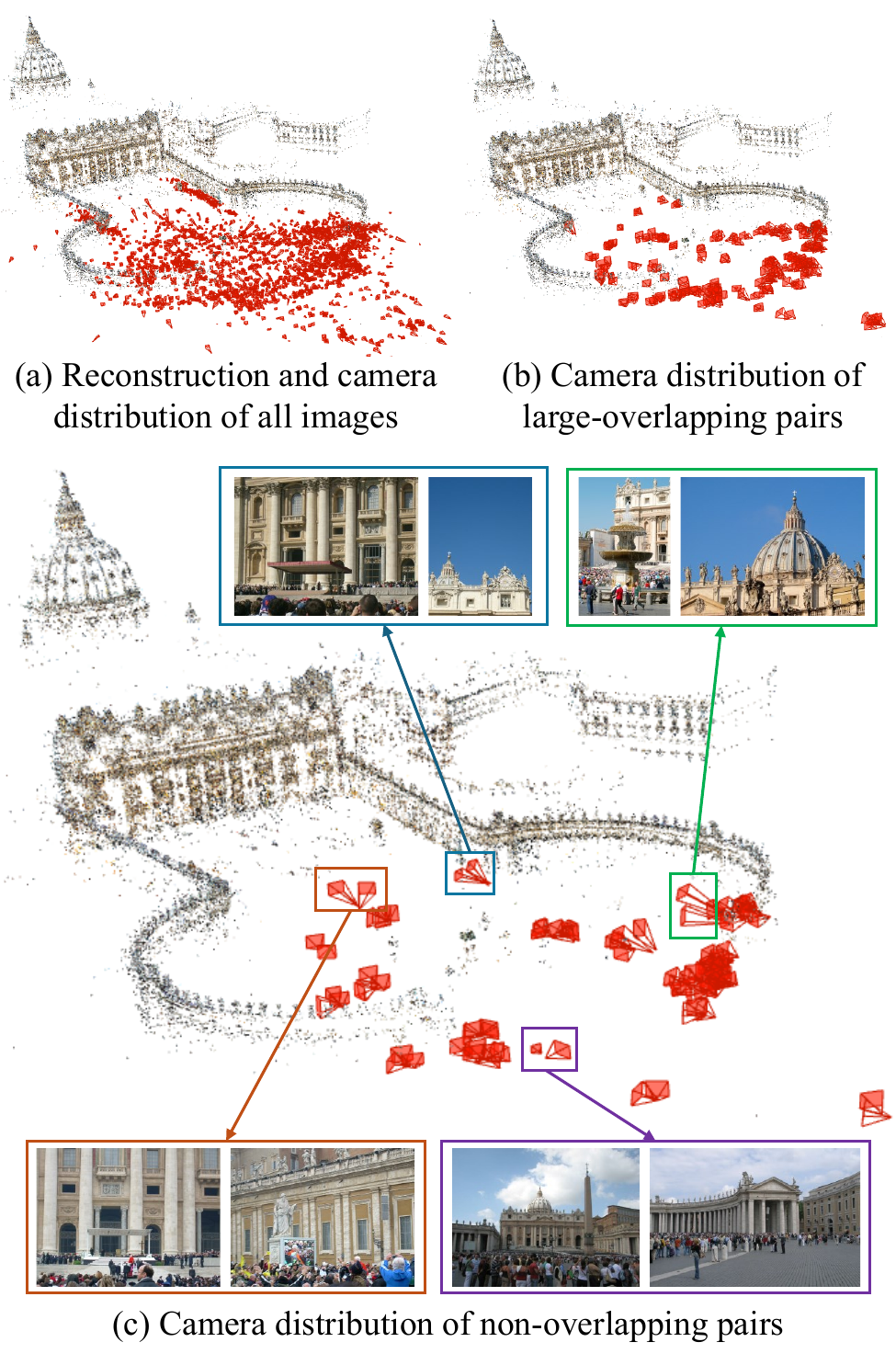}
    \vspace{-20pt}
    \caption{
    Camera distribution of the Vatican, Rome scene from the \dataset{} Dataset.
    We construct a dataset of real perspective image pairs with predominant rotational motion shown in (b) and (c) from the dense imagery reconstruction in (a).
    }
    \label{fig:dataset}
\end{figure}

\subsection{Dataset Construction}
\label{sec:construction}
To construct a dataset of \emph{real} perspective image pairs with varying degrees of overlap, we leverage available scene-level training data. Existing Internet image collections typically contain camera poses (predicted up to scale), which are determined using Structure-from-Motion (SfM) algorithms, such as COLMAP~\cite{schonberger2016structure}. In what follows, we describe how we extract real image pairs from this data, which can then be used for training and evaluating models.

\medskip
\noindent \textbf{Identifying Pairs with Predominant Rotational Motion}. Prior works targeting relative rotation estimation, in particular for non-overlapping views, mostly utilize panoramic views, focusing on image pairs with purely rotational motion. Pairs belonging to \emph{real} image collections almost always contain a non-negligible translation component. Furthermore, unlike in the StreetLearn~\cite{mirowski2019streetlearn} dataset used by prior work~\cite{cai2021extreme,dekel2024estimating} that provides exact translation values between consecutive panoramas which allows for filtering pairs with predominant rotational motions, reconstructed relative poses are only provided up to scale. 
The scale varies among different reconstructed scenes, and therefore
there's no global threshold on the relative translation values which can be used for identifying pairs with predominant rotational motion.

To automatically identify such pairs, we observe that available Internet collections require the existence of dense imagery, to compensate for the vast number of unknowns in the SfM optimization. We therefore construct \emph{mutual} nearest neighbors edge-weighted graphs, with one graph per landmark. In each graph $G$, nodes $v \in V$ correspond to images, and two images are connected by an edge $e \in E$ if they are both among each other's $K$ nearest neighbors, considering L2 distances between their translations ($K$ is empirically set to $5$). Note that images captured from sparser (outlier) regions in space are unlikely to be within the mutual $K$ nearest neighbors of images captured within denser regions, and hence won't be included in $G$. Finally, we select a subset of image pairs containing relatively small distances from each scene graph $G$, yielding a set of image pairs with predominant rotational motion; see the supplementary material for additional details.  

\medskip
\noindent \textbf{Extracting Level of Overlap}. Following prior work ~\cite{cai2021extreme,dekel2024estimating}, we are interested in training and evaluating models according to three different categories: \emph{Large}, \emph{Small} and \emph{None}, indicating image pairs with a varying amount of overlap. However, unlike prior work that use cropped images with a fixed $90^{\circ}$ FoV, Internet images contain varying FoV values. Thus, the relative rotation angle is not sufficient for extracting the pair's overlap level.  

Denote the FoV values of image $i\in [1,2]$ as $[\text{fov}^i_x,\text{fov}^i_y]$. We can parameterize a 3D rotation matrix $\mathbf{R}$ using three Euler angles $[\alpha,\beta,\gamma]$, denoting the relative roll, pitch and yaw angles, respectively:
\ignorethis{
\begin{equation}
        \mathbf{R}(\alpha, \beta, \gamma) = \mathbf{R}_z(\alpha)\mathbf{R}_x(\beta)\mathbf{R}_y(\gamma)
\end{equation}
\ruojin{is there a typo for switching the notation of $R_y$ and $R_x$, should it be updated as follows?}\hani{no. COLMAP coordinate system is defined like that.}\hec{Ruojin will double check}\ruojin{updated formula as below:}

\begin{equation}
       \mathbf{R}(\alpha, \beta, \gamma) = \mathbf{R}_z(\alpha)\mathbf{R}_y(\gamma)\mathbf{R}_x(\beta),
\end{equation}
following COLMAP's coordinate system.
}

\begin{equation}
       \mathbf{R}(\alpha, \beta, \gamma) = \mathbf{R}_z(\alpha)\mathbf{R}_y(\beta)\mathbf{R}_x(\gamma),
\end{equation}
following Dense Correlation Volumes' coordinate system.
We use the following conditions to determine the overlap level $o$:

\begin{equation}
    o=\left\{ \begin{array}{ll}
         \textit{Large} &  |\gamma| < \frac{\text{fov}^1_x + \text{fov}^2_x}{4}\land   |\beta| < \frac{\text{fov}^1_y + \text{fov}^2_y}{4}\\
        \textit{None} & |\gamma| > \frac{\text{fov}^1_x + \text{fov}^2_x}{2}\land   |\beta| > \frac{\text{fov}^1_y + \text{fov}^2_y}{2} \vspace{3pt}\\ 
        \textit{Small} & else \end{array} \right. 
\end{equation}
In other words, pairs with relative yaw and pitch angles that are smaller than a quarter of the average corresponding FoV values are considered as pairs with a large overlap ratio. Likewise, pairs with relative yaw and pitch angles that are larger than half the average corresponding FoV values are considered non-overlapping pairs. Pairs in-between these conditions are considered pairs with a small overlap ratio.

\medskip
\noindent \textbf{Additional Filtering}. In the scene-scale datasets we explore, large FoV disparities could result in one image focusing on specific architectural details like a statue, while the other captures a much broader scene perspective, further complicating the problem of estimating the relative rotations. To extract pairs consistent in scale, we limit the difference between the FoV values to be at most $5^\circ$. Furthermore, as most images contain small roll values, we rotate the scenes to match the gravity axis and horizontal axis and filtered images with rolls exceeding $10^\circ$. Finally, we excluded images captured from an aerial perspective, by filtering images with translation along y-axis exceeding a global threshold, empirically set to 1.

\begin{table}[t]
\setlength{\tabcolsep}{3.8pt}
 \def\arraystretch{0.95}
\centering
\resizebox{0.45\textwidth}{!}{%
\begin{tabular}{llcccccccccccc}
\toprule & & \multicolumn{4}{c}{\#Pairs} \\ 
\cline{3-7} 
Subset
& Source & \#Scenes &  Large & Small & None & Total  \\ 
\midrule
Train   & \cite{tung2024megascenes}  & 5883 & 33430                   & 13684    & 29481      & 76595                        \\
Validation   & \cite{tung2024megascenes}  & 515 & 3398                   & 710    & 4130      & 8238                        \\
Validation Balanced  & \cite{tung2024megascenes}  & 177 & 92                   & 55    & 707      & 8238                        \\
\cambridge{} & \cite{kendall2015cambridgelandmarks}    & 6 & 2512        & 827      & 1961    & 5300                            \\
\megadepth{}  & \cite{li2018megadepth} & 17 & 2700       & 829                      & 643                  &  4172                          \\

\bottomrule
\end{tabular}
}
\vspace{-8pt}
\caption{\textbf{\dataset{} Dataset Statistics}. Above, we report the number of image pairs extracted for each overlap level, split into train and test (with \cambridge{} denoting a \emph{single} camera setting and \megadepth{} denoting the ``in the \emph{wild}" setting).}
\label{tab:dataset-statistics-new}
\end{table}

\subsection{Dataset Size and Splits}
We apply the procedure described above over landmarks from the MegaScenes~\cite{tung2024megascenes} dataset to create a training set and a validation set. As we focus on outdoor scenes in our work, we filter reconstructions that capture indoor scenes. We obtain a total of nearly 34K non-overlapping pairs originating from over 2K unique landmarks.  The validation set was also balanced the set over the overall angle. 

For evaluation, we create two test sets, to separately examine image pairs captured in a \emph{single} camera setting with constant illumination (\cambridge{}) and image pairs captured in the \emph{wild} (\megadepth{}). Image pairs in the \cambridge{} test set contain images from the Cambridge Landmarks~\cite{kendall2015cambridgelandmarks} dataset, which contains videos capturing six different landmarks.
Image pairs in the \megadepth{} test set contain images from the MegaDepth dataset~\cite{li2018megadepth}, which contains Internet photos from Flickr for a set of large-scale landmarks. 

\begin{figure*}[t]
    \centering

    \includegraphics[width=0.98\textwidth]{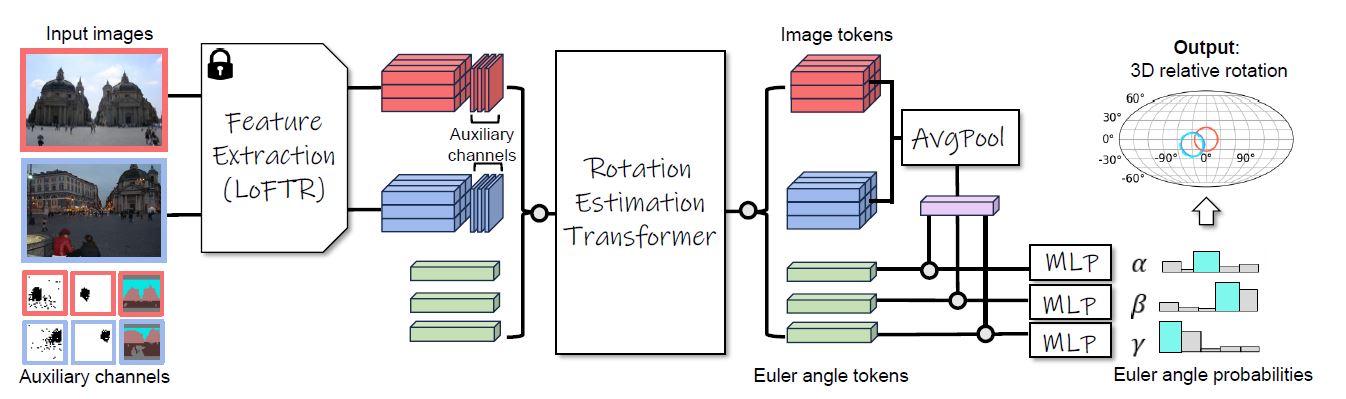}
    \vspace{-8pt}
    \caption{
    \textbf{Method architecture.} Given a pair of input Internet images, we extract image features using pretrained LoFTR. These features are combined with auxiliary channels, including keypoint and pairwise matches masks, and segmentation maps (visualized on the bottom left). 
    These image features are reshaped into tokens and concatenated with Euler angle position embeddings, which are then processed by our Rotation Estimation Transformer module. 
    The output Euler angle tokens and averaged image tokens are concatenated and processed by MLPs to predict the probability distribution of Euler angles, representing the relative 3D rotation between the input images.   
    }
    \label{fig:overview}
\end{figure*}

As there is some overlap between the landmarks in MegaScenes~\cite{tung2024megascenes} and MegaDepth~\cite{li2018megadepth}, we performed additional filtering, ensuring no overlap exists between the train and test sets. Furthermore, we filter test pairs if one of the images is \emph{highly} transient---\emph{i.e.}, if transient objects dominate the image. We quantify this using a pretrained segmentation model~\cite{enze2021segformer}, filtering images containing transient objects in over 40\% of the pixels. We also manually validate non-overlapping image pairs, allowing to further filter unlabeled objects, such as a stage or a market stand.
Finally, for non-overlapping image pairs, we observed that the overall relative rotations are highly imbalanced, and therefore we balance this set by the overall relative angle. Table \ref{tab:dataset-statistics-new} summarizes the number of image pairs and landmarks used for both training and test. 
\section{Method}

Given a pair of Internet images with (possibly) extreme relative motion, we estimate the relative rotation $\mathbf{R}$ between the images. Following prior works on extreme rotation estimation~\cite{cai2021extreme,dekel2024estimating}, we assume a camera-centric setting, where the two cameras have limited translation. However, our approach departs from prior works by operating on outdoor images captured by a crowd of photographers, with varying intrinsic parameters as well as appearances---\emph{e.g.}, due to illumination changes and dynamic objects.

Our model (detailed in Section \ref{sec:architecture}; see Figure \ref{fig:overview} for an overview) outputs three discrete Euler angles, denoting the relative roll, pitch and yaw angles. As illustrated in prior work~\cite{cai2021extreme}, this parameterization enables using a simple cross-entropy loss for training. In Section \ref{sec:train}, we describe our progressive learning scheme, allowing for gradually adapting the model to extreme Internet imagery.

 \subsection{Model} 
 \label{sec:architecture}

We extract image features using a pretrained LoFTR model~\cite{sun2021loftr}. In contrast to the features extracted using common convolutional neural networks pretrained on ImageNet~\cite{deng2009imagenet}, LoFTR is a Transformer-based model trained on Internet pairs, with the goal of extracting local feature matches -- a setting and task which is highly related to the one we address in our work, thus enabling extraction of better (\emph{i.e.}, more relevant) features.  

As we are interested in designing a network that can \emph{also} reason over image pairs with little or no overlap, we combine the extracted features with additional \emph{auxiliary} channels; see Figure~\ref{fig:overview} (bottom left). These include keypoint and pairwise matches masks, utilized previously for disambiguating images for similar structures~\cite{cai2023doppelgangers}. Intuitively, knowledge over pairwise matches can assist the model in cases of small overlap and for generalizing across different camera intrinsic properties. We also incorporate a segmentation map as an additional auxiliary channel, which segments images into several categories (such as sky, building, road and sidewalk). This channel allows for identifying additional cues, such as the skyline or transient objects, which can aid in determining the rotation for non-overlapping pairs. In Section \ref{sec:results}, we demonstrate the benefit of incorporating these auxiliary channels in our model.

We then reshape extracted features (and auxiliary channels) to tokens, concatenating these image tokens with learnable Euler angle position embeddings. These tokens are processed by our \emph{Rotation Estimation Transformer} module, which uses a transformer decoder architecture~\cite{vaswani2017attention}. The output Euler angle tokens obtain information from image features and auxiliary channels within transformer attention modules. These tokens are then processed by three different prediction heads (one per angle, denoted as \emph{MLP} in Figure \ref{fig:overview}).Each output prediction head uses as input averaged image tokens and one of the output Euler angle tokens, which provides the model with additional angle-specific information, allowing for achieving improved performance, as we show in the supplementary material. The prediction heads output a probability distribution over $N=360$ bins, capturing an angle in the range $[-180^\circ, 180^\circ]$.  

 \subsection{Learning} 
 \label{sec:train}
As detailed in Section \ref{sec:dataset}, we assemble real image pairs from Internet imagery, which can be used for training and evaluating models. However, even with our proposed \dataset{} dataset, the set of real image pairs is limited---\emph{e.g.}, only $\sim$36K non-overlapping image pairs are extracted, as available Internet datasets are typically scene-centric, with nearby cameras usually capturing overlapping views. Therefore, in what follows, we propose a \emph{progressive} learning scheme, which leverages panoramic images, and allows for gradually generalizing the model to images captured in-the-wild. All learning stages are optimized using a cross-entropy loss for each Euler angle prediction. Additional details are provided in the supplementary material.
 
\medskip \noindent \textbf{Initialization.} We begin by training our model on the perspective views cropped from panoramic images, using the data created by Cai et al~\cite{cai2021extreme}. Specifically, we use the image pairs cropped from panoramic images included in the StreetLearn~\cite{mirowski2019streetlearn} dataset, depicting various streets in Manhattan. As further detailed in Cai et al~\cite{cai2021extreme}, this training set includes roughly 1M image pairs sampled from the same panorama, split according to the overlap level.

 \begin{figure}[t]
\centering
\begin{tabular}{@{}l@{}c@{}c@{}c@{}}

    \makebox[58pt]{\includegraphics[height=1.95cm, trim={0 0 0 0},clip]{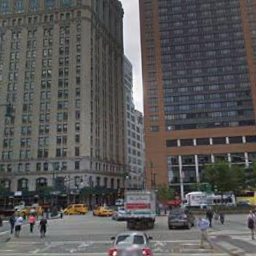}} \hspace{0.1cm} &
    \makebox[58pt]{\includegraphics[height=1.95cm, trim={0 0 0 0},clip]{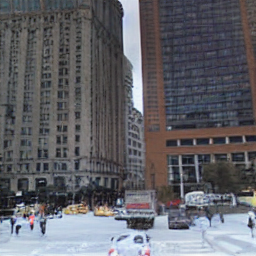}} &
    \makebox[58pt]{\includegraphics[height=1.95cm, trim={0 0 0 0},clip]{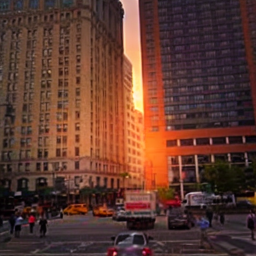}} &
    \makebox[58pt]{\includegraphics[height=1.95cm, trim={0 0 0 0},clip]{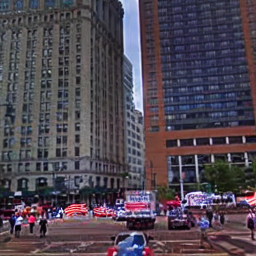}}\\
    \makebox[58pt]{} & \makebox[58pt]{\emph{snowy}} & \makebox[58pt]{\emph{sunset}} & \makebox[58pt]{\emph{4th of July}}\\
    \makebox[58pt]{\includegraphics[height=1.95cm, trim={0 0 0 0},clip]{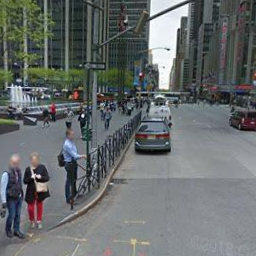}} \hspace{0.1cm} &
    \makebox[58pt]{\includegraphics[height=1.95cm, trim={0 0 0 0},clip]{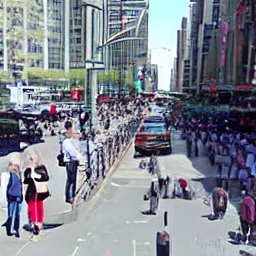}} &
    \makebox[58pt]{\includegraphics[height=1.95cm, trim={0 0 0 0},clip]{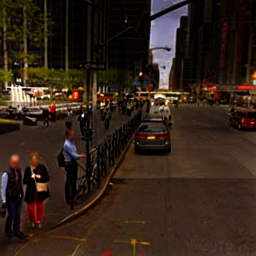}} &
    \makebox[58pt]{\includegraphics[height=1.95cm, trim={0 0 0 0},clip]{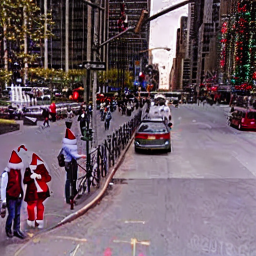}}\\
    \makebox[58pt]{} & \makebox[58pt]{\emph{busy street}} & \makebox[58pt]{\emph{night}} & \makebox[58pt]{\emph{Christmas}}

    \end{tabular}
    \vspace{-8pt}
    \caption{\textbf{Augmenting perspective images cropped from panoramic views with image-level appearance modifications.} Given an input image (left) and a target text prompt \emph{``Make it $\left<w\right>$''} ($\left<w\right>$ is specified above), we use a conditional Diffusion model~\cite{brooks2023instructpix2pix} to create semantic appearance augmentations which modify both the global image characteristics as well as local image regions. 
   }
    \label{fig:instruct_pair_before_and_after}
\end{figure}

\medskip \noindent \textbf{Training with Data Augmentations.} We observe that the dataset constructed by prior work could be augmented to better capture the distribution of image pairs captured in-the-wild. In particular, we focus on two types of data augmentations, which we elaborate on next: (i) field of view (FoV) augmentations (denoted henceforth as \fovexp{}) and (ii) image-level appearance augmentations (denoted henceforth as \instructexp{}). In Section \ref{sec:results}, we demonstrate how both types of augmentations improve the model's ability in generalizing to real Internet scenes.

Rather than cropping perspective images with a fixed $90^\circ$ FoV as was done in prior work, we analyze the FoV values of the images belonging to the \shortdataset{} training set. We compute the mean and standard-deviation values, denoted as $\mu$ and $\sigma$, respectively.  
We then construct new perspective images by sampling from a Gaussian distribution $G(\mu,a\cdot\sigma)$ that resembles the distribution of real data, setting $a=1.5$ for obtaining a more diverse set, which also bears higher similarity to the perspective images used during initialization. 
We also allow for FoV differences (of up to $5^\circ$) between the two images paired together. Additionally, rather than providing the model with the full content within these regions, we construct crops with various aspect ratios to further emulate real image pairs.

\begin{figure*}[ht]
    \centering

    \begin{tabular}{@{}c@{}r@{ }l@{}c@{}r@{ }l@{}c@{}r@{ }l@{}c@{}}
        \rotatebox{90}{\makebox[37pt][c]{Large}} &
        \includegraphics[height=1.25cm, trim={0 0 0 0},clip]{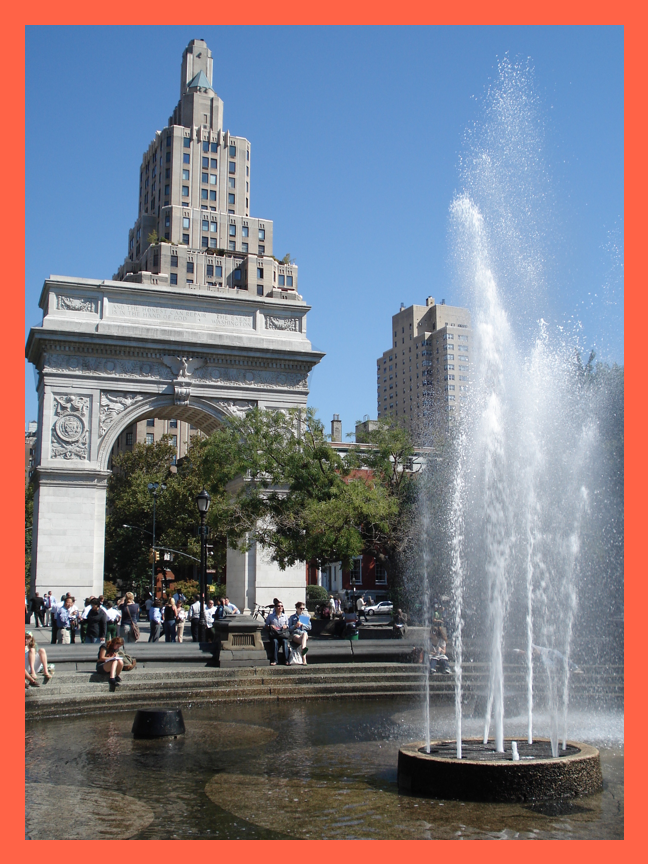}&
        \includegraphics[height=1.25cm, trim={0 0 0 0},clip]{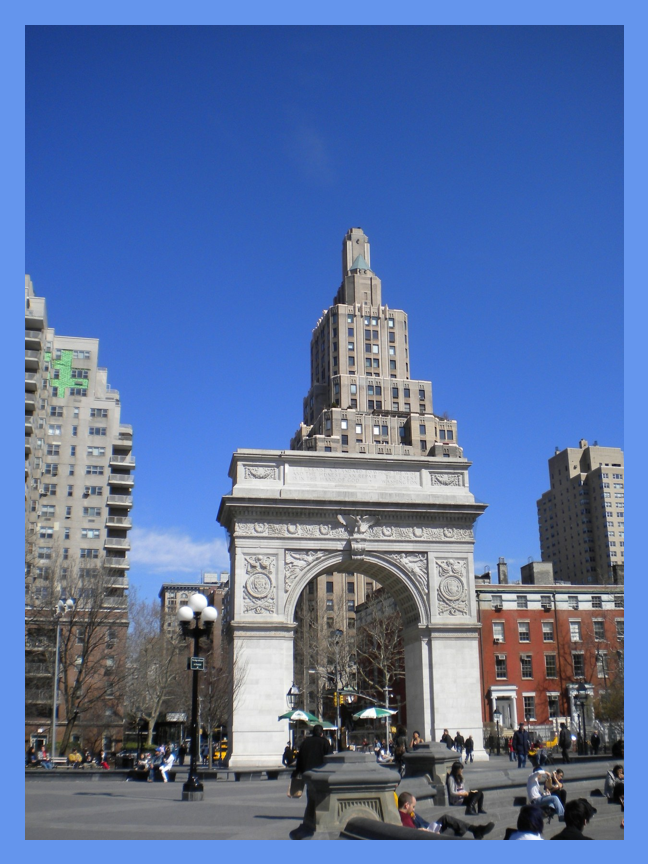} &
        \includegraphics[height=1.3cm, trim={0.2cm 0cm 0.4cm 0cm},clip]{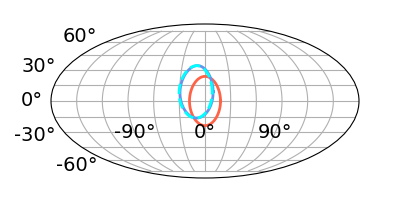} &
        \includegraphics[height=1.36cm, trim={0 0 0 0},clip]{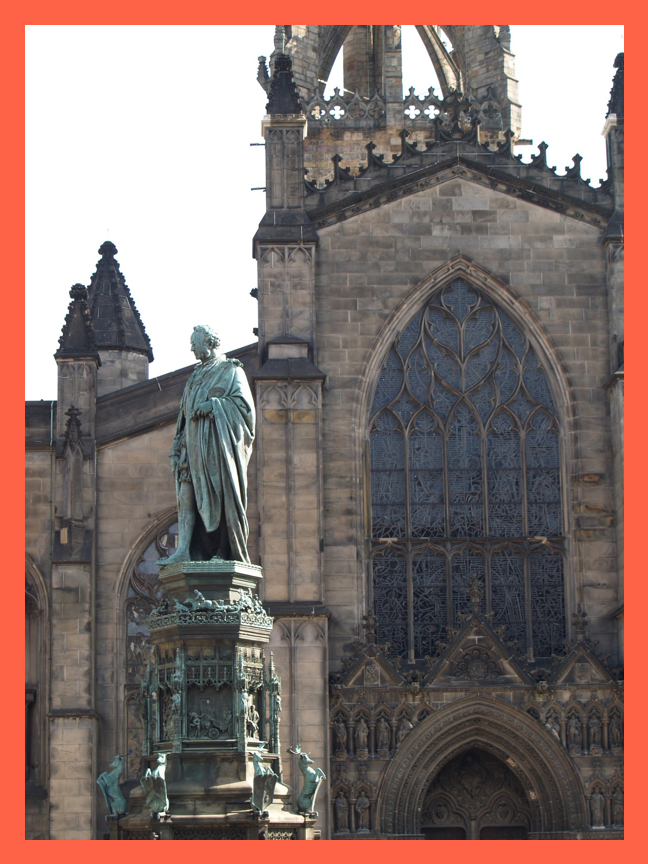}&
        \includegraphics[height=1.36cm, trim={0 0 0 0},clip]{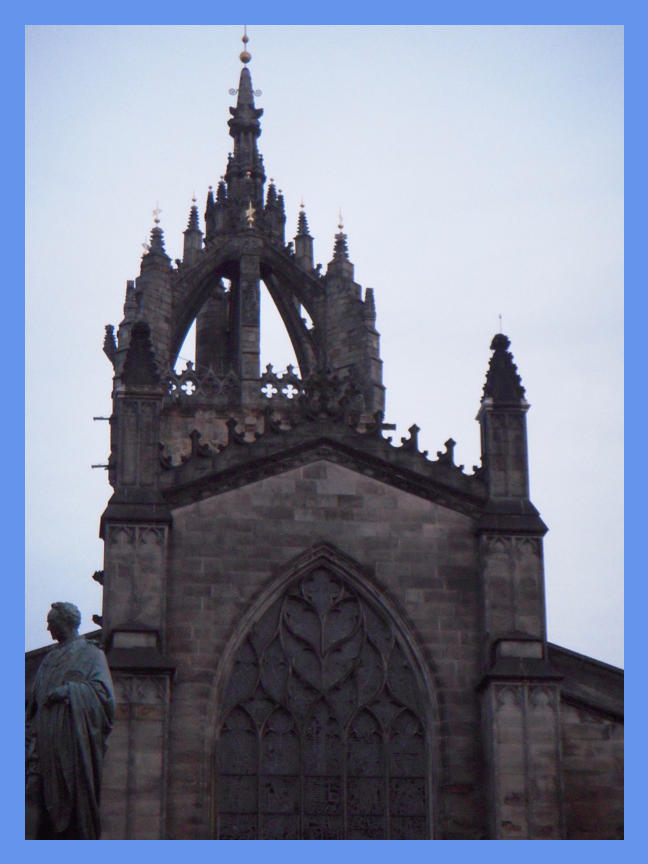} &
        \includegraphics[height=1.3cm, trim={0.2cm 0cm 0.4cm 0cm},clip]{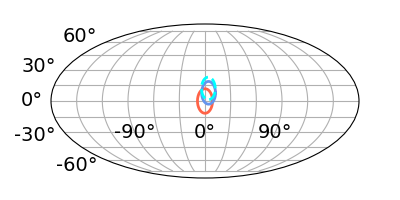} &
        \includegraphics[height=0.98cm, trim={0 0 0 0},clip]{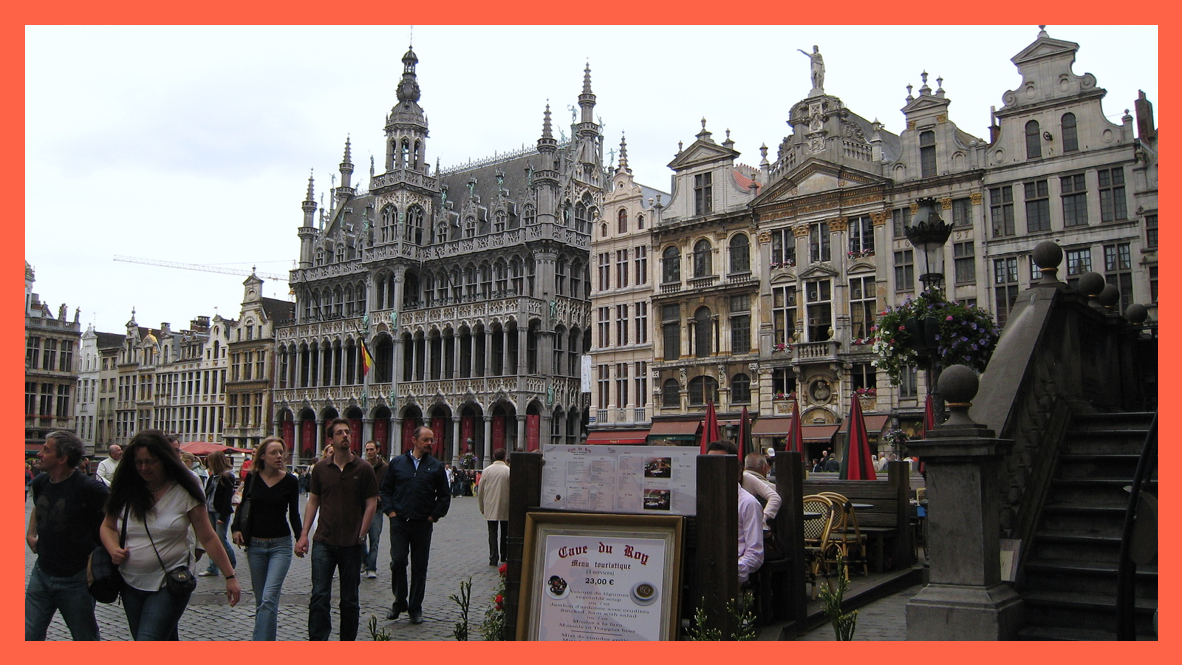}&
        \includegraphics[height=0.98cm, trim={0 0 0 0},clip]{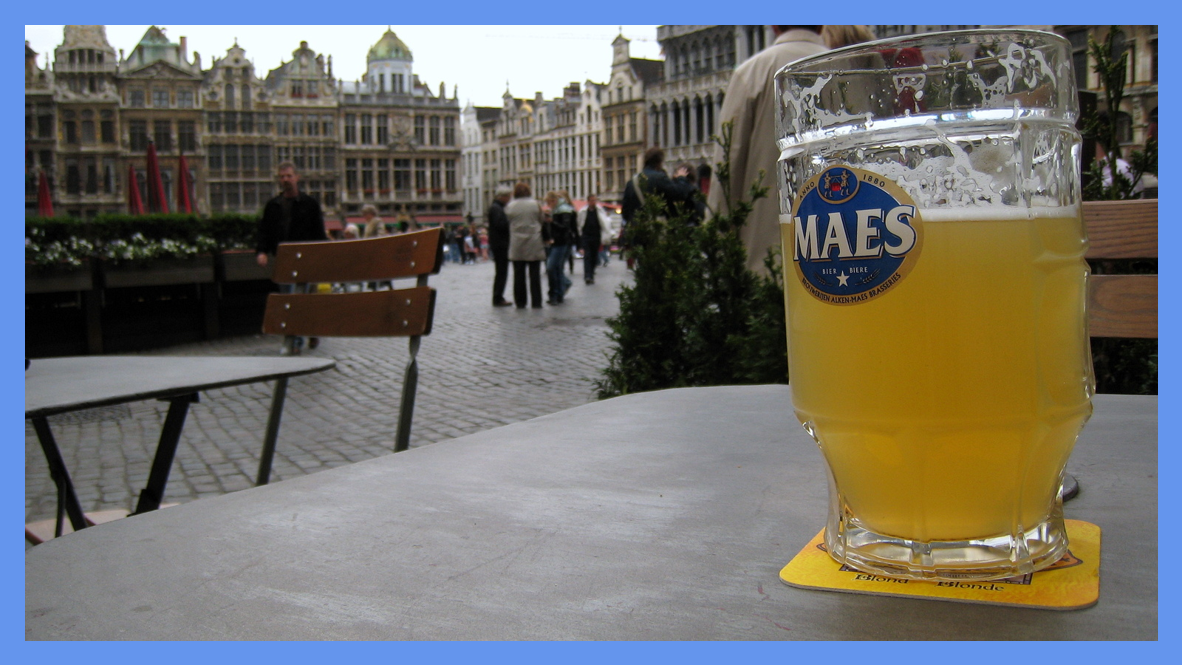} &
        \includegraphics[height=1.3cm, trim={0.2cm 0cm 0.4cm 0cm},clip]{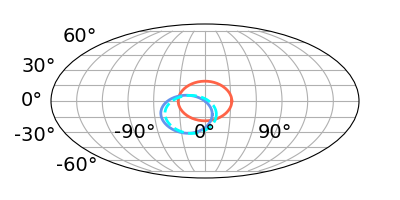} \\

        \rotatebox{90}{\makebox[37pt][c]{Small}} &

        \includegraphics[height=1.37cm, trim={0 0 0 0},clip]{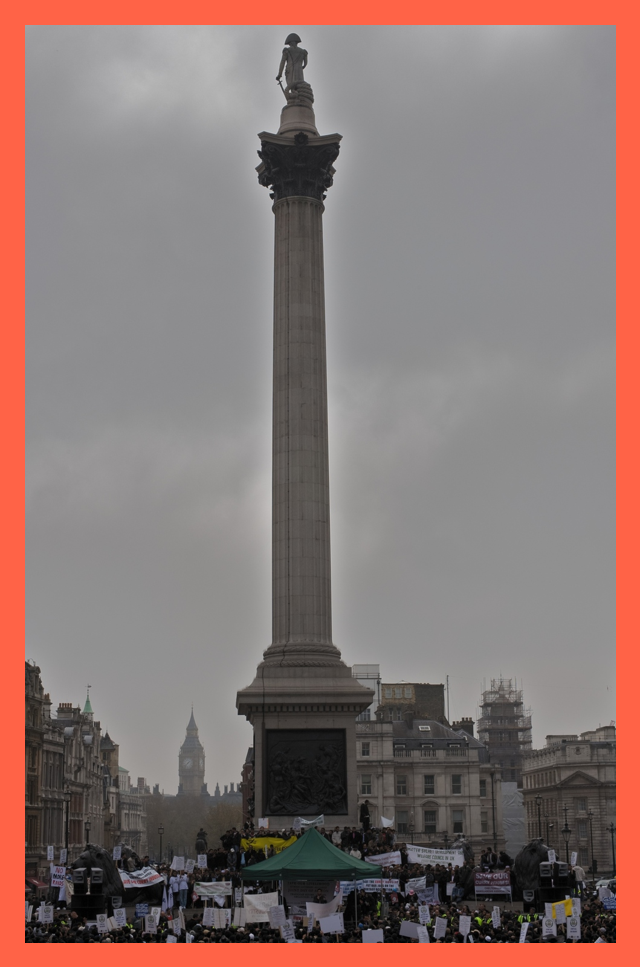}&
        \includegraphics[height=1.37cm, trim={0 0 0 0},clip]{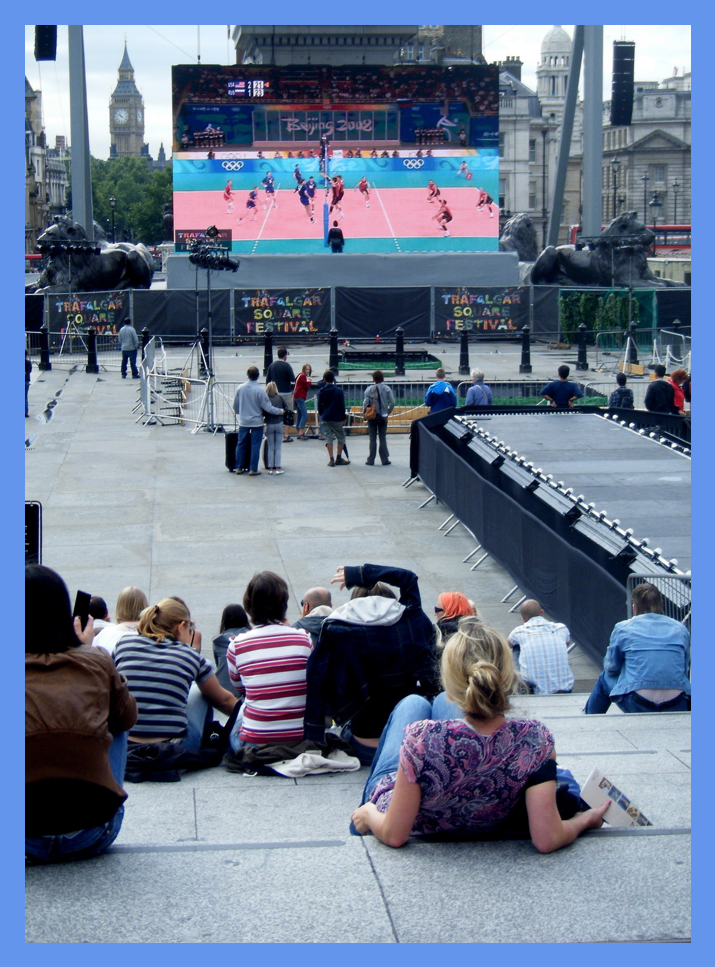} &
        \includegraphics[height=1.3cm, trim={0.2cm 0cm 0.4cm 0cm},clip]{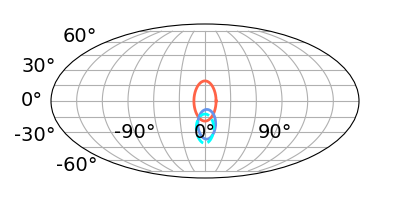} &
        \includegraphics[height=1.37cm, trim={0 0 0 0},clip]{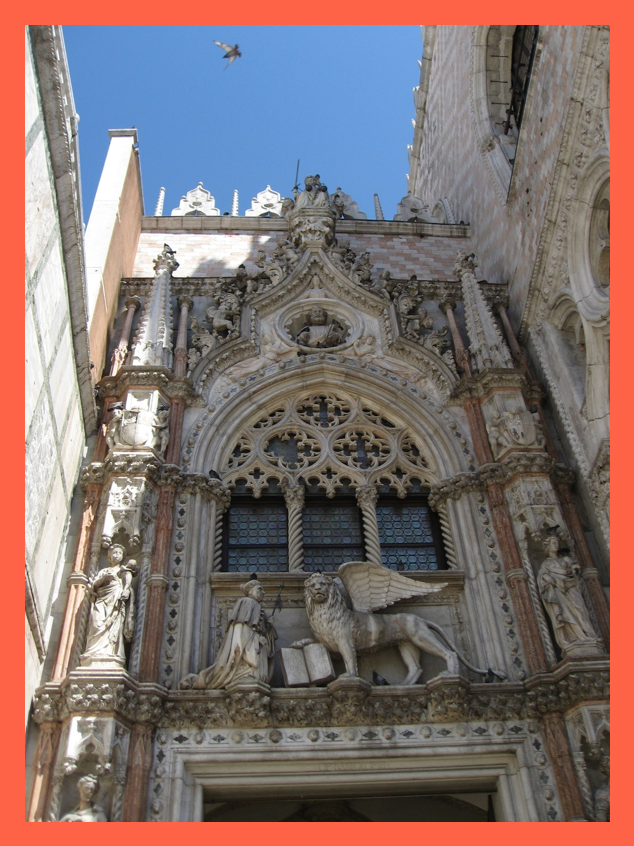}&
        \includegraphics[height=1.37cm, trim={0 0 0 0},clip]{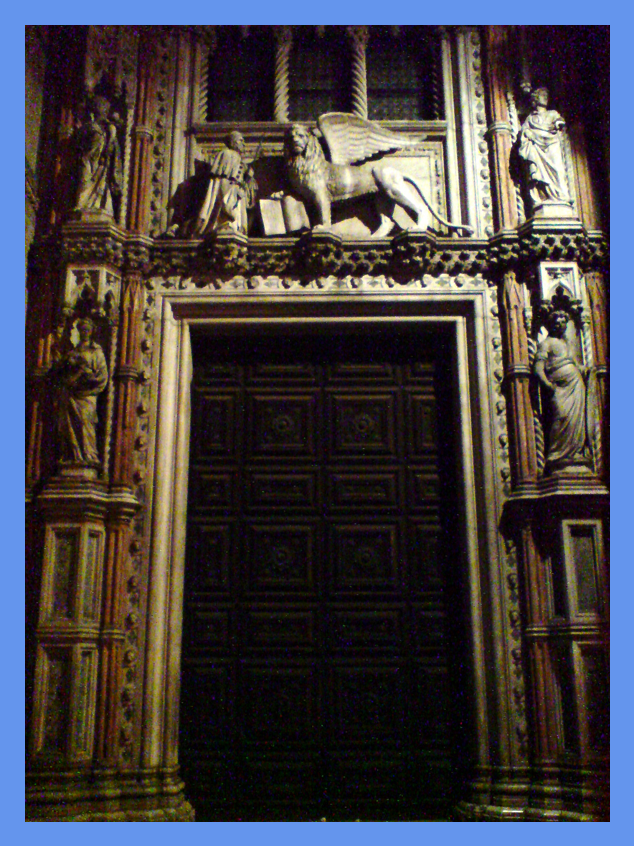} &
        \includegraphics[height=1.3cm, trim={0.2cm 0cm 0.4cm 0cm},clip]{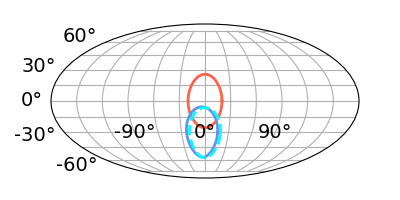} &
        \includegraphics[height=1.3cm, trim={0 0 0 0},clip]{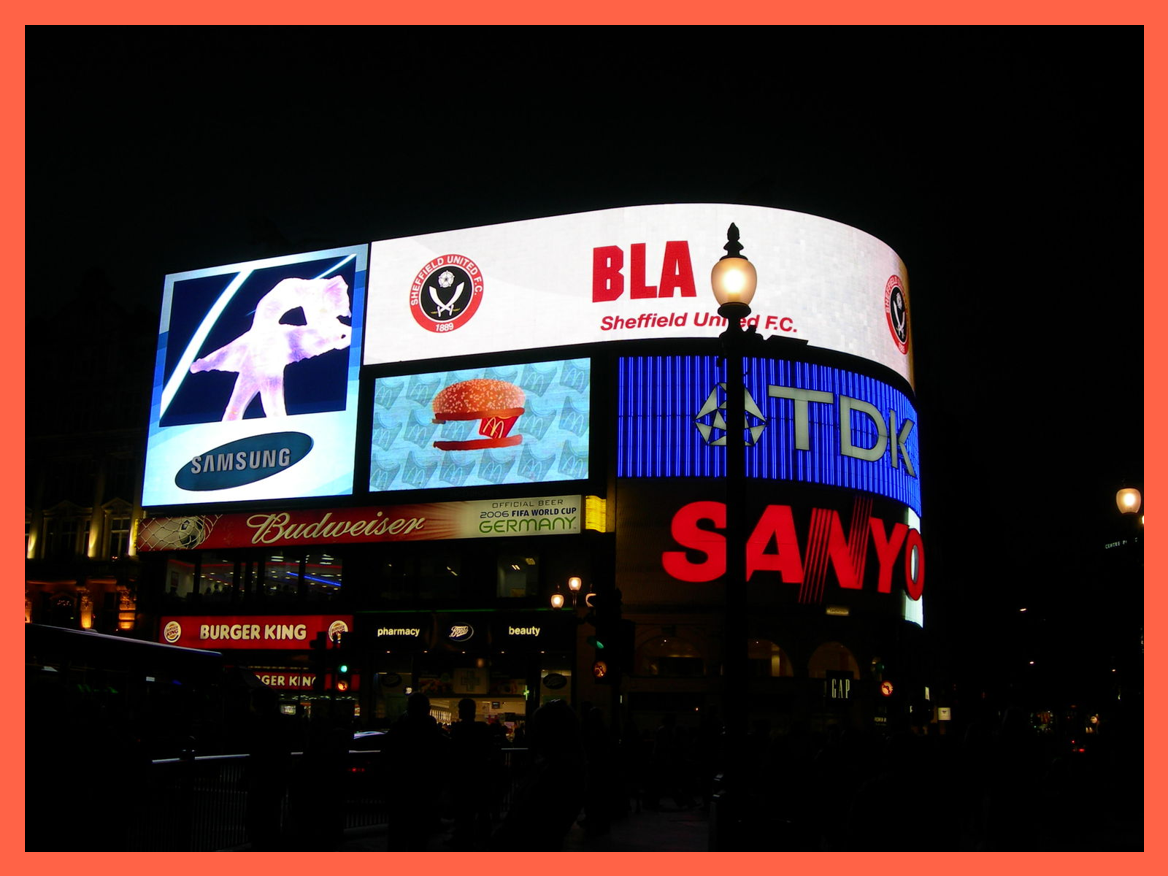}&
        \includegraphics[height=1.3cm, trim={0 0 0 0},clip]{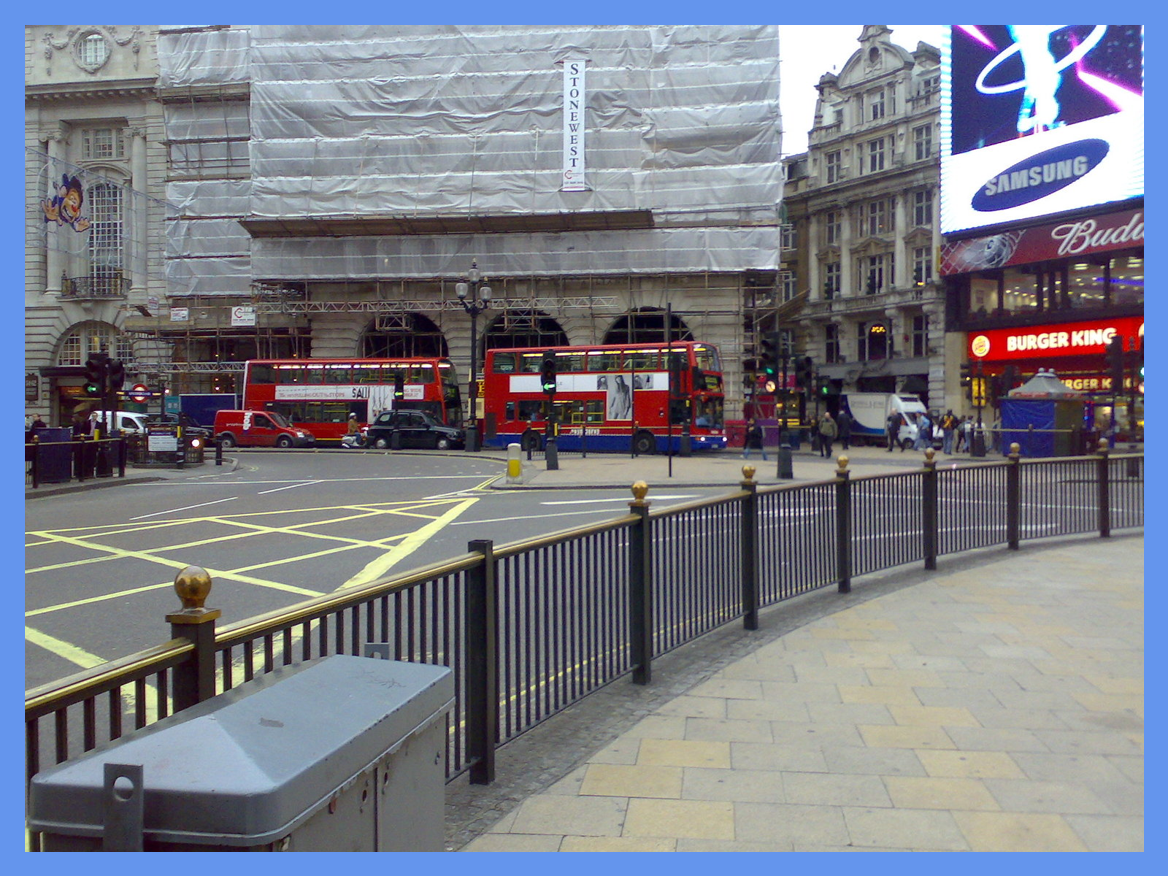} &
        \includegraphics[height=1.3cm, trim={0.2cm 0cm 0.4cm 0cm},clip]{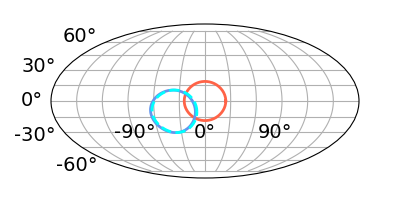} \\
        \rotatebox{90}{\makebox[37pt][c]{None}} &
 
        \includegraphics[height=1.37cm, trim={0 0 0 0},clip]{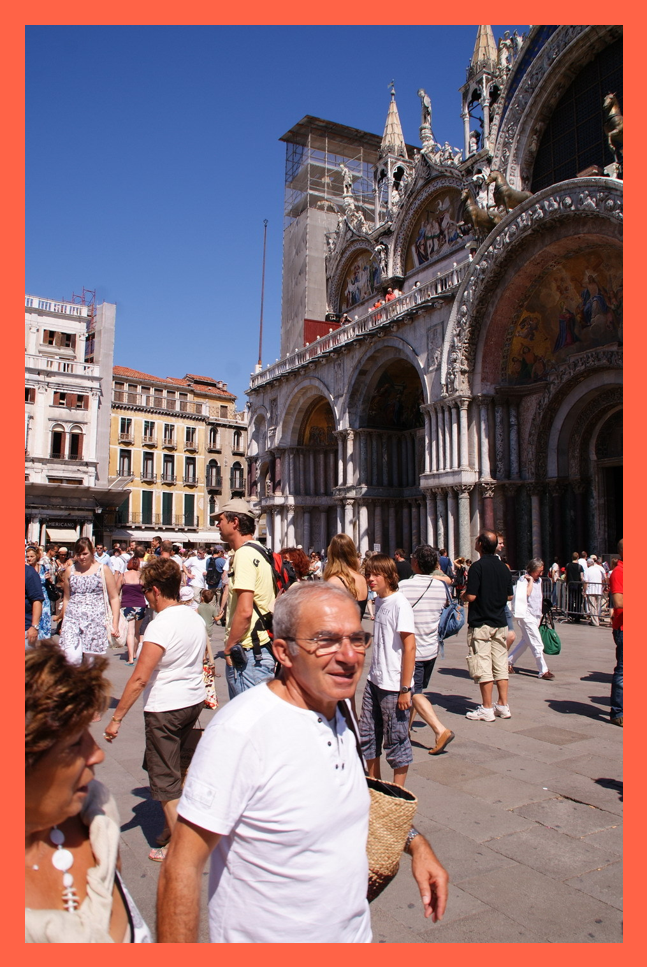}&
        \includegraphics[height=1.37cm, trim={0 0 0 0},clip]{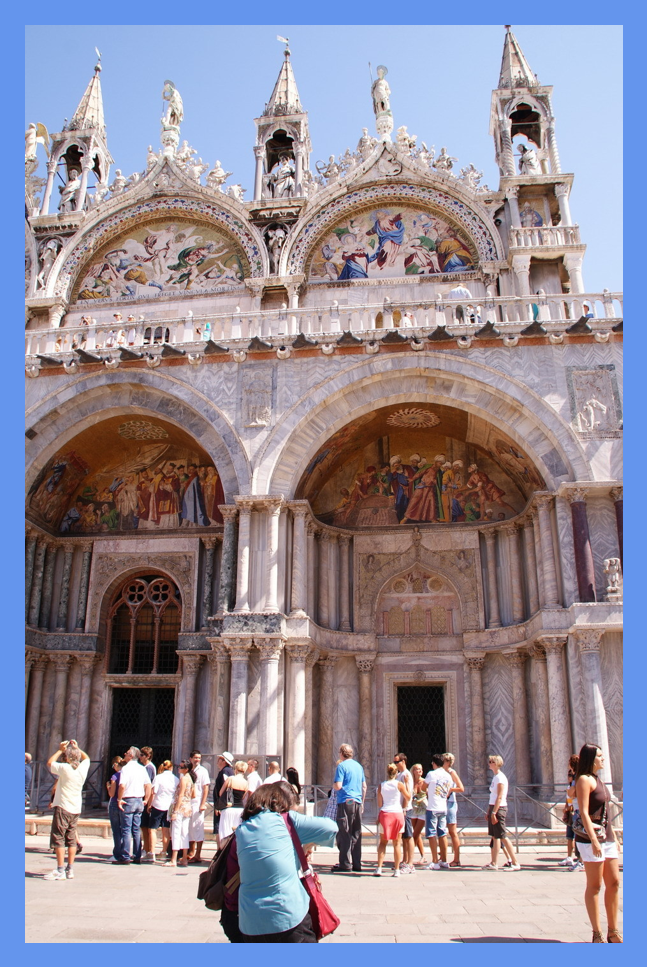} &
        \includegraphics[height=1.3cm, trim={0.2cm 0cm 0.4cm 0cm},clip]{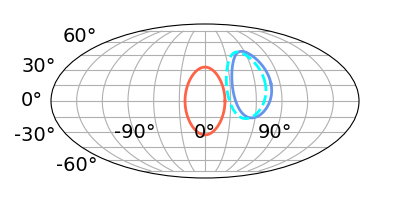} &
 
        \includegraphics[height=1.32cm, trim={0 0 0 0},clip]{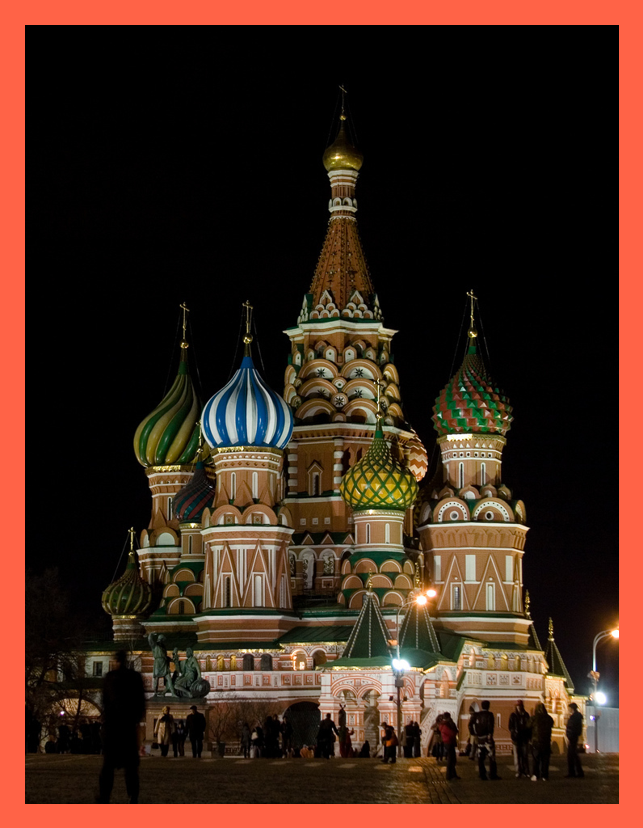}&
        \includegraphics[height=1.32cm, trim={0 0 0 0},clip]{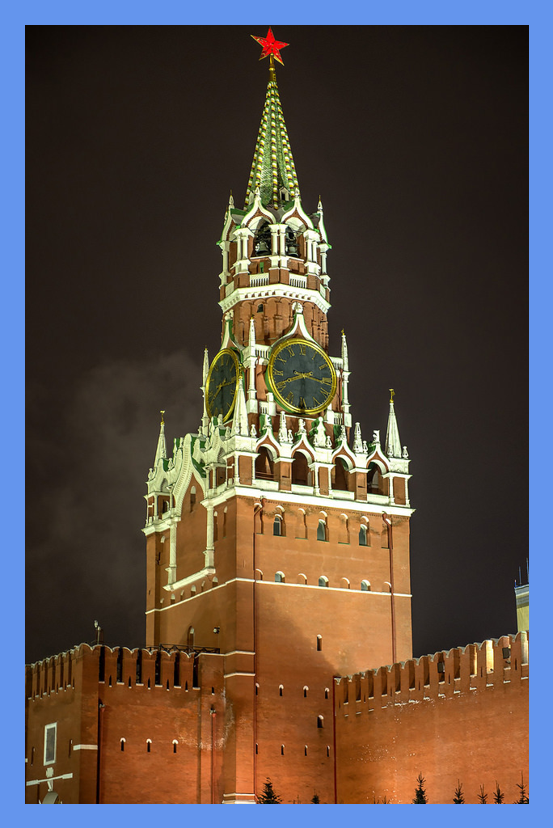} &
        \includegraphics[height=1.3cm, trim={0.2cm 0cm 0.4cm 0cm},clip]{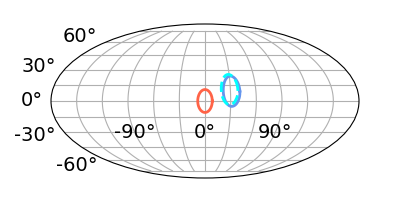} &
        \includegraphics[height=1.16cm, trim={0 0 0 0},clip]{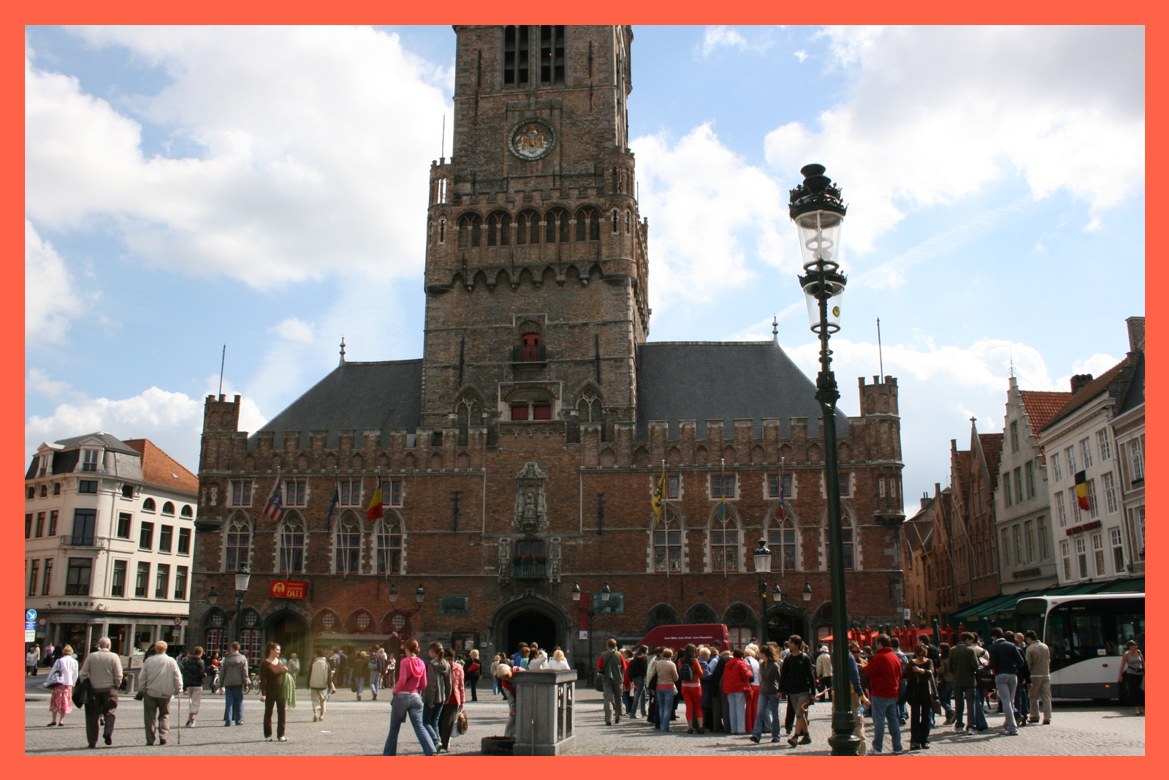}&
        \includegraphics[height=1.16cm, trim={0 0 0 0},clip]{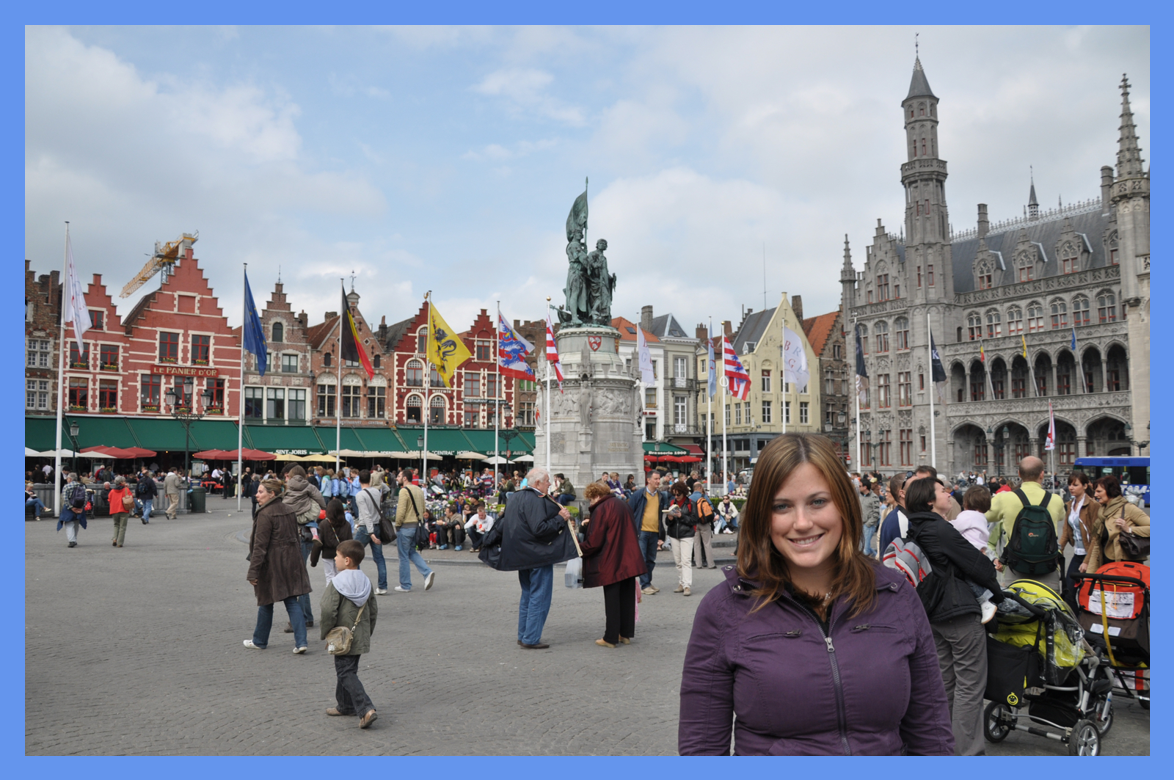} &
        \includegraphics[height=1.3cm, trim={0.2cm 0cm 0.4cm 0cm},clip]{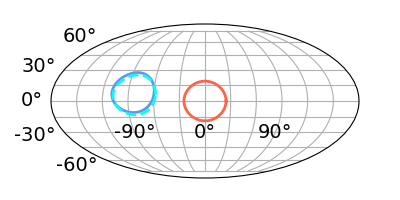} \\
    \end{tabular}
    \vspace{-8pt}
    \caption{\textbf{Qualitative results on the \megadepth{} test set}. We visualize the results of our model over different overlap levels, where the images on the \textcolor{RedOrange}{left} serve as the reference points, and their coordinate system determines the relative rotation, which defines the images on the \textcolor{RoyalBlue}{right}. The ellipsoids representing the ground truth are color-coded to match their respective images, with the estimated relative rotation illustrated by a cyan dashed line. As illustrated by the examples above, our method can accurately predict relative rotations for diverse image pairs containing varying appearances and intrinsic parameters. Please refer to the supplementary material for additional qualitative results.
    }
    \label{fig:Top 1}
\end{figure*}

In-the-wild images differ not only in their intrinsic parameters, but also in their appearance -- due to presence of varying illumination and dynamic objects. Therefore, in addition to augmenting the dataset with images of various field of views, we also perform image-level appearance augmentations, leveraging recent advancements in text-to-image Diffusion models~\cite{rombach2022high,hertz2022prompt,brooks2023instructpix2pix}. In particular, we apply the conditional InstructPix2Pix~\cite{brooks2023instructpix2pix} model on a subset of our data, using a set of text prompts of the format \emph{``Make it $\left<w\right>$''}, where $\left<w\right>$ captures diverse appearance modifications. See Figure \ref{fig:instruct_pair_before_and_after} for a few illustrative examples; note how some of these modify the global illumination (\emph{e.g.}, altering the time of day or the season) while others yield local edits (such as adding more people to make it \emph{``a busy street"}). We performed additional filtering on the augmented set, to avoid significant edits that also altered the structure of the scene; see the supplementary material for additional details.

\medskip \noindent \textbf{Training on Real Data.} Finally, we finetune the model over real image pairs from our proposed \shortdataset{} dataset. To prioritize learning over extreme image pairs, we first pass a batch of non-overlapping image pairs, followed by a batch of overlapping image pairs.

\section{Results and Evaluation}
\label{sec:results}

To validate our method, we compare our model with prior relative rotation estimation methods on the \cambridge{} and \megadepth{} test sets. In particular, our experiments seek to answer the following questions: 
\begin{itemize}
    \item How well do previous methods perform on our proposed task of extreme rotation estimation in the wild?
    \item How important is our progressive training scheme, and to what extent can the improved performance be attributed to training on the \dataset{} dataset?
    \item What is the impact of our design choices?
\end{itemize}
We also present qualitative results over different overlap levels in Figure~\ref{fig:Top 1}. These results illustrate our model's performance, and its robustness to varying illumination conditions and transient objects (such as the screen and the beer cup in the rightmost examples). Please refer to the supplementary material for many more qualitative results, including visualizations for baseline methods.


\subsection{Baselines} 
We compare our method to correspondence-based techniques, dedicated (end-to-end) relative rotation estimation techniques as well as recent relative pose estimation techniques. Specifically, we consider two correspondence-based methods: SIFT~\cite{lowe2004distinctive} and LoFTR~\cite{sun2021loftr}. These methods extract keypoint matches to compute an essential matrix using the RANSAC algorithm and calculate the rotation that stems from the essential matrix. 
We also compare against two prior works specifically targeting extreme rotation estimation using end-to-end deep learning frameworks: DenseCorrVol~\cite{cai2021extreme} and CascadedAtt~\cite{dekel2024estimating}. These methods are trained and evaluated on images sampled from panoramic views from StreetLearn~\cite{mirowski2019streetlearn}. Results for CascadedAtt are reported over a reproduced model, as a pretrained model is not provided; additional details are provided in the supplementary material.

We also consider 8PointViT~\cite{rockwell20228}, which estimates the relative pose between two images with a Vision Transformer that is modified to be close to the eight point algorithm. Following prior work~\cite{dekel2024estimating}, we only report results over overlapping settings for this baseline. 
Furthermore, as it assumes a single camera setting, we evaluate it only on the \cambridge{} test set. Finally, we consider Dust3R~\cite{wang2023dust3r}, a  recent technique for dense and unconstrained stereo 3D reconstruction. 
While this work does not specifically target the setting of extreme rotation estimation, we examine to what extent they can be adapted for this task in the wild. 
Note that Dust3R was trained on MegaDepth. Since our \megadepth{} test set is constructed from image pairs from this dataset, it is not entirely fair comparison; therefore, Dust3R's results on the \megadepth{} test set are highlighted in gray.

\subsection{Evaluation Metrics}
\label{sec:metrics}
For each image pair, we compute the geodesic error, defined as follows:
\begin{equation}
    \text{err} = \arccos \left( \frac{tr(\mathbf{R}^T\mathbf{R}^*)-1}{2}\right)
\end{equation}
where $\mathbf{R}$ is the predicted rotation matrix, and $\mathbf{R}^*$ is the ground truth relative rotation matrix. We report the median geodesic error (MGE) and relative rotation accuracy (RRA) for each test set and overlap ratio. RRA is reported over a predefined threshold $\tau$, indicating the percentage of image pairs with relative rotation error below $\tau$. We report this metric for $\tau = 15^\circ$ (RRA$_{15}$) and $\tau = 30^\circ$ (RRA$_{30}$).

Additionally, as we discretize the space of rotations and estimate three rotation angles, each predicted as a probability distribution over $N$ bins, we also report the performance of the \emph{Top 5} predictions in our ablation study. 
The \emph{Top 5} prediction reports the lowest error, considering the top-5 peaks in the relative yaw prediction only, as we observe that errors are mostly a function of the relative yaw angles.

\definecolor{graytext}{RGB}{130,130,130}
\begin{table}[t]
\setlength{\tabcolsep}{1.7pt}
 \def\arraystretch{1}
\centering
\resizebox{0.45\textwidth}{!}{%
\begin{tabular}{llcccccccccc} 

    \toprule
    \multicolumn{2}{c}{}&&\multicolumn{3}{c}{\cambridge{}}&&
    \multicolumn{3}{c}{\megadepth{}}\\
    \cline{4-6} \cline{8-10}
      & Method &&MGE\textdownarrow & RRA$_{15}$\textuparrow &RRA$_{30}$\textuparrow && MGE\textdownarrow & RRA$_{15}$ \textuparrow & RRA$_{30}$\textuparrow \\
    \midrule

    \multirow{6}{*}{\rotatebox[origin=c]{90}{Large}} & 
    SIFT*~\cite{lowe2004distinctive} && 1.95  & 92.3 & 95.3 && 2.94  & 74.6 & 80.8 \\
    & LoFTR*~\cite{sun2021loftr} &&   1.76 & 97.2 & 99.1 &&  \textbf{2.13}  & 85.2 & 93.8 \\
    & DenseCorrVol~\cite{cai2021extreme} &&   98.51 & 25.9 & 33.3 &&   120.53 & 7.0 & 13.0 \\   &CascadedAtt~\cite{dekel2024estimating}&&   29.75 & 42.7 & 50.0 && 170.62 & 7.3 & 9.2\\
        & 8PointViT~\cite{rockwell20228} &&   22.33 & 31.9 & 64.8 && -- & -- & -- \\
    & 
    Dust3R~\cite{wang2023dust3r} &&  \textbf{0.77} & \textbf{99.7} & \textbf{99.9} &&    \colorbox{lightgray}{1.01} &  \colorbox{lightgray}{98.4} &  \colorbox{lightgray}{99.2} \\

    & Ours&&   2.45 & 96.7 & 96.8 &&   {2.41}  & \textbf{97.5}  & \textbf{97.9}  \\
    \midrule
    \multirow{6}{*}{\rotatebox[origin=c]{90}{Small}} & 
    SIFT*~\cite{lowe2004distinctive} && 5.07  & 64.7 & 71.3 && \textcolor{gray}{7.27}  & \textcolor{gray}{61.4} & \textcolor{gray}{68.4} \\
    & LoFTR$^*$~\cite{sun2021loftr} &&   \textcolor{gray}{2.70} & \textcolor{gray}{81.5} & \textcolor{gray}{93.6} &&   \textcolor{gray}{6.80} & \textcolor{gray}{66.6} & \textcolor{gray}{81.2 }\\
    & DenseCorrVol~\cite{cai2021extreme} &&   143.47 & 2.8 & 9.4 &&   125.73 & 3.1 & 9.4 \\
   &CascadedAtt~\cite{dekel2024estimating}&&   148.44 & 0.0 & 3.0 && 139.14 & 2.7 & 4.4\\
        & 8PointViT~\cite{rockwell20228}  &&   51.30 & 1.7 & 12.8 && -- & -- & --\\
    & Dust3R~\cite{wang2023dust3r} &&   \textbf{1.96} & \textbf{95.9} & \textbf{94.6} && \colorbox{lightgray}{{2.80}} & \colorbox{lightgray}{{89.8}} & \colorbox{lightgray}{{94.4}}\\
    & Ours &&   4.35 & 88.3 & 89.0 &&  \textbf{4.47} &  \textbf{87.2} & \textbf{91.6} \\
    \midrule
    \multirow{6}{*}{\rotatebox[origin=c]{90}{None}} & 
    SIFT*~\cite{lowe2004distinctive} && \textcolor{gray}{121.94}  & \textcolor{gray}{2.7} & \textcolor{gray}{5.4} && \textcolor{gray}{122.84}  & \textcolor{gray}{0.0} & \textcolor{gray}{2.0} \\
    & LoFTR*~\cite{sun2021loftr} &&  \textcolor{gray}{ - }& \textcolor{gray}{ - }& \textcolor{gray}{- }&& \textcolor{gray}{ 56.54 }& \textcolor{gray}{0.0 }& \textcolor{gray}{33.0} \\
    & DenseCorrVol~\cite{cai2021extreme} && 77.10 & 9.0 & 27.0 &&   82.04 & 2.9 & 13.7 \\
    &CascadedAtt~\cite{dekel2024estimating}&&   69.69 & 8.4 & 23.1 && 78.60 & 7.5 & 20.8 \\
    & Dust3R~\cite{wang2023dust3r} &&   114.33 & 19.8 & 23.9 && \colorbox{lightgray}{{81.21}} & \colorbox{lightgray}{{15.4}} & \colorbox{lightgray}{{26.9}}\\
    & Ours &&   \textbf{13.62} & \textbf{52.7} & \textbf{59.7} &&   \textbf{26.97} & \textbf{36.1} & \textbf{50.7} \\
    \bottomrule
\end{tabular}
}
\vspace{-8pt}
\caption{\textbf{Rotation Estimation in the Wild.} We evaluate performance over the \cambridge{} and \megadepth{} test sets, separately considering Large (top), Small (middle) and Non-overlapping (bottom) pairs. 
$^*$ indicates median errors are computed only over successful image pairs, for which these algorithms output a pose estimate (failure over more than $50\%$ of the test pairs is shown in gray). 
\colorbox{lightgray}{Note that Dust3R was trained on images from \megadepth{}.}
}

\label{tab:main_result_val}
\end{table}

\subsection{Quantitative Comparison}

\noindent \textbf{Evaluation on the \shortdataset{} Test Sets}.
The main quantitative results comparing our method to other methods on rotation estimation in the wild are reported in Table~\ref{tab:main_result_val}.

As illustrated in Table~\ref{tab:main_result_val}, 
SIFT and LoFTR, which are correspondence-based methods, exhibit some robustness when handling highly overlapping Internet images, compared to methods that trained on images cropped from panoramas, achieving a median error of less than 3$^\circ$ for Large overlap pairs in both test sets. 
However, these methods rely on image overlap and may not always provide an output of estimated camera pose, as the geometric verification requires a sufficient number of detected inliers. Therefore, they struggle to produce reliable matches in cases of limited overlapping regions, as also observed in prior work focusing on extreme scenarios~\cite{cai2021extreme,dekel2024estimating}.

DenseCorrVol, CascadedAtt and 8PointViT exhibit relatively poor performance on the \shortdataset{} test sets, illustrating that models trained on images sampled from panoramas cannot easily generalize to Internet photos.
In the supplementary material, 
we show that our model significantly outperforms these prior work even when trained on the same dataset (\emph{i.e.}, images with a constant FoV and illumination cropped from panoramas).
As further discussed there, we believe this gap can be partially attributed to the usage of a pretrained LoFTR feature extractor, which is capable of encoding knowledge between Internet image pairs (which vary in their intrinsics and appearance). 

Dust3R operates on Internet datasets without the need for calibration and performs exceptionally well on overlapping pairs, achieving the lowest median error and highest success rates for Large and Small overlap categories for \cambridge{} test set. However, this method is designed for overlapping pairs as it initializes the model using pretrained CroCo~\cite{weinzaepfel2022croco}, which is trained to address cross-view completion problem from two overlapping views.
Furthermore, its output consists of a unified dense point cloud for each pair of images.  
Due to its design for overlapping pairs, its performance on non-overlapping views is extremely low, also on the \megadepth{} test set, which contains images from scenes included in its training set. 
Moreover, in terms of size, Dust3R (with a DPT head) contains 577 million parameters. In contrast, our model is significantly more compact, with only 80 million parameters. As shown in the table, despite its smaller size, our model yields significantly improved performance over non-overlapping pairs on both test sets.

\definecolor{graytext}{RGB}{130,130,130}
\begin{table}[t]
\setlength{\tabcolsep}{1.7pt}
 \def\arraystretch{1.02}
\centering
\resizebox{0.45\textwidth}{!}{%
\begin{tabular}{lccccccccccccccccc} 
    \toprule
    && \multicolumn{2}{c}{Large} && \multicolumn{2}{c}{Small} && \multicolumn{2}{c}{None} \\
     \cline{3-4} \cline{6-7} \cline{9-10}
     Method &&MGE \textdownarrow& RRA$_{10}$ \textuparrow&&MGE \textdownarrow& RRA$_{10}$\textuparrow &&MGE\textdownarrow & RRA$_{10}$ \textuparrow\\
    \midrule
DenseCorrVol~\cite{cai2021extreme} &&  1.09 & 99.4 &&  1.41 & 98.3 && \textbf{1.53} &  \textbf{96.4}\\
CascadedAtt~\cite{dekel2024estimating} && 1.42 & \textbf{100.0} && 1.89 & 98.3 && 2.06 & 96.2\\
8PointViT~\cite{rockwell20228}
&&  \textbf{0.52} &  \textbf{100.0} &&  \textbf{1.07} &  \textbf{100.0} && 101.51 & 6.0\\
Ours &&  {1.06} &  \textbf{100.0} &&  {1.09} &  \textbf{100.0} && 1.98 &  \textbf{96.4}\\
     \bottomrule
\end{tabular}
}
\vspace{-8pt}
\caption{\textbf{Evaluation on Images Cropped from Panoramas}. We compare results over the Streetlearn panoramas test set, first reported in Cai \emph{et al.}~\cite{cai2021extreme}. Note that all models (including ours) were trained on the same data; for this comparison, we report the performance of the model obtained after the \emph{initialization} stage.
}
\label{tab:network-ablation}
\end{table}

\medskip \noindent \textbf{Evaluation on Images Cropped from Panoramas.} In Table \ref{tab:network-ablation}, we conduct an evaluation over panoramic perspective images, using the training and test set reported in prior work~\cite{cai2021extreme,dekel2024estimating}. Note that all models are trained and evaluated on StreetLearn~\cite{mirowski2019streetlearn} images; no data augmentations or additional data sources are used for this evaluation. As illustrated in the table, our model yields comparable performance over such constrained image pairs, achieving state-of-the-art relative rotation accuracy for non-overlapping pairs, matching the performance reported in prior work.

\subsection{Ablation study}
\label{sec:ablation}
We conduct various ablations analyzing the effect of our progressive training scheme and other design choices,also reporting performance over the \emph{Top 5} predictions (considering the top-5 peaks in the relative yaw prediction, as detailed in Section \ref{sec:metrics}). Additional ablations, including an analysis of architectural components, are provided in the supplementary material.

\definecolor{graytext}{RGB}{130,130,130}
\begin{table}[t]
\setlength{\tabcolsep}{1.7pt}
 \def\arraystretch{1.02}
\centering
\resizebox{0.45\textwidth}{!}{%
\begin{tabular}{lllccccccccccccccccc} 
    \toprule
    & \multicolumn{3}{c}{}&\multicolumn{3}{c}{Top 1} & & \multicolumn{3}{c}{Top 5} \\
     \cline{5-7} \cline{9-11}
     &Overlap & Train data &&MGE \textdownarrow  & RRA$_{15}$ \textuparrow&RRA$_{30}$\textuparrow && MGE \textdownarrow & RRA$_{15}$\textuparrow & RRA$_{30}$\textuparrow \\
    \midrule
      & Large & \cite{cai2021extreme} && 13.65 & 35.4 & 73.5 && 12.22 & 61.4 & 84.7\\
    & & +\fovexp{} && 4.61 & 79.7 & 81.1 && 4.41 & 90.3& 98.9\\
    & & +\instructexp{} && 4.46 & 90.4 & 92.4 && 4.43 & 94.3 & 99.1 \\
    & & +\shortdataset{} && \textbf{2.41}  & \textbf{97.5}  & \textbf{97.9}  && \textbf{2.41}  &  \textbf{98.4}  & \textbf{99.4} \\
    &Small& \cite{cai2021extreme} && 55.28 & 3.7 & 29.1 && 29.83 & 15.0 & 50.3  \\
    & & +\fovexp{} && 12.91 & 56.2 & 68.2 && 10.97 & 66.0 & 85.4 \\
    & & +\instructexp{} && 11.46 & 62.5 & 80.6 && 10.73 & 68.0 & 91.0 \\
     & & +\shortdataset{} &&  \textbf{4.47} &  \textbf{87.2} & \textbf{91.6} && \textbf{4.24} &  \textbf{91.1} &  \textbf{97.2}  \\
    &None & \cite{cai2021extreme} && 74.94 & 12.8 & 25.3 && 25.11 & 26.1 & 58.8  \\
    & & +\fovexp{} && 61.62 & 25.0 & 38.4 && 16.82 & 44.2 & 75.0 \\
    & & +\instructexp{} && 68.31 & 25.0 & 36.1 && 16.21 & 45.7 & 78.2 \\
    & & +\shortdataset{} &&   \textbf{26.97} &  \textbf{36.1} &  \textbf{50.7} &&  \textbf{12.85} & \textbf{57.1} &  \textbf{85.8}  \\
    \bottomrule
\end{tabular}
}
\vspace{-8pt}
\caption{Ablation study, evaluating the effect of our progressive training scheme over the \megadepth{} test set. All experiments start with the cropped panoramas used in Cai \emph{et al.}~\cite{cai2021extreme}. 
}
\label{tab:ablation_val}
\end{table}

\medskip \noindent
\textbf{The effect of our progressive training scheme.} We conduct multiple experiments ablating the effect of our progressive training scheme. Our training process consists of four stages: initialization (following prior work~\cite{cai2021extreme}),  
incorporating multiple FoVs (+\fovexp{}), training with image-level appearance augmentations (+\instructexp{}), and training with \dataset{} pairs (+\shortdataset{}). 

Table \ref{tab:ablation_val} illustrates the impact of  each training stage on the results. As can be observed from the table, each training stage further refines the model's performance, often in a significant manner. For instance, 
the median  error in the \megadepth{} small overlap test set decreased from 55.3$^\circ$ to 12.9$^\circ$ when FoV augmentations were added. Additionally, the \dataset{} training set plays a crucial rule in finalizing our training process, yielding a significant reduction in the median error (more than half) for small and non-overlapping scenarios in the \megadepth{} test set.
While this median error remains relatively high for non-overlapping pairs, we observe that the top-5 scores show significant improvements, \emph{e.g.}, reducing the median error
from 27.0$^\circ$ to 12.8$^\circ$. 
This demonstrates that the model has learned this knowledge, although it cannot necessarily be recovered by the largest peak alone. In the supplementary material, we demonstrate that the intermediate training stages are indeed important and that the improved performance cannot be obtained with the \dataset{} training set alone.

Moreover, evidenced in Table \ref{tab:ablation_val}, our model demonstrates strong generalization capabilities even when trained exclusively on panorama-cropped images. The model generalizes to the  \megadepth{} test set significantly better (e.g., for Large overlap cases, our model achieves an MGE of 13.65$^\circ$ compared to values higher than 120$^\circ$ achieved by baselines), demonstrating that the improvement is not just due to our progressive training scheme.
We further examined architectural differences by conducting an additional ablation study, applying our progressive training scheme to the baseline models (see Table~\ref{tab:dcv_w_data} in the supplementary materials). This experiment revealed that prior models are not directly suitable for real-world applications, as these baseline models showed significantly poorer performance across all metrics.

\definecolor{graytext}{RGB}{130,130,130}
\begin{table}[t]
\setlength{\tabcolsep}{1.7pt}
 \def\arraystretch{1.02}
\centering
\resizebox{0.45\textwidth}{!}{%
\begin{tabular}{lllccccccccccccccccc} 
    \toprule
    & & \multicolumn{3}{c}{}&\multicolumn{3}{c}{Top 1} & & \multicolumn{3}{c}{Top 5} \\
     \cline{6-8} \cline{10-12}
     &Overlap & KP & SM &&MGE \textdownarrow& RRA$_{15}$\textuparrow &RRA$_{30}$\textuparrow && MGE \textdownarrow& RRA$_{15}$\textuparrow & RRA$_{30}$\textuparrow \\
    \midrule
    & {Large}  & $\times$ & $\times$ && \textbf{2.18}  & 97.4  & 98.1  && \textbf{2.18}  & 97.4  & 98.1  \\ 
    && $\times$ & \checkmark && 2.30  & 97.0  & 97.4  && 2.30  & \textbf{98.5}  & 99.4  \\
    && \checkmark & $\times$ &&  2.44 & \textbf{97.6}  &  \textbf{98.3}
    &&  2.31  & 98.4  & \textbf{99.7}  \\
    & & \checkmark & \checkmark && 2.41  & 97.5  & {97.9}  && 2.41  &  98.4  & 99.4 \\
    &{Small} & $\times$ & $\times$ && 4.50 & 87.9 & 91.6 && 4.50  & 87.9  & 91.7  \\ 
    && $\times$ & \checkmark && 4.49  & \textbf{88.1}  & 92.0  && 4.46 & 91.2  & 96.7  \\
    & & \checkmark & $\times$ && \textbf{4.41} & 87.5  & \textbf{92.2}   && 4.32  & \textbf{91.9}  & \textbf{97.6} \\
    & & \checkmark & \checkmark &&  4.47 &  87.2 & {91.6} &&  \textbf{4.24} &  91.1 &  97.2  \\
    &{None} 
    & $\times$ & $\times$ && 48.81  & 34.0  & 44.1  && \textbf{12.56}  & \textbf{57.5}  &  84.6  \\ 
    
    && $\times$ & \checkmark && 43.07  & 31.2  & 44.2  && 13.99  & 53.5  & 83.2  \\ 
    && \checkmark & $\times$ && 41.39 & 35.3  & 46.8  
    && 13.04  & 56.9  & \textbf{86.2}  \\
    & & \checkmark & \checkmark &&   \textbf{26.97} &  \textbf{36.1} &  \textbf{50.7} &&  12.85 & 57.1 &  85.8  \\
    \bottomrule
\end{tabular}
}
\vspace{-8pt}
\caption{Ablation study, evaluating the effect of the auxiliary channels added as input to our network. We train models without the keypoints and matches (KP) and without the segmentation maps (SM), comparing to our model over the \megadepth{} test set , after using a validation split that is balanced over the overall angle.} 
\label{tab:auxillary_val_balanced}
\end{table}

\medskip \noindent
\textbf{The effect of adding auxiliary channels.} We ablate the effect of adding auxiliary in Table \ref{tab:auxillary_val_balanced}, training models without keypoints and matches (KP) or the segmentation map (SM) provided as additional inputs. As illustrated in the table, these auxiliary channels boost performance almost across all metrics. In particular, both channels play a role in reducing the errors over non-overlapping image pairs (reducing the median error from over $40^\circ$ to $27.0^\circ$). 
\section{Conclusion}
We present a method and benchmark dataset for estimating relative 3D rotations between pairs of (possibly) non-overlapping RGB images. Our approach extends prior work addressing extreme rotations to real-world data that exhibit variation in both appearance in intrinsic camera parameters. 
While our model shows promising results on real-world Internet image pairs, it also highlights the inherent difficulty of the underlying task, suggesting that considerable progress can be achieved with future techniques that leverage our dataset. Our paired data could also potentially serve for exploring the challenging task of estimating extreme translations in real-world settings. Future work could also consider incorporating more views for enhancing performance in such extreme non-overlapping scenarios.

{
    \small
    \bibliographystyle{ieeenat_fullname}
    \bibliography{main}
}

  \clearpage
\setcounter{page}{1}
 \maketitlesupplementary


\medskip
\medskip
We refer readers to the interactive visualizations at our \href{https://tau-vailab.github.io/ExtremeRotationsInTheWild/}{project page} that show results for all presented models on the \shortdataset{} test sets. In this document, we provide  details regarding our proposed dataset (Section \ref{sec:dataset-supp}), additional implementation details (Section \ref{sec:details-supp}) and 
describe additional experiments and results (Section \ref{sec:results-supp}).

\section{Construction of the \shortdataset{} Dataset}
\label{sec:dataset-supp}
As mentioned in the main paper, we leveraged scene-level Internet photo collections for constructing the \shortdataset{} dataset, focusing on pairs with predominant rotational motion. Given scale is a degree of freedom in SfM reconstruction algorithms, we established a scene-based translation threshold as described below. We construct \emph{mutual} nearest neighbors edge-weighted graphs, with one graph per landmark. In each graph $G$, nodes $v \in V$ correspond to images, and two images are connected by an edge $e \in E$ if they are both among each other's $K$ nearest neighbors ($K$ is empirically set to $5$). The weights $w$ of the edges in each graph $G$ are set according to the L2 distances between the translation of the images. For each landmark, we compute the weighted node degree $d_v$ for each node, defined as the sum of the edge weights of edges incident to that node, divided by the number of such edges. 

For example, for an image node $v_{i}$ with translation $T_{i}$ and its mutual nearest neighbors represented by nodes $v_{j}$, where $j\in [0,...,k]$ and $k\leq K$, the weight of the edge $e_{i,j}$ is calculated as $w_{i,j}=||T_{i}-T_{j}||_2$.
The weighted node degree $d_{v_i}$ is then computed as $d_{v_i}=\frac{1}{k}{\sum_{j=0}^{k}}{w_{i,j}}$.
Finally, for the entire scene graph, we select image pairs with edge weights $w$ below the median value of weighted node degrees, specifically where $w<\text{med}\{d_v\}$. 

  The images for \shortdataset{} training set were curated from MegaScenes~\cite{tung2024megascenes} that utilizes COLMAP~\cite{schoenberger2016sfm} for its ground truth poses and uses a Manhattan world alignment. Our benchmark contains two test set, as follows:
 
   \textbf{\cambridge{}}. Image pairs in the \cambridge{} test set contain images from the Cambridge Landmarks~\cite{kendall2015cambridgelandmarks} dataset. The ground truth poses of Cambridge Landmarks Dataset~\cite{kendall2015cambridgelandmarks} are based on VisualSFM~\cite{wu2013visualsfm}. 
The ground truth poses were rotated so coordinate system would align with the gravity and horizontal axis using  \cite{jin2023percpectivefields}.

   \textbf{\megadepth{}}. Image pairs in the \megadepth{} test set contain images from the MegaDepth~\cite{li2018megadepth} dataset. MegaDepth also utilizes COLMAP~\cite{schoenberger2016sfm} for its ground truth poses and use Manhattan world alignment. 

   Both test sets underwent a filtering process to remove any images where transient objects occupied over 40\% of the image area. This selection was made using a SegFormer~\cite{enze2021segformer} segmentation mask, targeting specific transient object categories. Specifically, we consider: \emph{person}, \emph{car}, \emph{bus}, \emph{bicycle}, \emph{boat}, \emph{truck}, \emph{airplane}, \emph{van}, \emph{ship}, \emph{minibike}, and \emph{animal}.

\begin{table}[!t]
\centering
\resizebox{\linewidth}{!}{
\begin{tabular}{lcccccccc}
\toprule

Scene name & Scene number & Large &  Small & None & All\\ 
\midrule
\\
\textbf{\cambridge{}}\\
\\

Total & - & 2512 & 827 & 1961 & 5300 \\
GreatCourt & - & 548 & 163 & 248 & 959 \\
KingsCollege & - & 409 & 12 & 0 & 421 \\
StMarysChurch & - & 405 & 233 & 35 & 673 \\
OldHospital & - & 395 & 70 & 0 & 465 \\
Street & - & 494 & 342 & 1678 & 2514 \\
ShopFacade & - & 261 & 7 & 0 & 268 \\
\\\hline\\
\textbf{\megadepth{}}\\
\\
Total & - & 2700 & 829 & 643 & 4172 \\
Trafalgar Square, London & 1 & 566 & 207 & 157 & 930 \\ 
San Marco,Venice & 8 & 226 & 121 & 138 & 485 \\ 
Piazza del Popolo, Rome & 17 & 350 & 34 & 5 & 389 \\
Vatican, Rome & 15 & 206 & 101 & 68 & 375 \\
Piazza del Campo, Firenze & 115 & 186 & 111 & 37 & 334 \\
Red Square, Moscow & 559 & 231 & 64 & 26 & 321 \\
Piccadilly Circus, London & 16 & 166 & 92 & 55 & 313 \\
Wenceslas Square, Prague & 306 & 237 & 10 & 44 & 291 \\
Washington Square Park, New York City & 102 & 222 & 4 & 2 & 228 \\
Gendarmenmarkt ,Berlin & 258 & 87 & 11 & 21 & 228 \\
Place des Vosges ,Paris & 294 & 53 & 21 & 12 & 86 \\
Grand Place,Brussels & 61 & 21 & 23 & 36 & 80 \\
Royal Mile,Edinburgh & 162 & 52 & 10 & 5 & 67 \\
Bruges & 224 & 33 & 4 & 24 & 61 \\
Grote Markt,  Antwerp & 472 & 32 & 6 & 10 & 48 \\
Old Town , Stockholm & 238 & 16 & 7 & 2 & 25 \\
Marienplatz , Munich & 65 & 16 & 3 & 1 & 20 \\\\

\bottomrule 
\end{tabular}

}
\caption{The distribution of image pairs across different 3D scenes in the test sets of \dataset{} (\cambridge{} and \megadepth{}). The scene number denotes the number of landmark in MegaDepth~\cite{li2018megadepth} dataset.
}
\label{tab:dataset}
\end{table}

 In Table~\ref{tab:dataset}, we provide the image pair distribution across the different scenes of \cambridge{} and \megadepth{}. The distribution for the training set is provided in the accompanying \href{https://drive.google.com/file/d/13SeXcpj3hkSfeOrWUSBKteqlBAYIUi0-/view?usp=drive_link}{$train\_set\_scenes.txt$} file.

\section{Implementation Details}
\label{sec:details-supp}

\subsection{Network Architecture}
Our approach employs an encoder-decoder architecture to predict three Euler angles of relative rotation. Specifically, we utilize LoFTR as our image encoder, extracting its dense features (after LoFTR's stage 2) with the dimensions of  $256\times \frac{H}{8}\times \frac{W}{8}$. Then, by concatenating the two features along the third dimension, we obtain a feature map of size $256\times \frac{H}{8}\times \frac{2W}{8}$. We augment the feature dimension by concatenating three auxiliaries masks: keypoints, matches and segmentation mask. This augmented input is then projected into the transformer decoder embedding space, which has a dimensionality of 256. Additionally, we introduce three learnable tokens. Our rotation estimation transformer builds upon the DinoV2 ViT architecture with a patch size of 1, 4 attention heads, and a depth of 8. Finally, the transformer's output is normalized and average-pooled. To estimate the relevant angle, we concatenate the averaged feature token with the corresponding learnable token. Next, we input this concatenated representation into a stack of three fully connected layers resulting in a 360-dimensional output distribution.
Overall, our model comprises approximately 80 million parameters in total, including LoFTR and SegFormer, with 22 million of those being learnable parameters. 

\subsection{Input Euler angles} The Euler angles that are fed into the rotation estimation transformer are learnable tokens (previously demonstrated in ~\cite{hong2024lrm} and ~\cite{wang2023pf} ) 
These tokens are initially set with random values drawn from a standard normal distribution. During training, they adaptively identify and focus on relevant image tokens specific to each Euler angle. During inference, these learnable tokens are initialized by loading from the optimized weights.
\subsection{Auxillary Channels} 
\subsubsection*{Keypoints \& Matches} 
The keypoints are extracted from the LoFTR output. Additionally, for the matches mask, we use geometric verification using RANSAC (re-projection error set to 1 and confidence to 0.99) to estimate the Fundamental matrix and filter out any outliers. Only the keypoints with a confidence value which is greater than 0.8 are considered for geometric verification. We create binary masks for both keypoints and matches, which are then rescaled to match the dimensions of LoFTR features ($\frac{H}{8}\times\frac{W}{8}$). Finally, we concatenate these image masks side by side to obtain the resulting size of $2\times\frac{H}{8}\times\frac{2W}{8}$. 
\subsubsection*{Segmentation Mask} 
The segmentation map was generated using SegFormer-B3. \href{https://drive.usercontent.google.com/download?id=1-OmW3xRD3WAbJTzktPC-VMOF5WMsN8XT&authuser=0}{SegFormer~\cite{enze2021segformer}} has demonstrated strong performance on outdoor image datasets, such as Cityscapes and ADE-20K. We consider the following categories: \emph{sky}, \emph{building}, \emph{road}, \emph{sidewalk}, \emph{streetlight}. All remaining labels are labeled as \emph{other}. Additionally, \emph{road} and \emph{sidewalk} are grouped together, as the borders of their masks are noisy and both labels have similar 3D spatial context. Finally, we resize the modified segmentation mask to match the LoFTR feature dimensions using NEAREST\_EXACT interpolation mode. The resulting concatenated mask has a dimensionality of 
$1\times\frac{H}{8}\times\frac{2W}{8}$.

\subsection{Training Details}
In all of our experiments, we used Adam optimizer ($\beta_{1} = 0.5, \beta_{2} = 0.9$). The  first stages are trained with a single learning rate set to $1\times 10^{-4}$, the \shortdataset{} and \instructexp{} stages are trained with learning rate of  $1\times 10^{-5}$. The batch size is 20, except when finetuning over \shortdataset{}, where it is adjusted to 40 as it achieved a cleaner convergence. The  training duration on one Nvidia RTX A5000 are as follows: [5 days, 3 days, 1 hour, 12 hours] for the [\cite{cai2021extreme},\fovexp{},\instructexp{},\shortdataset{}] stages, respectively. The total number of epochs for the training process is 34. The number of iterations sufficient to convergence is roughly 700K iteration for the first two stages and 3K for the last 2 stages. While training on the \shortdataset{} dataset, we addressed the imbalance of overall relative rotations for non overlapping pairs in \dataset{} by using a weighted random sampler. This sampler assigned weights based on the overall rotation angle. Additionally, for the overlapping pairs, due to the overlap categories imbalance (40000 images for large overlap and 15492 for small overlap), we employed a weighted random sampler that weighted by the overlap category. We used the balanced validation split of  \dataset{} to monitor training progress of \shortdataset{}  (with a stopping criterion of MGE improvement dropping below 0.5$^\circ$).

\begin{figure}[ht]
\centering
    \includegraphics[height=5cm, trim={0 0 0 0},clip]{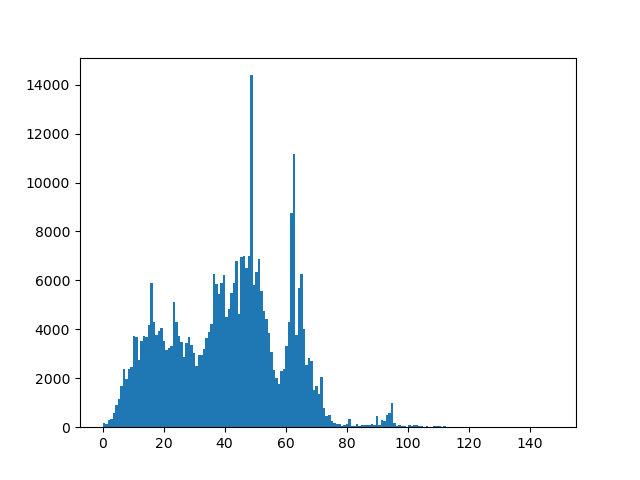}

    \caption{\textbf{Histogram of FoVs, corresponding to the \dataset{} training set}. As mentioned in the text, this was used for sampling square perspective images for \fovexp{} with a distribution that closely resembles in-the-wild image pairs. 
    }
    \label{fig:fov_hist}
\end{figure}
\subsection{Training with Data Augmentations}
\label{sec:app-aug}
\subsubsection*{Field of View Augmentations (\fovexp{})}
To better model the  distribution of FoV associated with in-the-wild image pairs, we analyzed the distribution of \shortdataset{} training set, as shown in Figure~\ref{fig:fov_hist} (considering $\text{fov}_x$ and $\text{fov}_y$ together).
The data revealed that both the median and average FoV values significantly deviate from 90 degrees which is assumed by prior work. Therefore, to better resemble to in-the-wild images, we curated a new dataset of panorama perspective images with a range of FoVs. These images were cropped into squares, with a FoV value selected from a Gaussian distribution with mean set according to the \shortdataset{} training set mean ($\mu_\shortdataset{}=40.9^\circ$) and a standard deviation which is 1.5 times the standard deviation of the \shortdataset{} training ($\sigma_\shortdataset{}=21.2$).The distribution was adjusted to exclude values below 30$^\circ$ and above 90$^\circ$ to maintain reasonable image quality. This was achieved by clipping these regions and subsequently re-normalizing the distribution. 
Following DUST3R~\cite{wang2023dust3r}, we also crop the images by the following aspect ratios [(256, 192), (256, 168), (256, 144), (256, 128)]. After cropping, we added zero padding to the remaining areas to achieve a uniform size of 256x256 for all images. For each batch, a single aspect ratio was chosen. We follow the method introduced in \cite{cai2021extreme},  
first training on overlapping pairs and then on non-overlapping pairs. The overlap training set includes 432992 pairs (35\% large overlap, 65\% small overlap), non overlap training set includes 1067764 pairs (15\% large overlap, 30\% small overlap and 55\% non overlapping pairs). 
\subsubsection*{Image-Level Appearance Augmentations (\instructexp{})}
As mentioned in the main paper, to perform image-level appearance augmentations, we apply the conditional InstructPix2Pix\cite{brooks2023instructpix2pix}  model on a subset of our data. The InstructPix2Pix editing process uses images that have been augmented with different field of views. The training set for this stage consists 18913 pairs. The model, \href{http://instruct-pix2pix.eecs.berkeley.edu/instruct-pix2pix-00-22000.ckpt}{instruct-pix2pix-00-22000.ckpt}, was configured with a text coefficient (cfg-text) of 7.5 and an image coefficient (cfg-image) of 1.5, over a total of 100 steps. 

 Given that InstructPix2Pix’s parameters are applied uniformly across all images, individual responses to the edits can vary, occasionally altering the image’s structure. Such alterations could potentially interfere with the cues necessary for estimating relative rotation. To mitigate this, a post-processing filtering stage was implemented in order to remove images whose structure was modified. 

\begin{figure*}[ht]
    \centering

    \begin{tabular}{@{}@{}c@{}r@{}l@{}c@{}r@{ }l@{}c@{}r@{ }l@{}c@{}r@{ }}

        \rotatebox{90}{\makebox[0pt][c]{High FAS}} \hspace{0.1cm}&
        \rotatebox{90}{\makebox[40pt][c]{Orig}}
        \hspace{0.1cm}&
        \includegraphics[height=1.4cm, trim={0 0 0 0},clip]{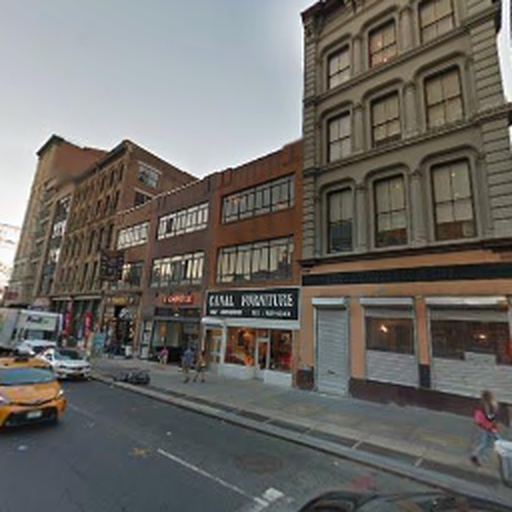}&

        \includegraphics[height=1.4cm, trim={0 0 0 0},clip]{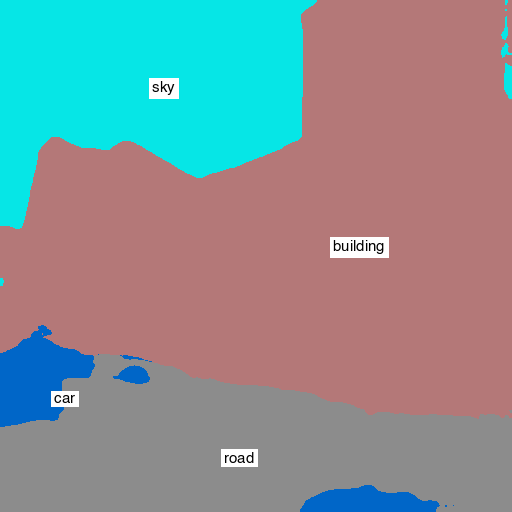} &
        \includegraphics[height=1.4cm, trim={0 0 0 0},clip]{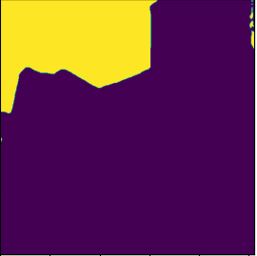}
        \hspace{0.3cm} &
        \includegraphics[height=1.4cm, trim={0 0 0 0},clip]{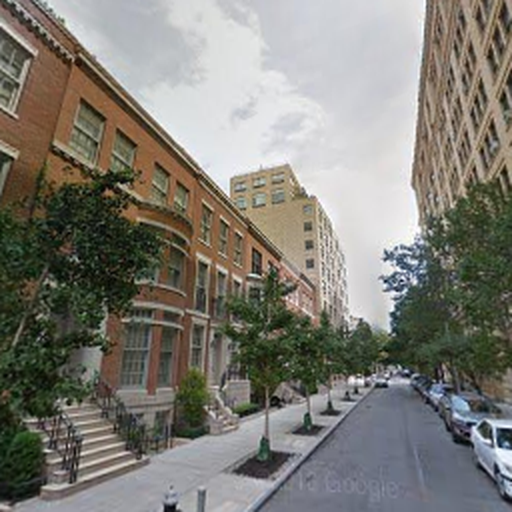}&
 
        \includegraphics[height=1.4cm, trim={0 0 0 0},clip]{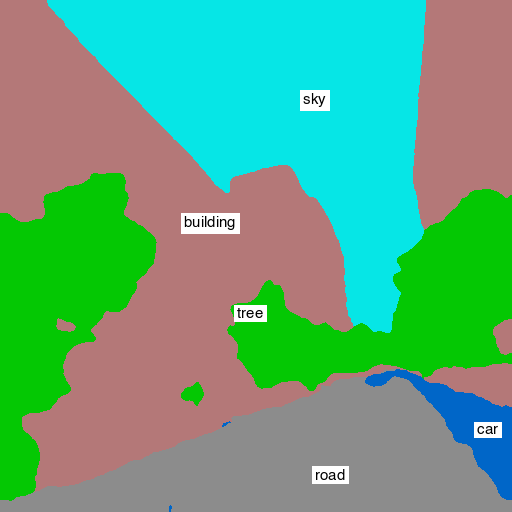} &
        \includegraphics[height=1.4cm, trim={0 0 0 0},clip]{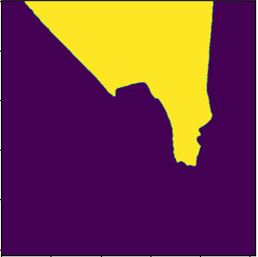}
        \hspace{0.3cm} &
        \includegraphics[height=1.4cm, trim={0 0 0 0},clip]{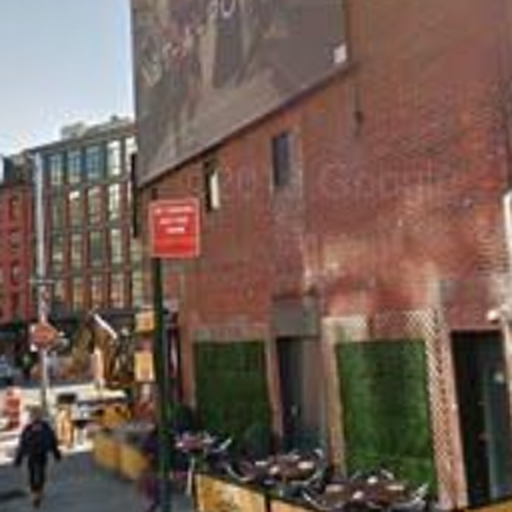}&

        \includegraphics[height=1.4cm, trim={0 0 0 0},clip]{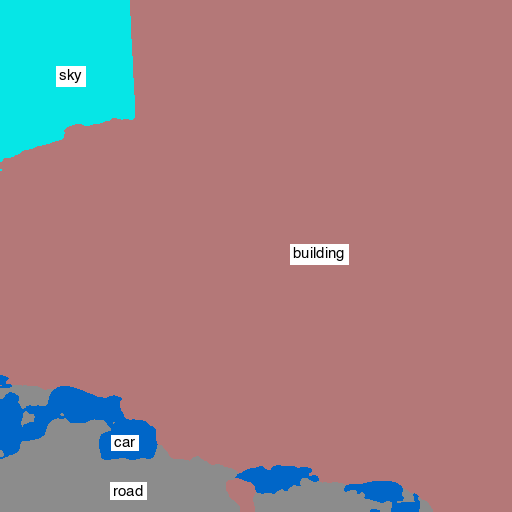} &
        \includegraphics[height=1.4cm, trim={0 0 0 0},clip]{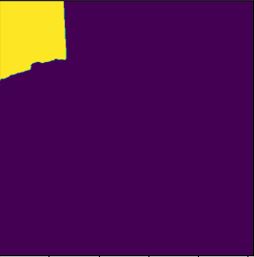}\\
        &\rotatebox{90}{\makebox[40pt][c]{Instruct}}
        \hspace{0.1cm}&
        \includegraphics[height=1.4cm, trim={0 0 0 0},clip]{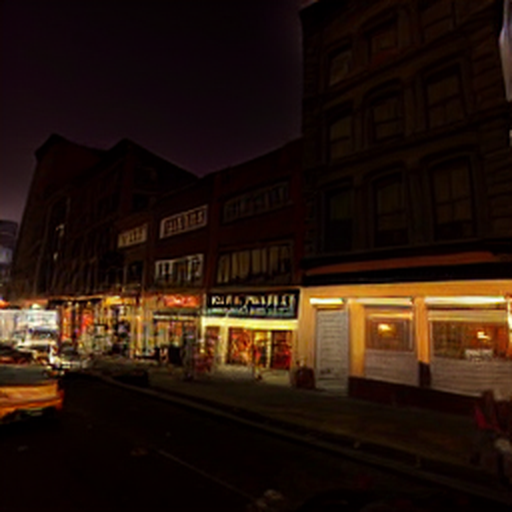}&

        \includegraphics[height=1.4cm, trim={0 0 0 0},clip]{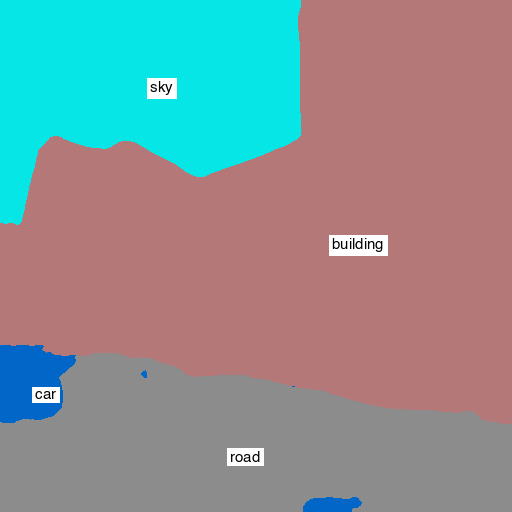}&
        \includegraphics[height=1.4cm, trim={0 0 0 0},clip]{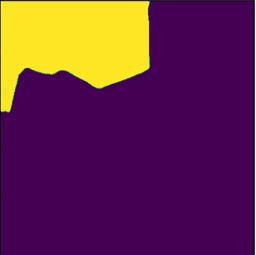}
        \hspace{0.3cm} &
        \includegraphics[height=1.4cm, trim={0 0 0 0},clip]{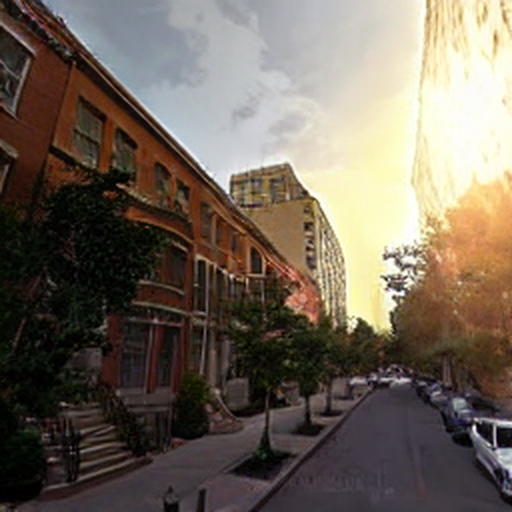}&

        \includegraphics[height=1.4cm, trim={0 0 0 0},clip]{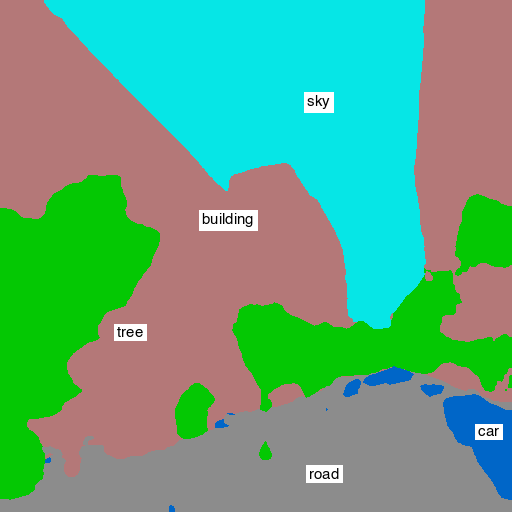} &  
        \includegraphics[height=1.4cm, trim={0 0 0 0},clip]{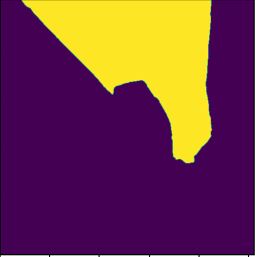}
        \hspace{0.3cm} &
        \includegraphics[height=1.4cm, trim={0 0 0 0},clip]{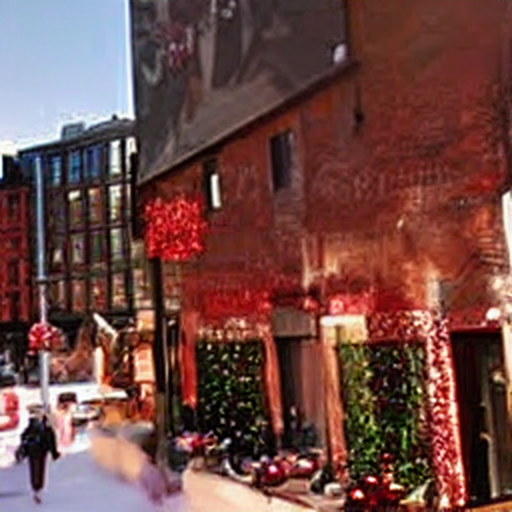}&

        \includegraphics[height=1.4cm, trim={0 0 0 0},clip]{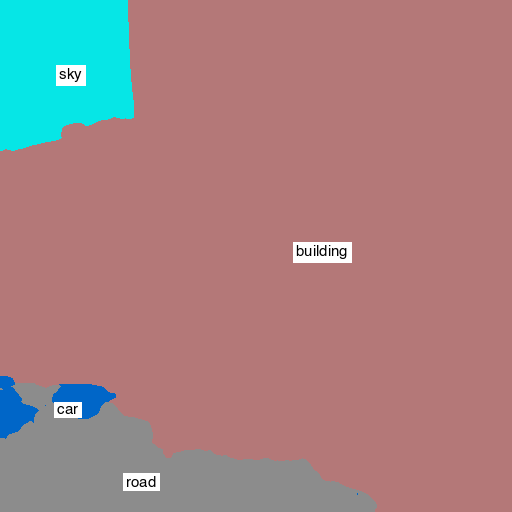} &
        \includegraphics[height=1.4cm, trim={0 0 0 0},clip]{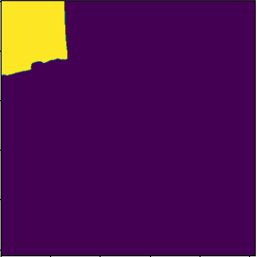}\\

        \rotatebox{90}{\makebox[0pt][c]{Low FAS}} &
        \rotatebox{90}{\makebox[40pt][c]{Orig}}\hspace{0.1cm}&
        \includegraphics[height=1.4cm, trim={0 0 0 0},clip]{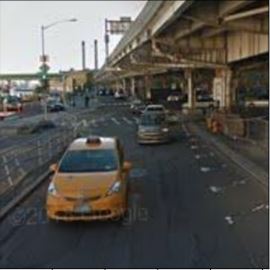}&

        \includegraphics[height=1.4cm, trim={0 0 0 0},clip]{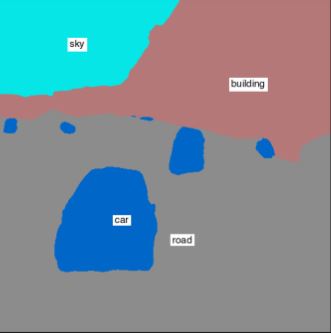} &
        \includegraphics[height=1.4cm, trim={0 0 0 0},clip]{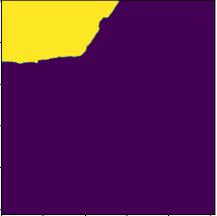}
        \hspace{0.3cm} &
        \includegraphics[height=1.4cm, trim={0 0 0 0},clip]{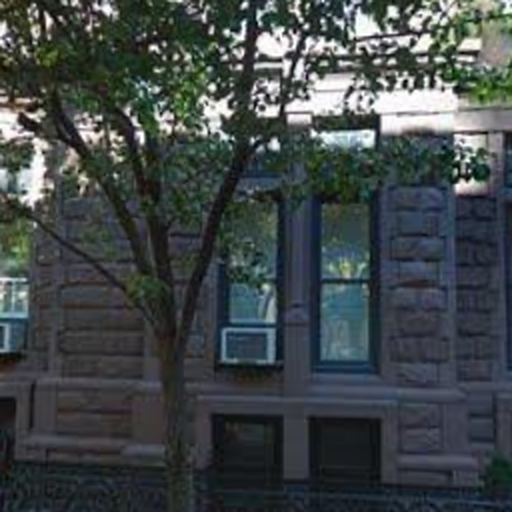}&

        \includegraphics[height=1.4cm, trim={0 0 0 0},clip]{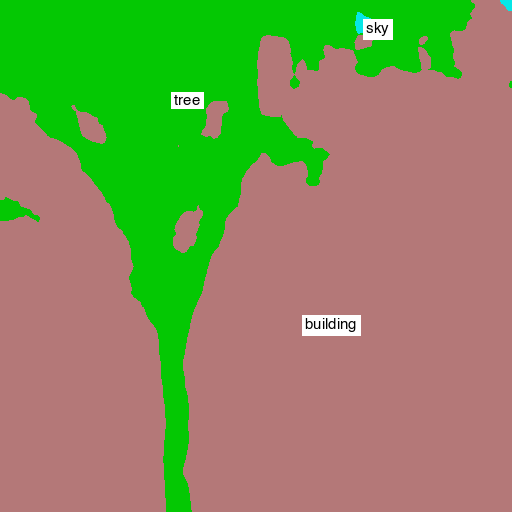} &
        \includegraphics[height=1.4cm, trim={0 0 0 0},clip]{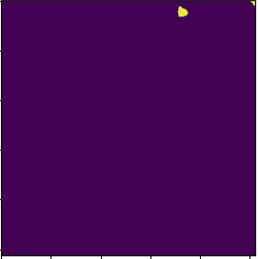}
        \hspace{0.3cm} &
        \includegraphics[height=1.4cm, trim={0 0 0 0},clip]{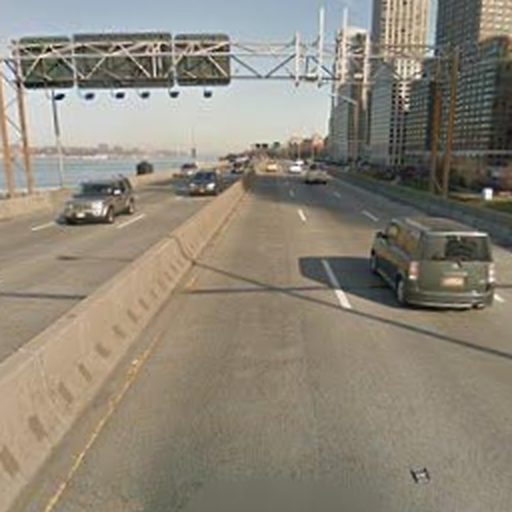}&

        \includegraphics[height=1.4cm, trim={0 0 0 0},clip]{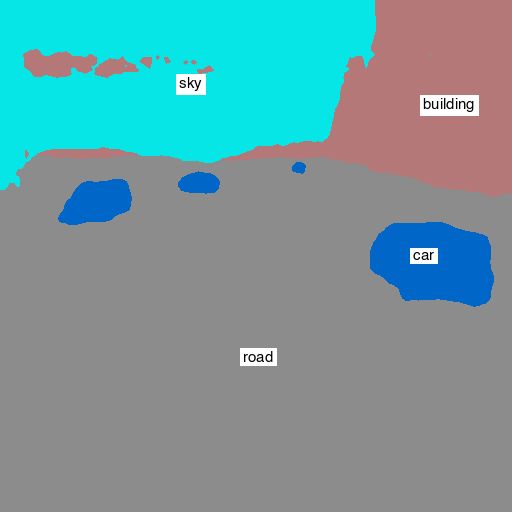} &
        \includegraphics[height=1.4cm, trim={0 0 0 0},clip]{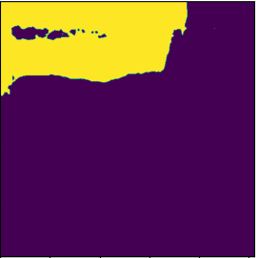}\\
        &\rotatebox{90}{\makebox[40pt][c]{Instruct}} \hspace{0.1cm}&
        \includegraphics[height=1.4cm, trim={0 0 0 0},clip]{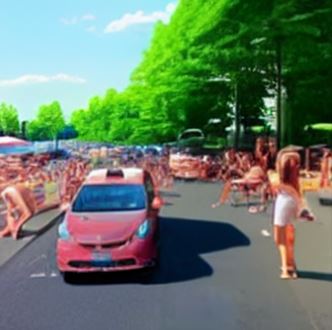}&

        \includegraphics[height=1.4cm, trim={0 0 0 0},clip]{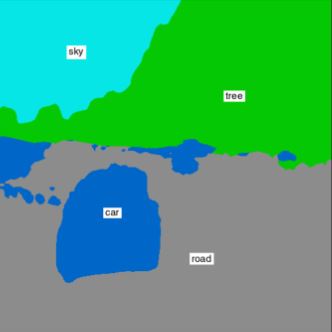} &
        \includegraphics[height=1.4cm, trim={0 0 0 0},clip]{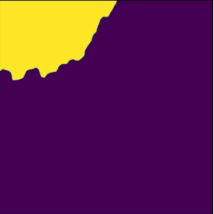}
        \hspace{0.3cm} &
        \includegraphics[height=1.4cm, trim={0 0 0 0},clip]{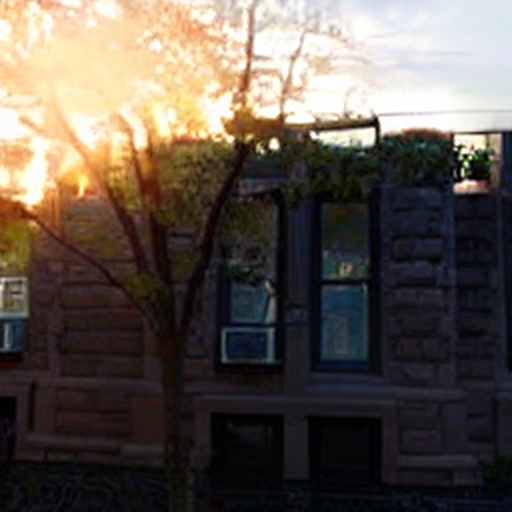}&

        \includegraphics[height=1.4cm, trim={0 0 0 0},clip]{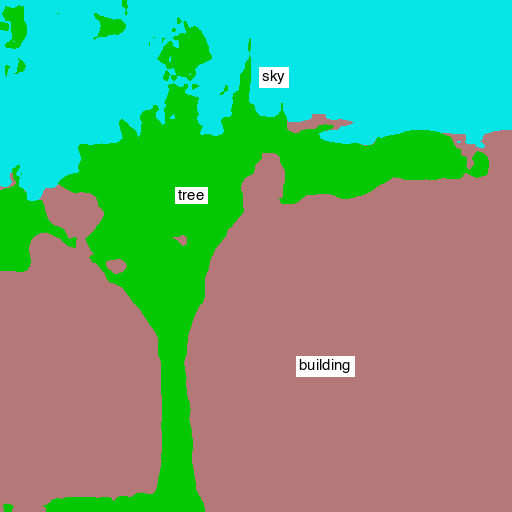} & 
        \includegraphics[height=1.4cm, trim={0 0 0 0},clip]{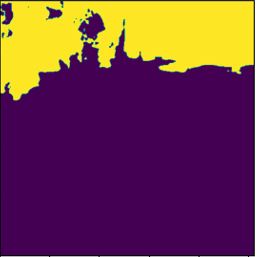}
        \hspace{0.3cm} &
        \includegraphics[height=1.4cm, trim={0 0 0 0},clip]{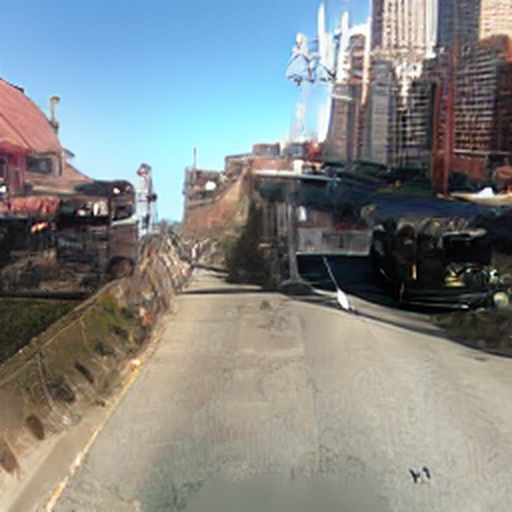}&

        \includegraphics[height=1.4cm, trim={0 0 0 0},clip]{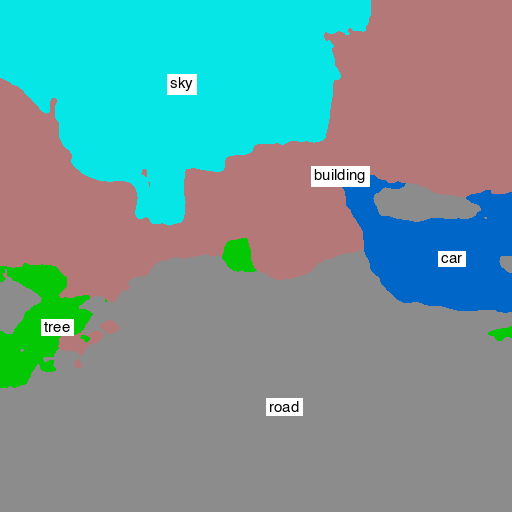} &
        \includegraphics[height=1.4cm, trim={0 0 0 0},clip]{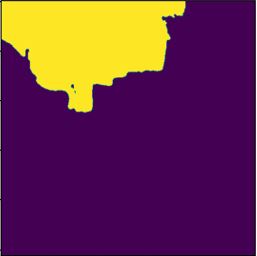}\\

    \end{tabular}
    \caption{\textbf{Filtering InstructPix2Pix Image augmentations.} The \instructexp{} training set was filtered using our Filtering Augmentation Score (FAS), detailed in Section \ref{sec:app-aug}. Above, we present three examples with high (top rows) and low (bottom rows) FAS values. Each example shows two rows: the original image with its segmentation map and binary mask ($M_0$), followed by the InstructPix2Pix transformation with its corresponding segmentation and mask ($M_t$). As illustrated above, these binary masks provide meaningful cues for whether or not the model's output modifies the scene structure, which is undesirable for our problem setting. }
    \label{fig:instrut-filtering}
\end{figure*}

The filtering process quantifies structural consistency by analyzing the primary scene boundaries (predominantly the skyline) through a comparative analysis of the segmentation maps from both the source image and its InstructPix2Pix transformation.
These segmentation maps are generated using \href{https://drive.usercontent.google.com/download?id=1-OmW3xRD3WAbJTzktPC-VMOF5WMsN8XT&authuser=0}{SegFormer\cite{enze2021segformer}}. Prior to filtering, we exclude images that lack static structural elements (roads, buildings) or containing predominantly transient features like cars and people. 
Additionally, indoor scenes (e.g., tunnels) are excluded due to their limited relevant segmentation labels, which generate noisy segmentation maps.

 To identify the main borderlines, we focus on the specific labels ‘building’, ‘road’, ‘sky’, ‘tree’, and ‘car’, and apply the softmax function exclusively to these categories. We define a main category (C) whose boundary will serve as the major boundary.  The main category is selected as the first available mask in the following order: ‘sky’, ‘building’, ‘road’. The categories not chosen as the main category are defined as secondary categories.

 The binary mask $M_{c}$ is designed to  highlight changes in the major boundary and assign a score reflecting the degree of change. Let $M_{o}$ represent the binary mask of C of the original image and $M_{t}$ the binary mask of C  after InstructPix2Pix transformation. To identify the original boundary, we apply erosion and dilation techniques to $M_{o}$ (using a disk size of 5 for ‘sky’ and 10 for ‘building’ and ‘road’). The difference between the eroded and the dilated mask of the original image creates a pronounced border around the main category, denoted as $M_{b}^{o}$. Next, we construct $M_{c}$, the binary mask for the altered pixels, using the formula $M_{c} = M_{o} \cup M_{t}-M_{o} \cap M_{t} - M_{b}^{o}$. Additionally, we exclude any transient elements like ‘trees’ and ‘cars’ from the count of altered pixels in $M_{c}$. 

 The score for the major borderline $S_c$, is then calculated by the following formula $S_c = 1-\frac{\sum_{i,j} {M_{c}}_{i,j}}{\sum_{i,j} {M_{o} \cup M_{t}}-_{i,j}}$
, which gives us a measure of the amount of unwanted change for C.  Additionally, if the total number of changed pixels exceeds 500, the score for the main category is set to zero.

 The final Filtering Augmentation Score (FAS) is composed from a multiplication of $S_c$ and a binary score for the secondary categories.
 We assign a binary score to the secondary categories to evaluate if they have undergone a significant change. If a second category’s mask exceeds 10,000 pixels and more than 50\% of the pixels have changed, we assign a score of 0; otherwise, we assign a score of 1. We empiricially set the FAS threshold to 0.92.
 
In Figure~\ref{fig:instrut-filtering}, we provide three examples of InstructPix2Pix transformations with high FAS and three with low FAS (\emph{i.e.}, which are excluded from the final training set). For each example, the top row respectively shows the original image, the reduced label segmentation map, and $M_{o}$. The second row displays the image after InstructPix2Pix transformation, its reduced label segmentation map, and $M_{t}$. The left column on the top row shows a transformation to a night time scene. Note that SegFormer effectively identifies the building despite the darkness, and $M_{o}$ is closely matches  $M_{t}$. The middle column illustrates a transformation to sunrise. In the top example, SegFormer accurately identified the building lines despite that severe transformation. However, in the bottom example, the transformation was too extreme, turning the building into sky (which is accurately reflected in both $M_{o}$ and $M_{t}$). In the right column, the top example demonstrates that the transformation into a Christmas scene altered the scene (and the original segmentation map). However, by reducing the labels, it becomes clear that the relevant semantic regions remained unchanged. In the bottom right example, InstructPix2Pix algorithm's transformation failed, and on the bottom left example the transformation to summer changed the structure of the scene. Both correctly identified by reduced label segmentation map. Note that although the change of the skyline (the change between $M_{o}$ and $M_{t}$) in the bottom left example is not drastic, it is accurately assigned a low FAS score.

\subsection{Baselines}
We compare our method with six pre-existing methods for rotation extraction, including both classical (SIFT) and learning-based methods (LoFTR, 8PointVit, ExtremeRotation, CascadedAtt, Dust3R). The methods estimate the pose for each pair, and the evaluation metrics are calculated from the pose estimation, as described in the main paper. For SIFT and LoFTR, the pose estimation might fail due to an insufficient number of points that pass the geometric method (RANSAC). These pairs are excluded from the evaluation metric statistics, as is done in previous works. It is important to mention that this exclusion provides an advantage to these methods, as removing the invalid pairs likely reduces the number of “hard” pairs, thereby increasing their corresponding metric statistics. The percentage of invalid pairs is presented to capture this phenomenon.

 \subsubsection*{SIFT}
The SIFT pipeline uses the OpenCV Python library, and the default random number generator is initialized with the seed value 12345. Images are resized so that their largest dimension is 256, and their intrinsic matrices are adjusted accordingly.  
Keypoints and descriptors for both grayscale images are detected using the SIFT detector. The 'Flann' KNN algorithm is used to match the key points, followed by filtering matches using Lowe's ratio test, ensuring that the nearest distance is smaller than the next nearest by a factor of 0.7.
Image pairs with less than 6 matches are filtered.
Since image pairs can have different intrinsic camera parameters, the keypoints are adjusted so that their intrinsic camera parameters are represented by an identity matrix for essential matrix calculation. Essential matrix is then calculated using RANSAC  confidence = 0.999, threshold = 0.01. Finally, the pose is recovered from the essential matrix and the valid matched key points.  The success rates for large, small, none categories for \cambridge{} test set: 91.4\%, 58.6\%, 5.7\% and for \megadepth{} test set: 65.6\%, 43.2\%, 7.7\% . 

 \subsubsection*{LoFTR}
 The LoFTR pipeline also utilizes the OpenCV Python library, and the default random number generator is initialized with the seed value 12345. Images are resized so that their largest dimension is 256, and their intrinsic matrices are adjusted accordingly.  
Key points and descriptors for both grayscale images are detected using the \href{https://drive.google.com/drive/folders/1xu2Pq6mZT5hmFgiYMBT9Zt8h1yO-3SIp}{LOFTR pretrained model('outdoor\_ds')}. Image pairs with less than 20 matches are filtered.
Since image pairs can have different intrinsic camera parameters, the keypoints are adjusted so that their intrinsic camera parameters are represented by an identity matrix for essential matrix calculation. Essential matrix is then calculated using RANSAC  confidence = 0.999, threshold = 0.01. Finally, the pose is recovered from the essential matrix and the valid matched key points.
The success rates for large, small, none categories for \cambridge{} test set: 97.0\%, 39.0\%, 0.0\% and for \megadepth{} test set: 82.9\%, 33.3\%, 0.46\% .

\definecolor{graytext}{RGB}{130,130,130}
\begin{table}[t]
\setlength{\tabcolsep}{1.7pt}
 \def\arraystretch{1}
\centering
\resizebox{0.45\textwidth}{!}{%
\begin{tabular}{llcccccccccccccc} 

    \toprule
    \multicolumn{2}{c}{}&&\multicolumn{3}{c}{StreetLearn}&&\multicolumn{3}{c}{\cambridge{}}&&
    \multicolumn{3}{c}{\megadepth{}}\\
    \cline{4-6} \cline{8-10}\cline{12-14}
      & Method &&MGE\textdownarrow & RRA$_{15}$\textuparrow &RRA$_{30}$\textuparrow && MGE\textdownarrow & RRA$_{15}$\textuparrow &RRA$_{30}$\textuparrow && MGE\textdownarrow & RRA$_{15}$\textuparrow &RRA$_{30}$\textuparrow\\
    \midrule

    \multirow{2}{*}{\rotatebox[origin=c]{90}{Large}} & 
    SIFT~\cite{lowe2004distinctive}&& 0.10  & 100.0 & 100.0 && 2.14  & 91.5 & 94.1 && 4.26  & 70.8 & 79.1  \\
    & LoFTR~\cite{sun2021loftr}&& 0.11 & 100.0 & 100.0 &&   1.84 & 98.6 & 99.6 &&  2.47 & 90.6 & 96.6  \\
    \midrule
    \multirow{2}{*}{\rotatebox[origin=c]{90}{Small}} & 
    SIFT~\cite{lowe2004distinctive}&& 0.36  & 96.1 & 98.2 && 4.98  & 67.0 & 72.0 && \textcolor{gray}{9.51}  & \textcolor{gray}{57.8} & \textcolor{gray}{68.0}\\
    & LoFTR~\cite{sun2021loftr}&& 0.48 & 100.0 & 100.0 &&   2.58 & 94.4 & 98.4 &&   \textcolor{gray}{5.57} & \textcolor{gray}{77.7} & \textcolor{gray}{92.6 }\\
    \midrule
    \multirow{2}{*}{\rotatebox[origin=c]{90}{None}} & 
    SIF*~\cite{lowe2004distinctive}&& \textcolor{gray}{97.94} & \textcolor{gray}{36.2} & \textcolor{gray}{38.1} && \textcolor{gray}{140.22}  & \textcolor{gray}{0.0} & \textcolor{gray}{1.8} && \textcolor{gray}{152.33}  & \textcolor{gray}{3.2} & \textcolor{gray}{6.4} \\
    & LoFTR~\cite{sun2021loftr}&& \textcolor{gray}{ - }& \textcolor{gray}{ - }& \textcolor{gray}{- } &&  \textcolor{gray}{ - }& \textcolor{gray}{ - }& \textcolor{gray}{- }&& \textcolor{gray}{14.10}& \textcolor{gray}{60.0}& \textcolor{gray}{100.0} \\
    \bottomrule
\end{tabular}
}

\vspace{-8pt}
\caption{\textbf{Homography-based estimation for feature-based techniques}. 
We evaluate performance over the \cambridge{} and \megadepth{} test sets, separately considering Large (top), Small (middle) and Non-overlapping (bottom) pairs. 
Note that median errors are computed only over successful image pairs, for which these algorithms output a pose estimate (failure over more than $50\%$ of the test pairs is shown in gray). }

\label{tab:homography-based}
\end{table}
In Table ~\ref{tab:main_result_val}, we followed prior works to calculate the essential matrix for SIFT and LoFTR. We also conducted an experiment to evaluate these methods using homography, with the results presented in Table ~\ref{tab:homography-based}. We observe that homography-based estimation yields improvements over the StreetLearn dataset where over our real-world test sets, the transformation appears to cause  distruptions, as evidenced by the increase in the MGE of SIFT for Large overlap from 2.1$^\circ$ to 2.5$^\circ$ in \megadepth{}. 
 \subsubsection*{8PointViT}
 The 8PointVit pipeline uses the \href{https://fouheylab.eecs.umich.edu/~cnris/rel_pose/modelcheckpoints/pretrained_models.zip}{'streetlearn' pretrained model}. Images are cropped around the center to a square dimension and resized to (256,256). The pose is then evaluated using the 8PointVit network.
 \subsubsection*{ExtremeRotation}
 The ExtremeRotation pipeline uses the \href{https://drive.google.com/drive/folders/1xA6O-FYAKWj0Ed2E3qIu-tKnw29C9q1Z?usp=sharing&pli=1}{'streetlearn\_cv\_distribution'} pretrained model. Images are cropped around the center to a square dimension and resized to (256,256). The relative Euler angles are then evaluated using the classification network.
 \subsubsection*{CascadedAtt}
 Since the model of CascadedAtt\cite{dekel2024estimating} 
was not released, we compare to it by directly implementing its encoder using weight-sharing Siamese residual U-nets, followed by cross-decoding with a weight-sharing transformer. These features are concatenated with Euler angle position embeddings and processed by our Rotation Estimation Transformer module, with the output rotation represented as relative Euler angles. We 
compare performance on the Streetlearn dataset evaluated in their work, finding that our re-implementation is mostly comparable, achieving only slightly lower MGE scores.

 \subsubsection*{Dust3R}
The checkpoint utilized was the \href{https://download.europe.naverlabs.com/ComputerVision/DUSt3R/DUSt3R_ViTLarge_BaseDecoder_512_dpt.pth}{512 DPT head}, which yielded superior outcomes compared to the Dust3R checkpoints. The optimization technique that led to the pose estimation is PnP RANSAC. The pair is fed into the model in two instances, with their positions reversed in each case, resulting in two separate estimations of relative rotation. The chosen relative rotation is the one associated with the greatest confidence score, which is determined by the product of the average values from the pair’s confidence maps.

 \section{Additional Results and Visualizations}
 \label{sec:results-supp}
\begin{figure}[ht]
\centering
    \rotatebox{90}{\makebox[50pt]{Large}}
    \includegraphics[height=1.95cm, trim={0 0 0 0},clip]{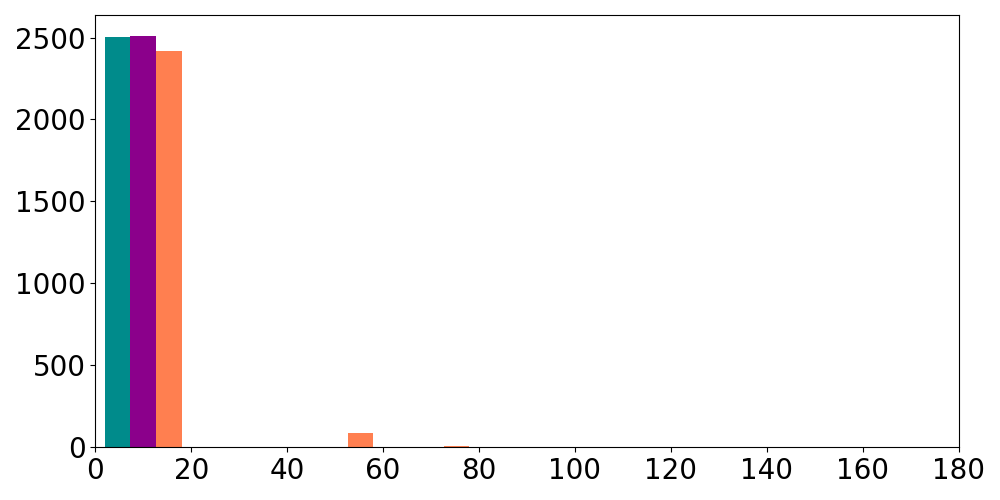}
    \includegraphics[height=1.95cm, trim={0 0 0 0},clip]{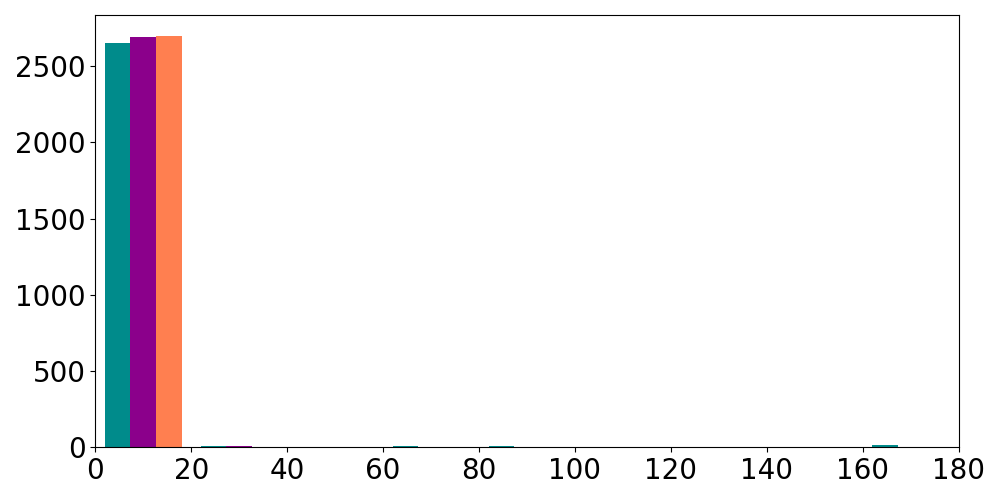}
    \\
    \rotatebox{90}{\makebox[50pt]{Small}}
    \includegraphics[height=1.95cm, trim={0 0 0 0},clip]{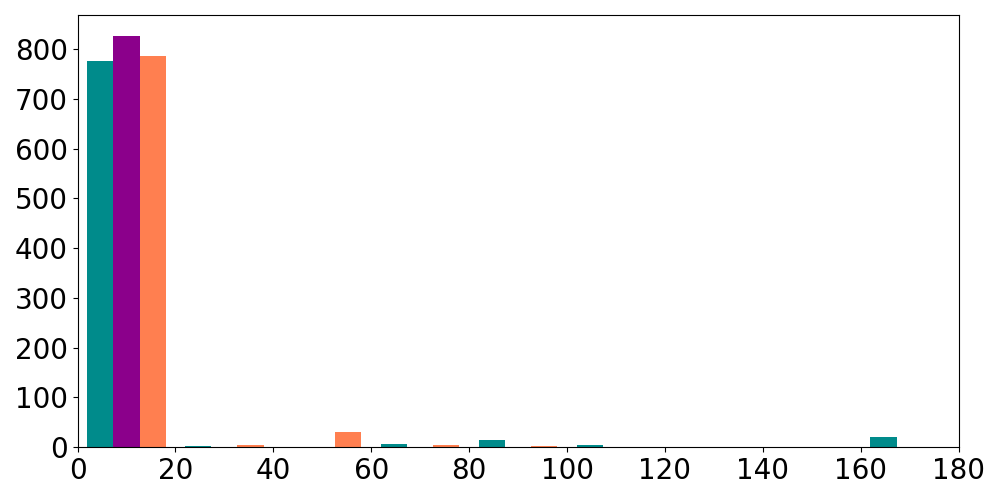}
    \includegraphics[height=1.95cm, trim={0 0 0 0},clip]{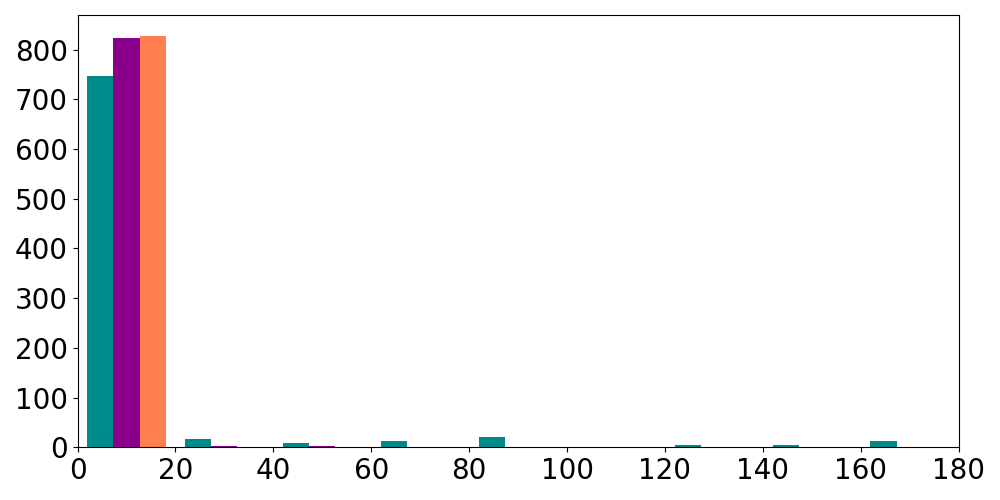}
    \\
    \rotatebox{90}{\makebox[50pt]{None}}
    \includegraphics[height=1.95cm, trim={0 0 0 0},clip]{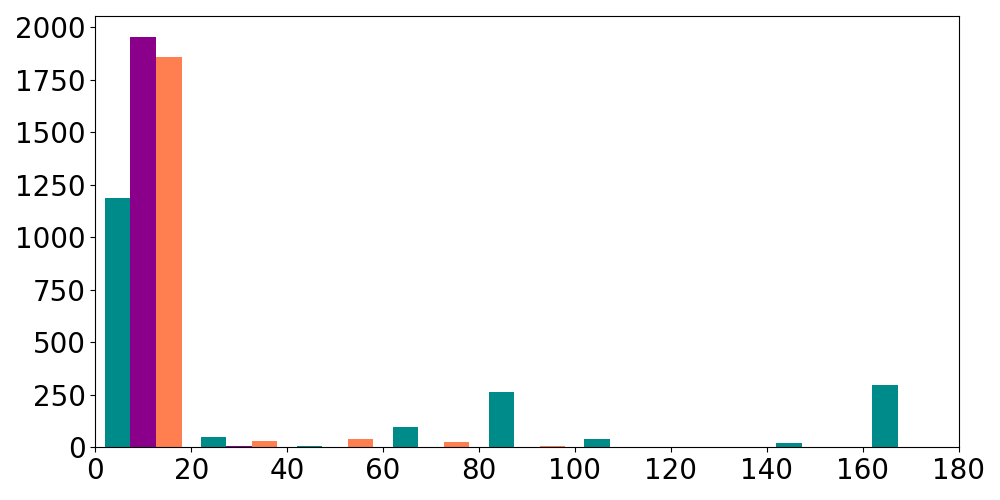}
    \includegraphics[height=1.95cm, trim={0 0 0 0},clip]{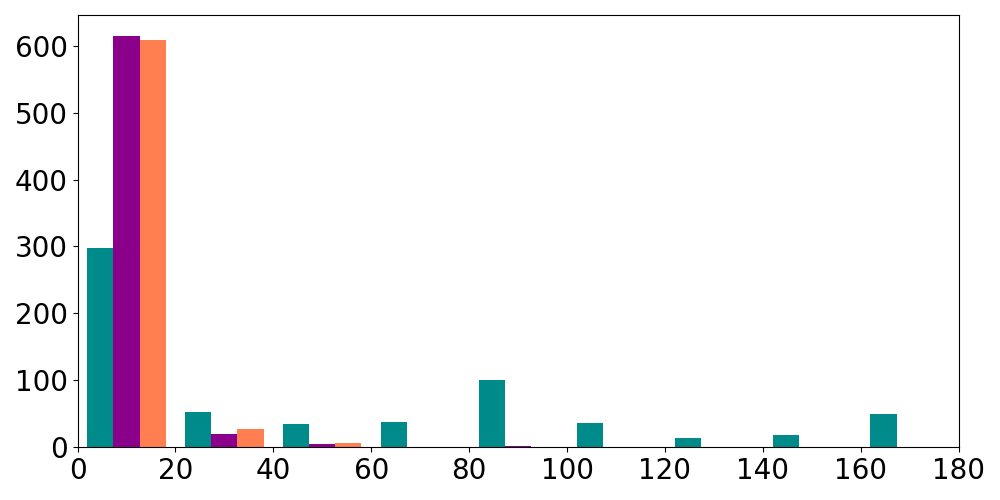}
    \\
    \makebox[120pt]{\cambridge}
    \makebox[100pt]{\megadepth}
    
    \caption{\textbf{Histograms over the {\color{MidnightBlue}{yaw}}, {\color{RedViolet}{pitch}} and {\color{BurntOrange}{roll}} relative rotation errors}. As illustrated above, yaw errors contribute significantly to the total error count. As detailed in \cite{cai2021extreme}, these yaw errors typically signify the uncertainties associated with non-overlapping pairs.  Notably, the error peaks tend to occur at intervals that are multiples of 90 degrees. 
    } 

    \label{fig:yaw_pitch_roll_hist}
\end{figure}

Another measure we use is \emph{Top 5}, where output rotation matrix is selected from the best of 5 picks in the 360-D yaw angle output (and the top1 pick for pitch and roll). 
Figure~\ref{fig:yaw_pitch_roll_hist} shows the geodesic error histograms for each relative rotation angle. The analysis reveals that yaw errors contribute significantly to the total error count. As detailed in ~\cite{cai2021extreme}, these yaw errors typically signify the uncertainties associated with non-overlapping pairs.  Notably, the error peaks tend to occur at intervals that are multiples of 90 degrees.

\begin{figure}[ht]
    \centering
    \resizebox{0.47\textwidth}{!}{%
    \begin{tabular}{r@{ }l@{}c@{}}

        \includegraphics[height=0.655cm, trim={0 0 0 0},clip]{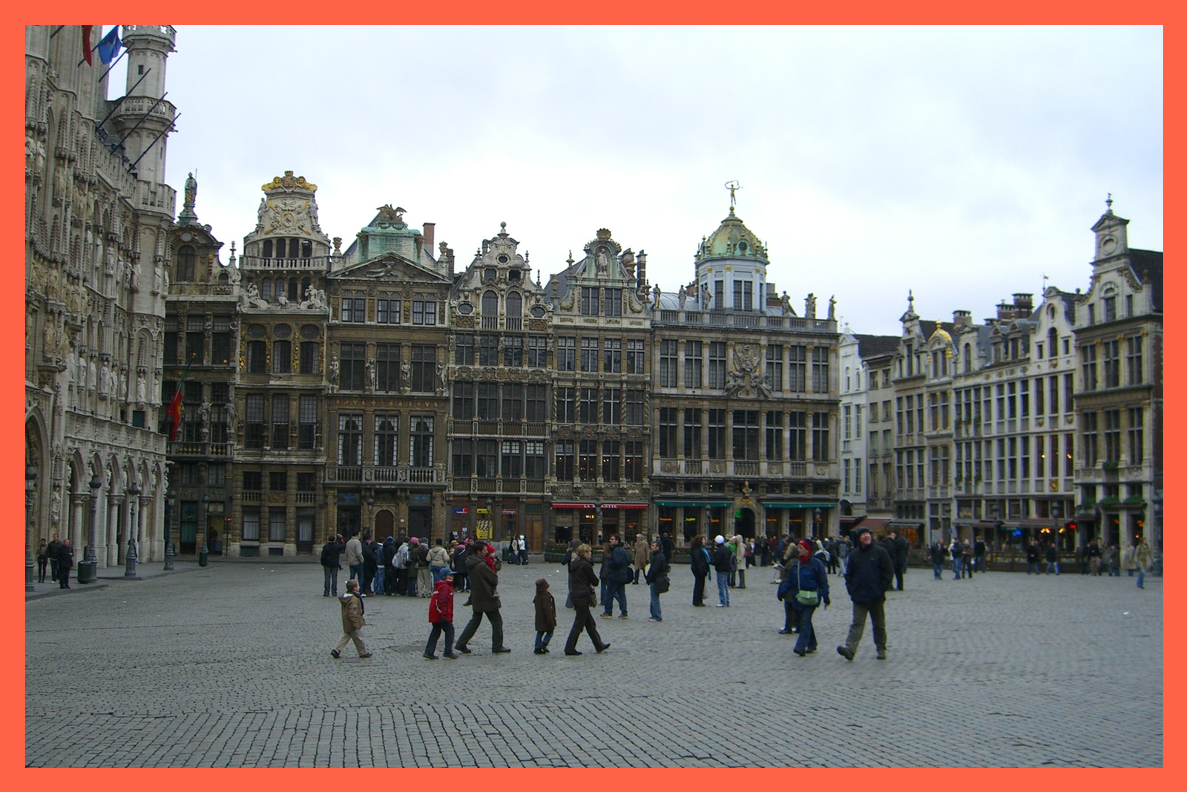}&
        \includegraphics[height=0.655cm, trim={0 0 0 0},clip]{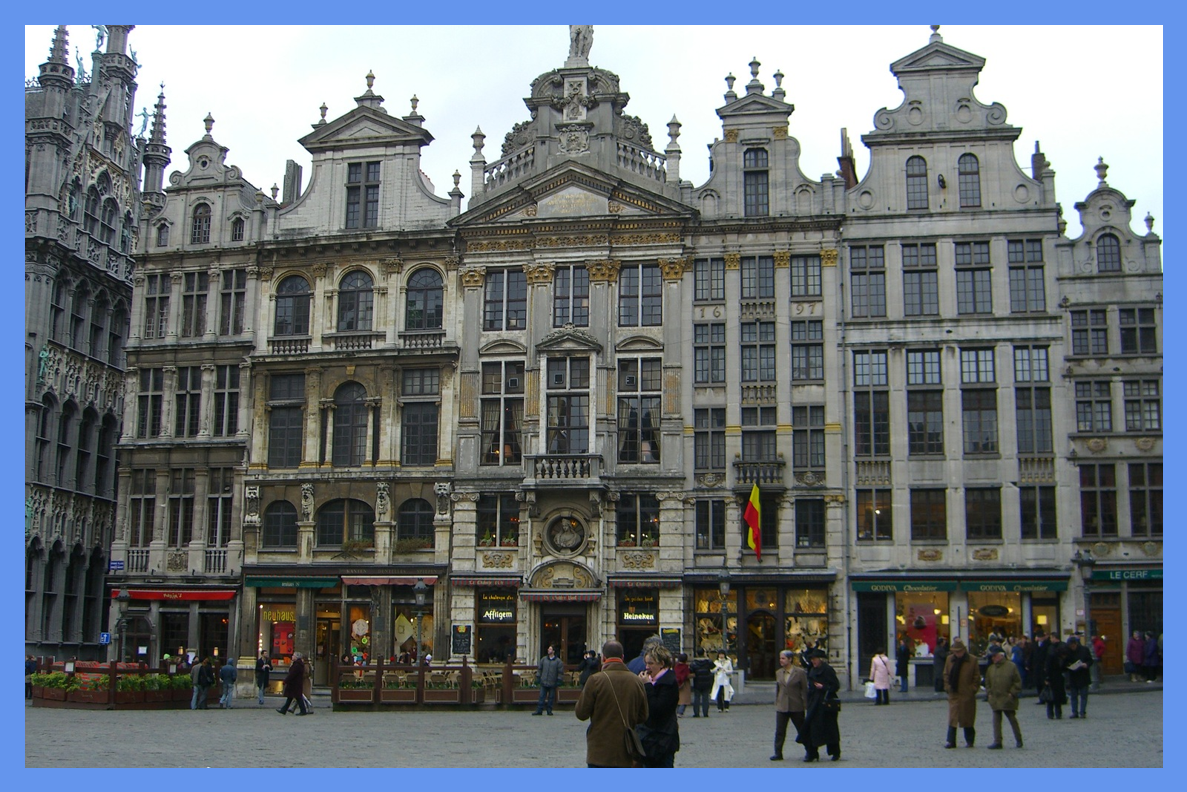} &
        \includegraphics[height=0.67cm, trim={0.2cm 0cm 0.4cm 0cm},clip]{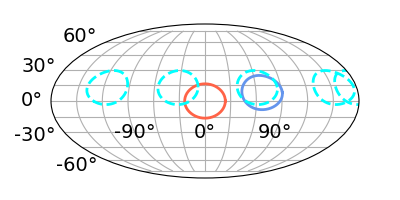} \\
        \includegraphics[height=0.73cm, trim={0 0 0 0},clip]{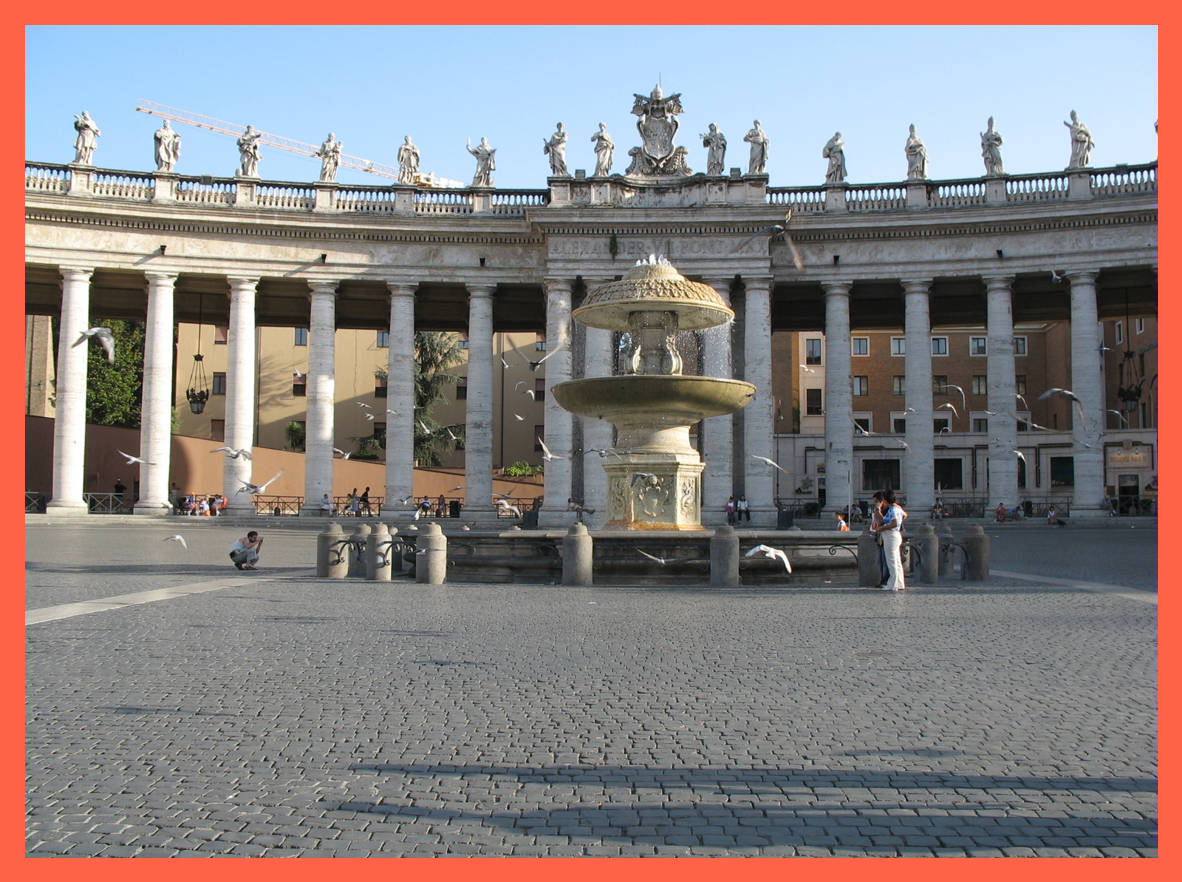}&
        \includegraphics[height=0.73cm, trim={0 0 0 0},clip]{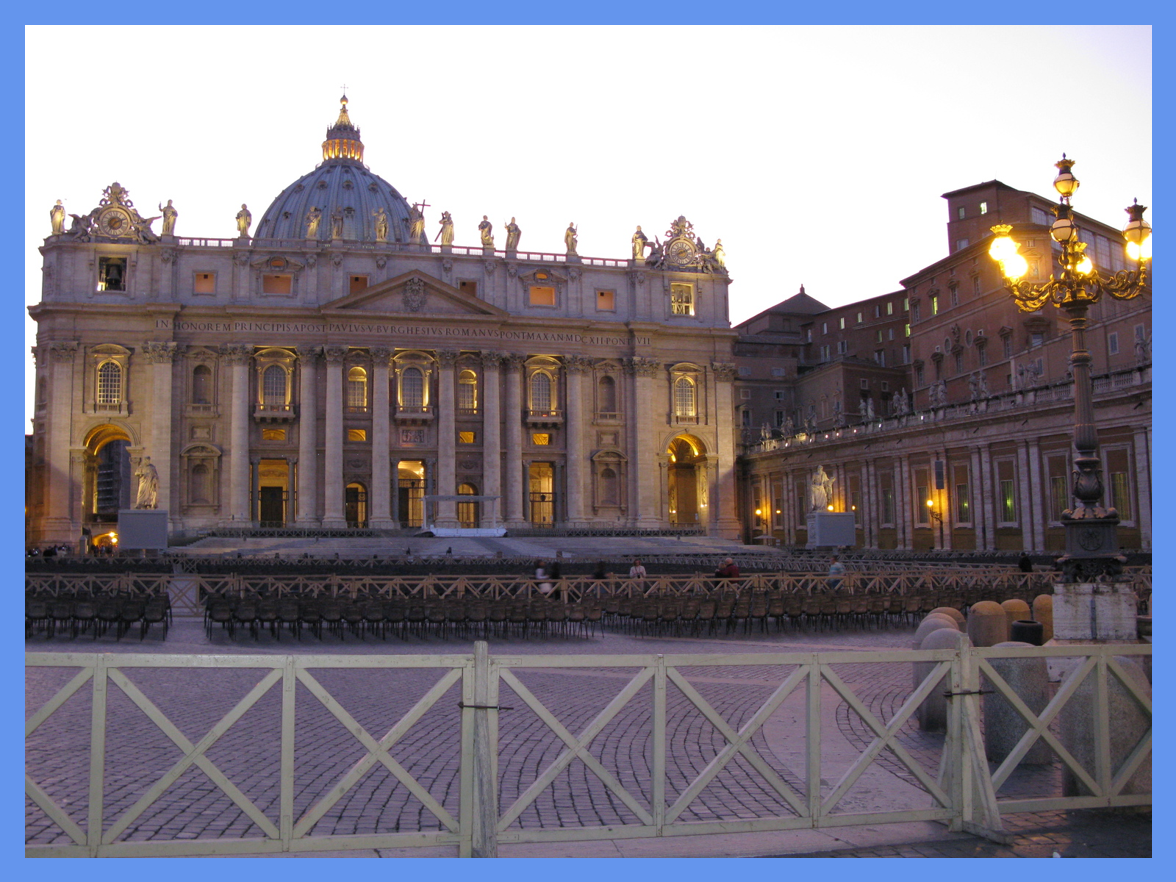} &
        \includegraphics[height=0.67cm, trim={0.2cm 0cm 0.4cm 0cm},clip]{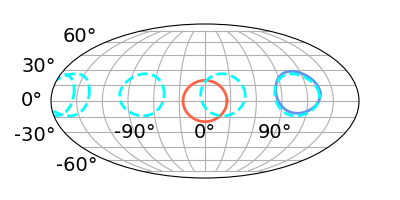} \\
        \includegraphics[height=0.74cm, trim={0 0 0 0},clip]{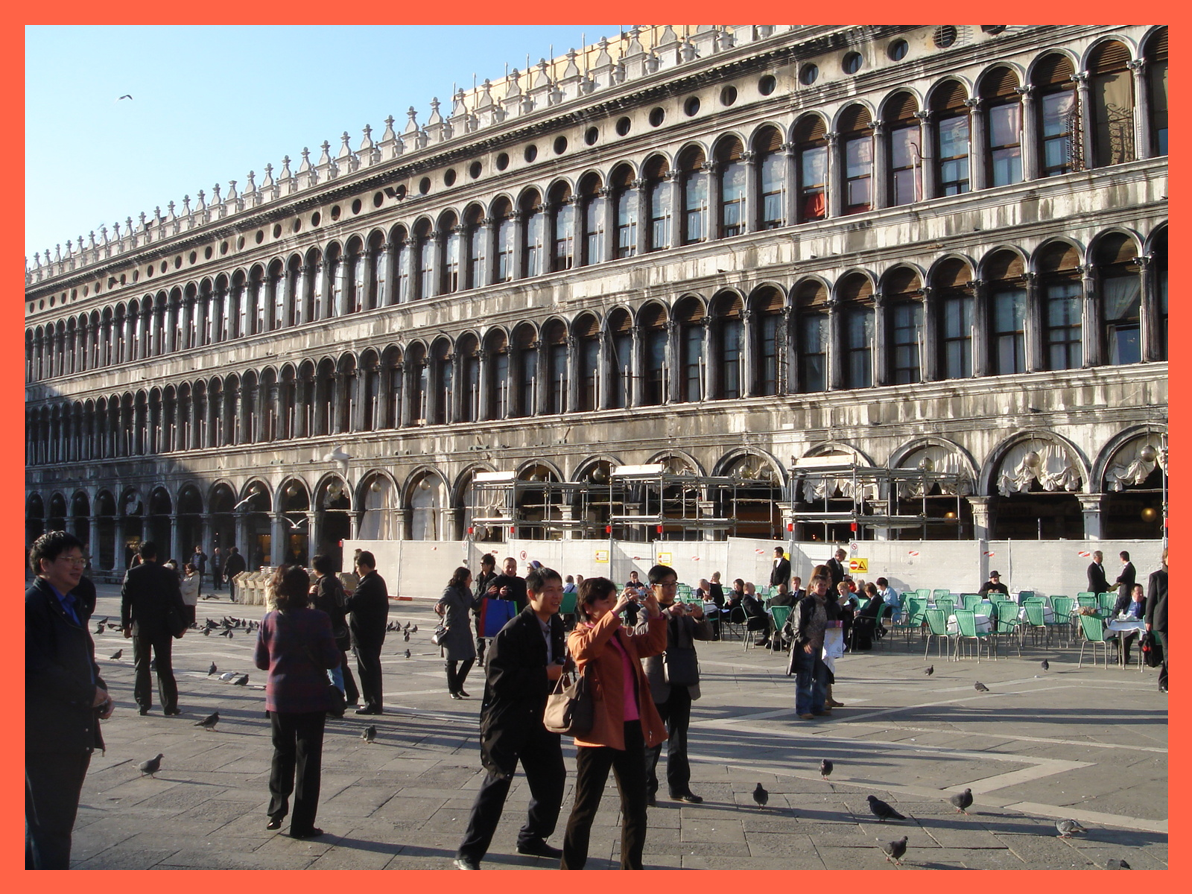}&
        \includegraphics[height=0.74cm, trim={0 0 0 0},clip]{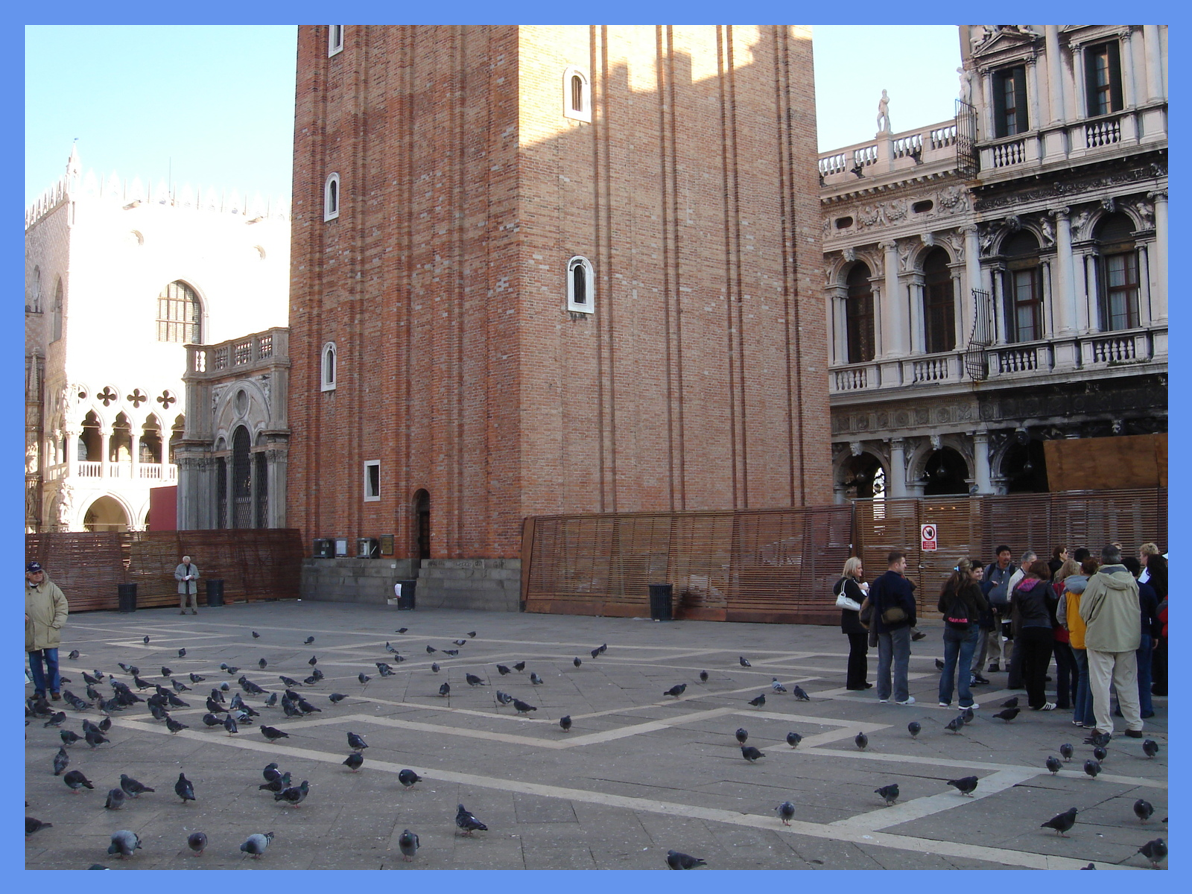} &
        \includegraphics[height=0.67cm, trim={0.2cm 0cm 0.4cm 0cm},clip]{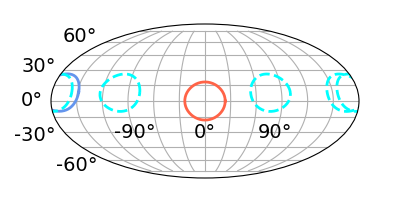} \\

    \end{tabular}
    }
    \caption{\textbf{Visualizing the top-5 predictions for non-overlapping pairs with high geodesic top-1 errors}. As illustrated by the examples above and also in our quantitative ablations, valuable insights can be found not only in the top predictions, but also among the subsequent selections. Images on the \textcolor{RedOrange}{left} serve as the reference points, and their coordinate system detemines the relative rotation, which defines the images on the \textcolor{RoyalBlue}{right}. The ellipsoids representing the ground truth are color-coded to match their respective images, with the estimated relative rotation illustrated by a cyan dashed line. 
}
    \label{fig:Top 5}
\end{figure}

In Figure~\ref{fig:Top 5}, we demonstrate the top-5 predictions of our model for several non-overlapping pairs from the \megadepth{} test set with high geodesic top-1 errors. 
To identify the \emph{Top 5} predictions, we start with the 360-D yaw angle output, and apply a softmax function to it. Next, we smooth the distribution by applying a Gaussian kernel on it (using a standard deviation of 5). Finally, we locate the top 5 local maximas on the smoothed 
output.

\begin{figure}[ht]
\centering
    \includegraphics[height=4.0cm, trim={0 0 0 0},clip]{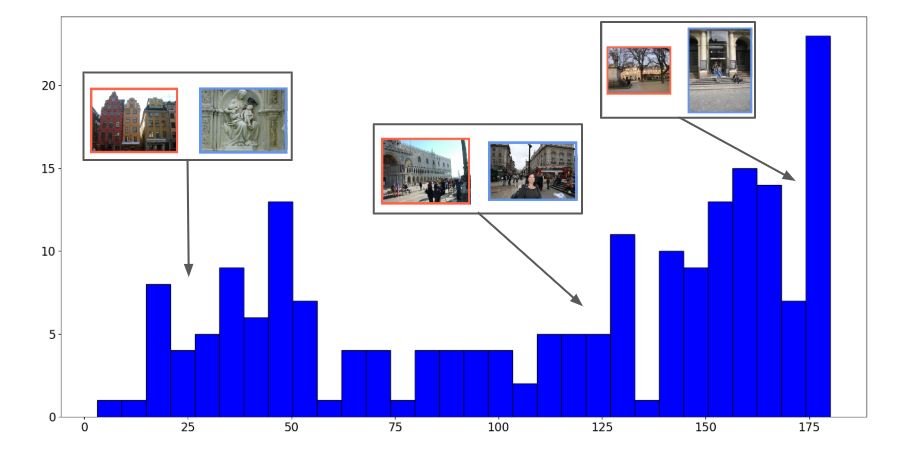}
    
    \caption{\textbf{Histogram of overall angle estimation of 200 randomly sampled image pairs from different scenes}. 
    }
    \label{fig:mixed_images_hist}
\end{figure}

We also examined the behavior of pairs sampled from entirely different places. We conducted an experiment over 200 such image pairs, sampled from different scenes. Their histogram (Figure ~\ref{fig:mixed_images_hist}) reveals that the overall angle spans almost the entire range, with a noticeable tendency to larger angles (the overall angle of $64.5\%$ of the pairs is above 90 degrees).

\definecolor{graytext}{RGB}{130,130,130}
\begin{table}[t]
\setlength{\tabcolsep}{1.7pt}
 \def\arraystretch{1.2}
\centering
\resizebox{0.45\textwidth}{!}{%
\begin{tabular}{lllccccccccccccccccc} 
    \toprule
    & \multicolumn{6}{c}{}&\multicolumn{3}{c}{Top 1} & & \multicolumn{3}{c}{Top 5} \\
     \cline{8-10} \cline{12-14}
     &Overlap && \fovexp{} & \instructexp{} & \shortdataset{} &&MGE\textdownarrow & RRA$_{15}$\textuparrow &RRA$_{30}$ \textuparrow&& MGE\textdownarrow & RRA$_{15}$\textuparrow & RRA$_{30}$\textuparrow \\
    \midrule

      \multirow{15}{*}{\rotatebox[origin=c]{90}{\cambridge{}} } & Large &&   $\times$&$\times$& $\times$&&  8.17 & 86.4 & 97.1 && 8.17 & 87.5 & 98.1\\
      
      & && $\vee$&$\vee$&$\vee$ && 2.87  & \textbf{97.7}  & \textbf{98.8}  && 2.82  & \textbf{99.2}  & \textbf{99.8}  \\

    & && \checkmark&$\times$& $\times$&& 3.18 & 98.2 & 98.2 && 3.21 & 97.9& 98.7 \\

    & &&\checkmark&\checkmark& $\times$ && 3.10 & 99.4 & 99.5 && 4.04 & 98.5 & 98.9\\

    & && \checkmark&\checkmark& \checkmark &&   \textbf{2.45} & {96.7} & {96.8} && \textbf{2.44} & {96.8} & {97.0} \\ 

    &Small && $\times$&$\times$& $\times$ && 26.44 & 6.3 & 60.2 && 22.32 & 17.4 & 69.5\\

    & && $\vee$&$\vee$&$\vee$&& 5.72  & \textbf{87.7}  & \textbf{92.1}  && 5.56  & \textbf{92.6}  & \textbf{97.7}  \\

    & && \checkmark&$\times$& $\times$ && 5.14 & 79.8 & 82.6 && 5.24 & 85.5 & 92.1 \\

     & && \checkmark&\checkmark& $\times$ && 5.91 & 83.9 & 88.4 && 5.84 & 88.3 & 95.5 \\

    & && \checkmark&\checkmark& \checkmark &&  \textbf{4.35} & {88.3} & {89.0} &&  \textbf{4.15} & {91.6} & {93.7} \\

    &None && $\times$&$\times$& $\times$ && 76.63 & 17.9 & 32.9 && 17.54 & 42.0 & 78.9 \\

    & && $\vee$&$\vee$&$\vee$&& 23.92  & 41.5  & 52.2  && 11.51  & 64.7  & 87.7  \\

    & && \checkmark&$\times$& $\times$ && 28.22 & 42.4 & 49.7 && 10.52 & 71.0 & 90.8 \\

    & && \checkmark&\checkmark& $\times$ && 14.20 & 50.1 & 60.1 && 9.76 & 72.7 & 92.6 \\

     & && \checkmark&\checkmark& \checkmark &&  \textbf{13.62} & \textbf{52.7} & \textbf{59.7} &&   \textbf{9.25} & \textbf{75.4} & \textbf{93.6} \\
     
    \midrule
     \multirow{15}{*}{{\rotatebox[origin=c]{90}\megadepth{}}} & Large && $\times$&$\times$& $\times$ && 13.65 & 35.4 & 73.5 && 12.22 & 61.4 & 84.7\\

     & && $\vee$&$\vee$&$\vee$&& 2.76  & 97.0 & 98.2  && 2.72 & 97.9  & \textbf{99.6}  \\

    & && \checkmark&$\times$& $\times$ && 4.61 & 79.7 & 81.1 && 4.41 & 90.3& 98.9\\

    & && \checkmark&\checkmark& $\times$ && 4.46 & 90.4 & 92.4 && 4.43 & 94.3 & 99.1 \\

    & && \checkmark&\checkmark& \checkmark && \textbf{2.41}  & \textbf{97.5}  & \textbf{97.9}  && \textbf{2.41}  &  \textbf{98.4}  & {99.4} \\

    &Small&& $\times$&$\times$& $\times$ && 55.28 & 3.7 & 29.1 && 29.83 & 15.0 & 50.3  \\

    & && $\vee$&$\vee$&$\vee$&& 6.46  & 81.9  & 90.1  && 5.99  & 86.1  & \textbf{97.7}  \\

    & && \checkmark&$\times$& $\times$ && 12.91 & 56.2 & 68.2 && 10.97 & 66.0 & 85.4 \\

    & && \checkmark&\checkmark& $\times$ && 11.46 & 62.5 & 80.6 && 10.73 & 68.0 & 91.0 \\

     & && \checkmark&\checkmark& \checkmark &&  \textbf{4.47} &  \textbf{87.2} & \textbf{91.6} && \textbf{4.24} &  \textbf{91.1} &  {97.2}  \\

    &None && $\times$&$\times$& $\times$ && 74.94 & 12.8 & 25.3 && 25.11 & 26.1 & 58.8  \\

    & && $\vee$&$\vee$&$\vee$ && 65.74  & 24.1  & 37.7  && 16.73  & 45.1  & 72.6  \\

    & && \checkmark&$\times$& $\times$ && 61.62 & 25.0 & 38.4 && 16.82 & 44.2 & 75.0 \\
    & && \checkmark&\checkmark& $\times$ && 68.31 & 25.0 & 36.1 && 16.21 & 45.7 & 78.2 \\
      & && \checkmark&\checkmark& \checkmark && \textbf{26.97} &  \textbf{36.1} &  \textbf{50.7} &&  \textbf{12.85} & \textbf{57.1} &  \textbf{85.8}  \\
    \bottomrule
\end{tabular}
}
\caption{Ablation study, evaluating the effect of our progressive training scheme over the two test sets (\cambridge{} in the top rows, and \megadepth{} in the bottom rows). All experiments start with the cropped panoramas used in Cai \emph{et al.}~\cite{cai2021extreme}. We also assess the necessity of the  progressive training scheme compared to training on all pairs simultaneously. Comparison denoted with $\vee$ is training with all of the datasets together. Best results are in bold. }
\label{tab:ablation_ch_val}
\end{table}
\definecolor{graytext}{RGB}{130,130,130}
\begin{table}[t]
\setlength{\tabcolsep}{1.7pt}
 \def\arraystretch{1.2}
\centering
\resizebox{0.45\textwidth}{!}{%
\begin{tabular}{lllccccccccccccccccc} 
    \toprule
    & & \multicolumn{3}{c}{}&\multicolumn{3}{c}{Top 1} & & \multicolumn{3}{c}{Top 5} \\
     \cline{6-8} \cline{10-12}
     &Overlap & KP & SM &&MGE\textdownarrow & RRA$_{15}$\textuparrow &RRA$_{30}$ \textuparrow&& MGE\textdownarrow & RRA$_{15}$\textuparrow & RRA$_{30}$\textuparrow \\
     \midrule
    \multirow{9}{*}{\rotatebox[origin=c]{90}{\cambridge{}} } & {Large} & $\times$ & $\times$ && \textbf{2.27} & {97.9}  & {98.4}  && \textbf{2.27}  &98.0  & 98.6 \\
    &&  $\times$ & \checkmark && 2.31 & 96.8  & 97.1  && 2.28  &98.0  & 99.0 \\
      & & \checkmark & $\times$ && 2.40 & \textbf{99.5}  & \textbf{99.7}  && 2.40  &\textbf{99.5}  & \textbf{99.8} \\
    & & \checkmark & \checkmark &&   2.45 & {96.7} & {96.8} && 2.44 & {96.8} & {97.0} \\
    &{Small} & $\times$ & $\times$ && 12.72 & 91.1  & 92.7  && 4.56  &94.0  & 97.1 \\ 
    && $\times$ & \checkmark && 4.07 & 87.8  & 89.1  && \textbf{4.00}  &91.8  & 96.1 \\
    && \checkmark & $\times$  && \textbf{4.04} & \textbf{92.9}  & \textbf{93.8}  && \textbf{4.00}  &\textbf{96.3}  & \textbf{98.8} \\
    & & \checkmark & \checkmark &&  {4.35} & {88.3} & {89.0} &&  {4.15} & {91.6} &{93.7} \\
    &{None} & $\times$ & $\times$ && 22.85  & 42.9  & 51.0  && 10.48  & 66.6  & 86.6  \\ 
    && $\times$ & \checkmark && 16.41 & 47.9 & 54.5  && 10.26  &68.7  & 85.5 \\
    && \checkmark & $\times$ && \textbf{12.92}  &   \textbf{55.0}  &  \textbf{63.2}  && 9.43  &  \textbf{76.8}  &  \textbf{95.3} \\
    & & \checkmark & \checkmark &&  {13.62} & {52.7} & {59.7} &&   \textbf{9.25} & {75.4} & {93.6} \\
    \hline
    
    \multirow{9}{*}{\rotatebox[origin=c]{90}{\megadepth{} }}
    & {Large}  & $\times$ & $\times$ && \textbf{2.18}  & 97.4  & 98.1  && \textbf{2.18}  & 97.4  & 98.1  \\ 
    && $\times$ & \checkmark && 2.30  & 97.0  & 97.4  && 2.30  & \textbf{98.5}  & 99.4  \\
    && \checkmark & $\times$ &&  2.44 & \textbf{97.6}  &  \textbf{98.3}
    &&  2.31  & 98.4  & \textbf{99.7}  \\
    & & \checkmark & \checkmark && 2.41  & 97.5  & {97.9}  && 2.41  &  98.4  & 99.4 \\
    &{Small} & $\times$ & $\times$ && 4.50 & 87.9 & 91.6 && 4.50  & 87.9  & 91.7  \\ 
    && $\times$ & \checkmark && 4.49  & \textbf{88.1}  & 92.0  && 4.46 & 91.2  & 96.7  \\
    & & \checkmark & $\times$ && \textbf{4.41} & 87.5  & \textbf{92.2}   && 4.32  & \textbf{91.9}  & \textbf{97.6} \\
    & & \checkmark & \checkmark &&  4.47 &  87.2 & {91.6} &&  \textbf{4.24} &  91.1 &  97.2  \\
    &{None} 
    & $\times$ & $\times$ && 48.81  & 34.0  & 44.1  && \textbf{12.56}  & \textbf{57.5}  &  84.6  \\ 
    
    && $\times$ & \checkmark && 43.07  & 31.2  & 44.2  && 13.99  & 53.5  & 83.2  \\ 
    && \checkmark & $\times$ && 41.39 & 35.3  & 46.8  
    && 13.04  & 56.9  & \textbf{86.2}  \\
    & & \checkmark & \checkmark &&   \textbf{26.97} &  \textbf{36.1} &  \textbf{50.7} &&  12.85 & 57.1 &  85.8  \\
    \bottomrule
\end{tabular}
}
\caption{Ablation study, evaluating the effect of the auxiliary channels added as input to our network. We train models without the keypoints and matches (KP) and without the segmentation maps (SM), and compare with our full model that is provided with both.} 
\label{tab:auxillary_long_val}
\end{table}

 \section{Additional Ablations}
 We report results over the \cambridge{} test set for the ablations discussed in the main paper. Table \ref{tab:ablation_ch_val} shows the effect of our progressive training scheme, while Table \ref{tab:auxillary_long_val} demonstrates the impact of adding additional channels. As illustrated, these results are consistent with the results reported in the main paper over the \megadepth{} test set. 
  Next we provide additional ablations which further ablate various design considerations and our progressive training scheme.

 \subsection{The impact of our progressive training scheme}
 To further evaluate the efficacy of our progressive training approach, we compared it to training on all pairs simultaneously. We used a pre-trained model initially trained on panoramic images with a 90 degree field of view (FoV). All-at-once model (indicated by $\vee$  was then further trained on a combined dataset comprising \fovexp{}, \instructexp{}, and \dataset{} training sets, using three separate batches for each dataset. The results are presented in Table \ref{tab:ablation_ch_val}, and demonstrate that our progressive training approach is particularly beneficial in challenging scenarios, such as when there is no overlap with a single camera (\cambridge{} test set), and for the \megadepth{} test set. This method helps the model gradually adapt to the diverse challenges posed by real-world images. Another interesting finding from this ablation is the improved generalization capabilities of our model, trained only on 90 degrees FOV panorama perceptive images, in comparison to prior work \cite{cai2021extreme} for in-the-wild image pairs. This result supports our decision to switch to a LoFTR encoder that was trained on Internet images, which could narrow the disparity between real-life images and panorama crops. 
 
 \definecolor{graytext}{RGB}{130,130,130}
\begin{table}[t]
\setlength{\tabcolsep}{1.7pt}
 \def\arraystretch{1.02}
\centering
\resizebox{0.45\textwidth}{!}{%
\begin{tabular}{lllccccccccccccccccc} 
    \toprule
    & & \multicolumn{3}{c}{}&\multicolumn{3}{c}{Top 1} & & \multicolumn{3}{c}{Top 5} \\
     \cline{6-8} \cline{10-12}
     &Overlap & \instructexp{} & SM &&MGE\textdownarrow & RRA$_{15}$\textuparrow &RRA$_{30}$ \textuparrow&& MGE\textdownarrow & RRA$_{15}$\textuparrow & RRA$_{30}$\textuparrow \\
    \midrule

    & {Large}  & $\times$ & $\times$ && 2.43  & 97.3  & 98.1  && 2.34  & 98.0  & 99.5  \\ 
    && $\times$ & \checkmark && \textbf{2.40}  & \textbf{97.6}  & 98.1  && 2.39  & \textbf{98.5}  & 99.4  \\
    && \checkmark & $\times$ &&  2.44 & \textbf{97.6}  &  \textbf{98.3}
    &&  \textbf{2.31}  & 98.4  & \textbf{99.7}  \\
    & & \checkmark & \checkmark && 2.41  & 97.5  & {97.9}  && 2.41  &  98.4  & 99.4 \\
    &{Small} & $\times$ & $\times$ && 4.64  & 86.7  & 90.6  && 4.44  & 90.3  & 97.2  \\ 
    && $\times$ & \checkmark && 4.50  & 87.2  & 92.0  && 4.34  & 91.3  & \textbf{97.8}  \\
    & & \checkmark & $\times$ && \textbf{4.41} & 87.5  & \textbf{92.2}   && 4.32  & \textbf{91.9}  & {97.6} \\
    & & \checkmark & \checkmark &&  4.47 &  87.2 & {91.6} &&  \textbf{4.24} &  91.1 &  97.2  \\
    &{None} 
    & $\times$ & $\times$ && 46.08  & 35.4  & 46.1  && 13.35  & 55.8  & 83.8  \\ 
    && $\times$ & \checkmark &&40.01  & 36.1  & 50.2 && 13.21  & 56.3  & 84.4  \\
    && \checkmark & $\times$ && 41.39 & 35.3  & 46.8  
    && 13.04  & 56.9  & \textbf{86.2}  \\
    & & \checkmark & \checkmark &&   \textbf{26.97} &  \textbf{36.1} &  \textbf{50.7} &&  \textbf{12.85} & \textbf{57.1} &  85.8  \\
    \bottomrule
\end{tabular}
}
\vspace{-8pt}
\caption{Ablation study, evaluating the effect of the priors of heavy neural networks - InstructPix2Pix and SegFormer.} 
\label{tab:no_heavy_net}
\end{table}

To assess the contribution of our components, we conducted an ablation study removing both InstructPix2Pix and SegFormer networks (Table \ref{tab:no_heavy_net}),  
finding that results are slightly worse in this case. For example, for non-overlapping pairs the MGE is 46.08$^\circ$.  Importantly, even without these additional cues, our method still outperforms the baselines in the challenging no-overlapping scenario, demonstrating that while these networks do enhance our results, our improved performance is not solely dependent on these additional cues.

\ignorethis{

\subsection{The impact of adding additional channels} \hec{is this what we call it in the paper?}
\hec{please first explain what we are ablating here before reporting performance}
Table \ref{tab:auxillary}  we can see that
the model without the additional channels shows inferior performance across all overlapping categories for top1. However, it achieves comparable results for top5, suggesting that the additional channels help the model in selecting the correct peak. 
For overlapping cases, the model with keypoints but without the segmentation map often yields the lowest errors; for instance, we can see that it obtains the highest RRA scores in the large overlap category of both test sets and in the small overlap category for \cambridge{} test set.
This suggests that in situations with visual overlap, the segmentation map does not contribute much to the model's performance. The keypoint and matches masks, on the other hand, offer more illumination-invariant geometric information to the model, enhancing the prediction of the relative rotation also in overlapping cases. These masks improve the model’s adaptability to different camera intrinsic parameters and help determine the areas of overlap, as evidenced by the improved performance across all categories.

}
\definecolor{graytext}{RGB}{130,130,130}
\begin{table}[t]
\setlength{\tabcolsep}{1.7pt}
 \def\arraystretch{1.2}
\centering
\resizebox{1\columnwidth}{!}{%
\begin{tabular}{llccccccccccccccccc} 
    \toprule
    & & && &&\multicolumn{3}{c}{Top 1} & & \multicolumn{3}{c}{Top 5} \\
     \cline{7-9} \cline{11-13}
     &Overlap  & \cite{cai2021extreme} & \fovexp{} & \instructexp{} &&MGE\textdownarrow & RRA$_{15}$\textuparrow &RRA$_{30}$ \textuparrow&& MGE\textdownarrow & RRA$_{15}$\textuparrow & RRA$_{30}$\textuparrow \\
    \midrule

    \multirow{12}{*}{\rotatebox[origin=c]{90}{\cambridge{}} } & {Large} & $\times$&$\times$& $\times$ && 11.89  &63.8  & 97.3 && 11.28  & 68.3 & 98.8 \\
    & & \checkmark&$\times$& $\times$ && 3.12  & 98.9  & 99.5 && 2.96  & 99.2 & 99.6 \\
    & & \checkmark&\checkmark& $\times$ && \textbf{2.42}  &96.3  & 96.4 && \textbf{2.42}  & 96.4 & 96.7 \\
    &  & \checkmark& \checkmark& \checkmark &&   2.45 & \textbf{96.7} & \textbf{96.8} && 2.44 & \textbf{96.8} & \textbf{97.0} \\
    &{Small} & $\times$&$\times$& $\times$ && 37.39  & 14.4  & 36.9 && 18.61  & 38.9 & 77.5 \\
    & & \checkmark &$\times$& $\times$ && 6.34  &83.0  & 89.6 && 5.84  & 88.3 & 96.4 \\
    & & \checkmark & \checkmark& $\times$ && 4.39  &87.3  & 88.4 && 4.24  & 90.9 & 93.5 \\
    & & \checkmark & \checkmark& \checkmark &&  \textbf{4.35} & \textbf{88.3} & \textbf{89.0} &&  \textbf{4.15} & \textbf{91.6} & \textbf{93.7} \\
    &{None}& $\times$&$\times$& $\times$ && 107.20  & 5.5  & 13.5 && 23.31  & 29.5 & 63.3 \\
    & & \checkmark &$\times$& $\times$ && 24.26  &37.6  & 53.8 && 12.29  & 61.3 & 84.8 \\
     & & \checkmark & \checkmark& $\times$ && 14.58  &50.8  & 58.4 && 9.47  & 74.6 & 93.2 \\
    &  & \checkmark & \checkmark& \checkmark &&  \textbf{13.62} & \textbf{52.7} & \textbf{59.7} &&   \textbf{9.25} & \textbf{75.4} & \textbf{93.6} \\
    \midrule
    \multirow{12}{*}{\rotatebox[origin=c]{90}{\megadepth{} }}
 & {Large} & $\times$ & $\times$& $\times$ && 10.07 & 69.5  & 95.9 && 9.47  & 75.1 & 98.7 \\
    & & \checkmark & $\times$ & $\times$ && 3.35  & 94.2  & 96.5 && 3.18  & 95.2 & 97.5 \\
   & & \checkmark & \checkmark& $\times$ && \textbf{2.40}  & \textbf{97.6}  & \textbf{98.1} && \textbf{2.39}  & \textbf{98.5} & \textbf{99.4} \\
    &  & \checkmark & \checkmark& \checkmark && {2.41}  & {97.5}  & {97.9}  && {2.41}  &  {98.4}  & \textbf{99.4} \\
    
    &{Small} & $\times$ & $\times$ & $\times$ && 30.26  & 21.5  & 49.3 && 14.39  & 52.2 & 83.6 \\
    & & \checkmark & $\times$ & $\times$ && 9.48  & 67.8  & 83.7 && 8.34  & 72.7 & 90.8 \\
    & & \checkmark & \checkmark& $\times$ && 4.50  & \textbf{87.2}  & 92.0 && 4.34  & \textbf{91.3} & \textbf{97.8} \\
    &  & \checkmark & \checkmark& \checkmark &&  \textbf{4.47} &  \textbf{87.2} & \textbf{91.6} && \textbf{4.24} &  {91.1} &  {97.2}  \\
    &{None} & $\times$&$\times$& $\times$ && 75.44  & 4.8  & 18.8 && 23.58  & 26.0 & 63.1 \\
    &&  \checkmark & $\times$ & $\times$ && 48.77  & 24.1  & 39.3 && 16.43  & 46.7 & 79.9 \\
    &  & \checkmark & \checkmark& $\times$ && 29.74  & \textbf{36.1}  & 50.2 && 13.21  & 56.3 & 84.4 \\
    &  & \checkmark & \checkmark& \checkmark &&   \textbf{26.97} &  \textbf{36.1} &  \textbf{50.7} &&  \textbf{12.85} & \textbf{57.1} &  \textbf{85.8}  \\
    \bottomrule
\end{tabular}
}
\caption{We evaluate to what extent the progressive training scheme is needed, in contrast to simply training on real image pairs. In the first rows (three $\times$'s), we train models on \dataset{} \emph{only}. In the rows below, we first train on the panoramas, according to our progressive training scheme. As illustrated above, training on panoramas, along with the various data augmentations we propose, significantly improve performance, particularly for non-overlapping image pairs. } 
\label{tab:data_aug_val}
\end{table}
 To demonstrate that the improved performance cannot be simply obtained with the \dataset{} training set, we report performance obtained without training on images cropped from panoramas in Table \ref{tab:data_aug_val}. We also compare performance to models trained on multiple stages of our progressive training scheme, followed by training on the \dataset{} dataset as the final step.
As illustrated in Table \ref{tab:data_aug_val}, 
relying solely on real image pairs from the \shortdataset{} train set yields significant performance degradation, particularly for extreme scenarios. Initialization with images cropped from panoramas improves performance. 
In particular, for large overlap on the \cambridge{} test set, it achieves the best RRA percentages. 
However, as overlap between the images decreases, the errors remain relatively high. Our data augmentations allow for further improving performance, by creating diverse perspective images that better resemble the real samples. 

 \definecolor{graytext}{RGB}{130,130,130}
\begin{table}[t]
\setlength{\tabcolsep}{1.7pt}
 \def\arraystretch{1.2}
\centering
\resizebox{0.45\textwidth}{!}{%
\begin{tabular}{lccccccccccccccccc} 
    \toprule
    && \multicolumn{2}{c}{Large} && \multicolumn{2}{c}{Small} && \multicolumn{2}{c}{None} \\
     \cline{3-4} \cline{6-7} \cline{9-10}
     Method &&MGE \textdownarrow& RRA$_{10}$ \textuparrow&&MGE \textdownarrow& RRA$_{10}$\textuparrow &&MGE\textdownarrow & RRA$_{10}$ \textuparrow\\
    \midrule
Ours w/o LT &&  \textbf{0.98} &  \textbf{100.0} &&   \textbf{1.06} & 98.7 && 2.61 & 87.7\\
Ours &&  1.06 &  \textbf{100.0} &&  1.09 &  \textbf{100.0} &&  \textbf{1.98} &  \textbf{96.4}\\
     \bottomrule
\end{tabular}
}
\caption{\textbf{Network Architecture Ablation}. For our model, we show ablated architecture, ablating the learnable tokens (LT). The best results are in bold.}
\label{tab:network-ablation_wo_LT}
\end{table}
\subsection{Architectural Ablations}
We conduct an ablation to evaluate the impact of using learnable tokens.  
 As illustrated in Table~\ref{tab:network-ablation_wo_LT}, the performance without learnable tokens significantly deteriorates for non overlapping pairs. Specifically, $RRA_{10}$ decreases from 96.4\% to 87.7\% and the median increases from 1.98 to 2.61.

\definecolor{graytext}{RGB}{130,130,130}
\begin{table}[t]
\setlength{\tabcolsep}{1.7pt}
 \def\arraystretch{1}
\centering
\resizebox{0.45\textwidth}{!}{%
\begin{tabular}{llcccccccccc} 

    \toprule
    \multicolumn{2}{c}{}&&\multicolumn{3}{c}{\cambridge{}}&&
    \multicolumn{3}{c}{\megadepth{}}\\
    \cline{4-6} \cline{8-10}
      & Method &&MGE\textdownarrow & RRA$_{15}$\textuparrow &RRA$_{30}$\textuparrow && MGE\textdownarrow & RRA$_{15}$ \textuparrow & RRA$_{30}$\textuparrow \\
    \midrule

    \multirow{2}{*}{\rotatebox[origin=c]{90}{Large}} & 
     DenseCorrVol~\cite{cai2021extreme} && 4.40 & 86.3 & 91.7 && 5.21 & 74.5 & 82.3 \\  
    & Ours&&   \textbf{2.41} & \textbf{96.1} & \textbf{96.1} &&   \textbf{2.41}  & \textbf{97.5}  & \textbf{97.9} \\
    \midrule
    \multirow{2}{*}{\rotatebox[origin=c]{90}{Small}} & 
     DenseCorrVol~\cite{cai2021extreme} && 15.4 & 49.2 & 59.2 && 63.54 & 22.2 & 32.1 \\
    & Ours &&   \textbf{4.27} & \textbf{87.4} & \textbf{88.4} &&  \textbf{4.47} &  \textbf{87.2} & \textbf{91.6} \\
    \midrule
    \multirow{2}{*}{\rotatebox[origin=c]{90}{None}} & 
     DenseCorrVol~\cite{cai2021extreme} && 103.97 & 17.9 & 29.4 && 95.46 & 11.2 & 19.0 \\
    & Ours &&   \textbf{14.16} & \textbf{51.4} & \textbf{58.4} &&   \textbf{26.97} &  \textbf{36.1} &  \textbf{50.7}\\
    \bottomrule
\end{tabular}
}
\vspace{-8pt}
\caption{\textbf{Finetune baseline with \instructexp{} and \fovexp{}.} We evaluate performance over the  \cambridge{} and \megadepth{} test sets, separately considering Large (top), Small (middle) and Non-overlapping (bottom) pairs. }

\label{tab:dcv_w_data}
\end{table}
 We further investigated architectural differences through an additional ablation where we applied our progressive training scheme to the baseline models(Table~\ref{tab:dcv_w_data}). This experiment further shows that prior models are not directly applicable for real-world settings, as these baselines perform significantly worse across all metrics, particularly for Small and Non-overlapping cases. For instance, DenseCorrVol yields a MGE of 63.54$^\circ$ (Small) and 95.46$^\circ$ (None).
 
 \ignorethis{
\subsection{Generalization to in-the-wild image pairs} \ruojin{what does "real images" mean? Usually, people might think generalization to real images means we train on synthetic images, maybe we want to say something like real images in the wild?}
\hani{is it better than generalization to real images? do yo think that I should insert real also? "in-the-wild real image pairs"?- I changed below as well}\ruojin{"in-the-wild image pairs" is clearer than "real images" to me, maybe Hadar can take another pass on this.} \hec{is this really an ablation? or just an additional experiment?}
In Table~\ref{tab:generalization} we evaluate the models’ capacity to adapt and apply their learning to in-the-wild image pairs. The training of the models was conducted using panorama perspective images that have a 90-degree field of view. These models are then tested using the \cambridge{} and \megadepth{} test sets to determine their generalization capabilities. The data presented in the table reinforces our decision to switch to a LoFTR encoder that was trained on Internet images, which could narrow the disparity between real-life images and panorama crops. It is evident that the Dense Correlation Volumes method falls short in performance, particularly in scenarios with a large overlap, where even without extra channels, our model outperforms it by approximately a factor of two. Moreover, the advantages of the extra channels are clearly demonstrated, as our model that includes them shows superior performance across almost all metrics.

\definecolor{graytext}{RGB}{130,130,130}
\begin{table}[t]
\setlength{\tabcolsep}{1.7pt}
 \def\arraystretch{1}
\centering
\resizebox{0.45\textwidth}{!}{%
\begin{tabular}{llcccccccccc} 

    \toprule
    \multicolumn{2}{c}{}&&\multicolumn{3}{c}{\cambridge{}}&&
    \multicolumn{3}{c}{\megadepth{}}\\
    \cline{4-6} \cline{8-10}
      & Method &&MGE & RRA$_{15}$ &RRA$_{30}$ && MGE & RRA$_{15}$ & RRA$_{30}$ \\
    \midrule

    \multirow{3}{*}{\rotatebox[origin=c]{90}{Large}} 
     & DenseCorrVol~\cite{cai2021extreme} &&   98.51 & 25.9 & 33.3 &&   120.53 & 7.0 & 13.0 \\
    &\casenc{}&&   29.75 & 42.7 & 50.0 && 170.62 & 7.3 & 9.2\\
    & wo additional channels &&  8.40 & 73.8 & 81.6 &&  170.29 & 20.6 & 27.6\\
    & Ours &&  \textbf{8.17} & \textbf{86.4} & \textbf{97.1} &&  \textbf{13.65} & \textbf{54.5} & \textbf{73.5}\\
    \midrule
    \multirow{3}{*}{\rotatebox[origin=c]{90}{Small}} & DenseCorrVol~\cite{cai2021extreme} &&   101.93 & 3.9 & 19.5 &&   99.03 & \textbf{4.9} & 18.7 \\
    & \casenc{}&&   139.14 & 2.7 & 4.4 &&   148.44 & 0.6 & 3.0 \\
    & wo additional channels &&  119.20 & 2.2 & 28.7 &&   147.93 & 1.3 & 11.1\\
    & Ours && \textbf{26.44} & \textbf{6.3} & \textbf{60.2} &&  \textbf{55.28} & 4.1 & \textbf{29.1}\\
    \midrule
    \multirow{3}{*}{\rotatebox[origin=c]{90}{None}} & DenseCorrVol~\cite{cai2021extreme} && 86.15 & 6.4 & 23.0 &&   88.71 & 4.2 & 12.7 \\
    & \casenc{}&&   \textbf{69.69} & 8.4 & 23.1 &&   78.60 & 0.0 & 20.8 \\
    & wo additional channels &&  70.41 & \textbf{26.6} & \textbf{37.3} &&   79.83 & 11.4 & 24.1\\
    & Ours &&  76.63 & 17.9 & 32.9 && \textbf{74.46} & \textbf{12.8} & \textbf{25.3}\\
    \bottomrule
\end{tabular}
}
\caption{\textbf{Generalization to in-the-wild image pairs.} The above models were trained on panorama perspective images of 90 degrees FoV. "wo additional channels" is our model without the additional channels.}

\label{tab:generalization}
\end{table}


}
\ignorethis{
  \definecolor{graytext}{RGB}{130,130,130}
\begin{table}[t]
\setlength{\tabcolsep}{1.7pt}
 \def\arraystretch{1.2}
\centering
\resizebox{1\columnwidth}{!}{%
\begin{tabular}{llccccccccccccccccc} 
    \toprule
    & & && &&\multicolumn{3}{c}{Top 1} & & \multicolumn{3}{c}{Top 5} \\
     \cline{7-9} \cline{11-13}
     &Overlap  & \cite{cai2021extreme} & \fovexp{} & \instructexp{} && MGE & RRA$_{15}$ &RRA$_{30}$ && MGE & RRA$_{15}$ & RRA$_{30}$ \\
    \midrule

    \multirow{9}{*}{\rotatebox[origin=c]{90}{\cambridge{}} } & {Large} 

     & \checkmark&$\times$& $\times$ && 2.82  & 99.2  & 99.6  && 2.72  & 99.4  & 99.8  \\

    & & $\vee$&$\vee$&$\vee$ && 2.91  & \textbf{99.8}  & \textbf{99.8}  && 2.87  & \textbf{99.8}  & \textbf{100.0}  \\
    
   &  & \checkmark& \checkmark& \checkmark && \textbf{2.37} & 95.8 & 95.8 && \textbf{2.37} & 95.9 & 96.1\\
    &{Small} 

     & \checkmark &$\times$& $\times$ && 5.88  & 84.9  & 91.4  && 5.51  & 89.6  & 97.1  \\
    & & $\vee$&$\vee$&$\vee$&& 5.87  & \textbf{89.4}  & \textbf{93.2}  && 5.47  & \textbf{93.0}  & \textbf{99.5}  \\

    & & \checkmark & \checkmark& \checkmark && \textbf{4.50} & 88.2 & 89.3&& \textbf{4.38} & 92.7 & 94.6  \\

    &{None}
    & \checkmark &$\times$& $\times$ && 19.00  & 43.8  & 56.3  && 10.72  & 67.8  & 88.0  \\

    & & $\vee$&$\vee$&$\vee$&& 16.40  & 47.9  & 59.5  && 10.23  & 70.7  & 93.5  \\

    &  & \checkmark & \checkmark& \checkmark && \textbf{13.27} & \textbf{52.7} & \textbf{59.9} &&  \textbf{9.17} & \textbf{76.9} & \textbf{94.0} \\
    \midrule
    \multirow{9}{*}{\rotatebox[origin=c]{90}{\megadepth{} }}
    & {Large} 
     & \checkmark & $\times$ & $\times$ && 2.96  & 96.0  & 97.3  && 2.88  & 96.9  & 98.5  \\
    & & $\vee$&$\vee$&$\vee$&& 2.67  & 96.9  & 98.0  && 2.62 & 98.2  & \textbf{99.7}  \\
    &  & \checkmark & \checkmark& \checkmark && \textbf{2.40}  & \textbf{97.7}  & \textbf{98.2}  && \textbf{2.39}  & \textbf{98.6}  & 99.4 \\
    
    &{Small}
    & \checkmark & $\times$ & $\times$ && 8.03  & 71.8  & 85.5  && 6.95  & 78.6  & 93.2  \\
    & & $\vee$&$\vee$&$\vee$&& 6.35  & 80.9  & 90.0  && 6.09  & 87.1  & 97.9  \\
    &  & \checkmark & \checkmark& \checkmark && \textbf{4.33} & \textbf{88.1} & \textbf{92.3} && \textbf{4.12} & \textbf{91.2} & \textbf{97.1}  \\

    &{None} 
    &  \checkmark & $\times$ & $\times$ && 44.66  & 27.2  & 42.0  && 14.96  & 50.4  & 83.4  \\

    & & $\vee$&$\vee$&$\vee$ && 66.87  & 26.1  & 36.2  && 15.53  & 47.3  & 75.0  \\
    &  & \checkmark & \checkmark& \checkmark &&   \textbf{27.49} & \textbf{36.2} & \textbf{50.5} && \textbf{13.03} & \textbf{57.1} & \textbf{85.1}  \\
    
    \bottomrule
\end{tabular}
}
\caption{\textbf{Progressive learning vs all at once} We evaluate to what extent the progressive training scheme is needed, in contrast to simply training on all pairs together. Comparison denoted with $\vee$ is training with all of the datasets together. }
\label{tab:all_vs_progressive}
\end{table}
 \subsection{Progressive training Vs. all at once} \hec{should we tie this to the previous subsection discussing the progressive training scheme?} In Table~\ref{tab:all_vs_progressive} we conducted an ablation study to assess the efficacy of our progressive training approach. We utilized a pre-trained model that was initially trained on panoramic images with a 90 degree field of view (FoV). All-at-once model (indicated by $\vee$  was then further trained on a combined dataset comprising \fovexp{}, \instructexp{}, and \dataset{} training sets, using three separate batches for each dataset. The results presented in the table demonstrate that our progressive training method is particularly beneficial in difficult scenarios, such as when there is no overlap with a single camera (\cambridge{} test set), and for the \megadepth{} test set. This approach aids the model in incrementally adapting to the diverse challenges presented by real-world images.
 }

\end{document}